\documentclass[a4paper,fleqn]{cas-dc}

\usepackage[numbers,sort&compress]{natbib}

\usepackage{array,graphicx}
\usepackage{lipsum}
\usepackage{etoolbox} 
\usepackage{booktabs}
\usepackage{pdflscape}
\usepackage{longtable, tabularx}
\usepackage{verbatim}
\usepackage{gensymb}
\usepackage{tikz}
\usepackage{makecell} 
\usepackage{booktabs}
\usepackage{multirow}
\usepackage{multicol} 
\usepackage{amsfonts}
\usepackage{amssymb,mathtools}
\usepackage{amsmath}
\usepackage{algpseudocode}
\usepackage[linesnumbered,ruled,vlined]{algorithm2e}
\usepackage[thinc]{esdiff}
\usepackage{enumitem}
\usepackage[svgnames]{xcolor}
\usepackage{threeparttable}
\usepackage{wasysym}
\usepackage{framed} 
\usepackage{nomencl} 
\usepackage{mdframed}
\usepackage{orcidlink}
\usepackage{hyperref}

\def\tsc#1{\csdef{#1}{\textsc{\lowercase{#1}}\xspace}}
\tsc{WGM}
\tsc{QE}
\tsc{EP}
\tsc{PMS}
\tsc{BEC}
\tsc{DE}

\makeatletter
\def\NAT@def@citea{\def\@citea{\NAT@separator}}
\makeatother

\newcommand\setrow[1]{\gdef\rowmac{#1}#1\ignorespaces}
\newcommand\clearrow{\global\let\rowmac\relax}
\clearrow

\ExplSyntaxOn
\NewDocumentCommand{\longdash}{ O{2} }
 {
  --\prg_replicate:nn { #1 - 1 } { \negthinspace -- }
 }
\ExplSyntaxOff

\newcommand{\mycomment}[1]{}

\ExplSyntaxOn

\NewDocumentCommand \vect { s o m }
 {
  \IfBooleanTF {#1}
   { \vectaux*{#3} }
   { \IfValueTF {#2} { \vectaux[#2]{#3} } { \vectaux{#3} } }
 }

\DeclarePairedDelimiterX \vectaux [1] {\lbrack} {\rbrack}
 { \, \dbacc_vect:n { #1 } \, }

\cs_new_protected:Npn \dbacc_vect:n #1
 {
  \seq_set_split:Nnn \l_tmpa_seq { , } { #1 }
  \seq_use:Nn \l_tmpa_seq { \enspace }
 }
\ExplSyntaxOff

\DeclareMathOperator*{\argmaxA}{arg\,max}
\newcommand{\cs}[1]{\texttt{\symbol{`\\}#1}}

\DeclareMathOperator*{\argminA}{arg\,min}

\makeatletter
\newcommand*{\compress}{\@minipagetrue}
\makeatother

\newcommand{\myfrac}[2]{%
    \setbox0\hbox{$#1$}        
    \dimen0=\wd0               
    \setbox1\hbox{$#2$}        
    \dimen1=\wd1               
    \ifdim\wd0<\wd1            
        \dfrac{#1\hfill}{#2}   
    \else                      
        \dfrac{#1}{#2\hfill}   
    \fi
}

\let\origclearpage\clearpage

\let\oldmaketitle\maketitle

\renewcommand{\maketitle}{%
  \oldmaketitle
  \let\clearpage\relax \onecolumn\tableofcontents\twocolumn\newpage
\hspace{1cm}
}

\makenomenclature
\renewcommand*\nompreamble{\begin{multicols}{2}}
\renewcommand*\nompostamble{\end{multicols}}

\newcommand{\colfigwidth}{\dimexpr\columnwidth-0.75em\relax}

\begin{document}
\let\WriteBookmarks\relax
\def\floatpagepagefraction{1}
\def\textpagefraction{.001}

\shorttitle{}

\shortauthors{Michael Herman et~al.}

\title [mode = title]{Sun sensor calibration algorithms: A systematic mapping and survey}                      

%
\author[1]{Michael Herman\,\orcidlink{0009-0001-1000-4553}}

\ead{hermanm@gatech.edu}



\affiliation[1]{organization={Aerospace Systems Design Laboratory, School of Aerospace Engineering, Georgia Institute of Technology},
    city={Atlanta},
    state={GA},
    postcode={30332}, 
    country={USA}}

\author[1]{Olivia J. Pinon Fischer\,\orcidlink{0000-0002-2233-1391}}

\ead{olivia.pinon@asdl.gatech.edu}

\author[1]{Dimitri N. Mavris\,\orcidlink{0000-0001-8783-4988}}

\ead{dimitri.mavris@aerospace.gatech.edu}


\begin{abstract}
Attitude sensors determine the spacecraft attitude through the sensing of an astronomical object, field or other phenomena. The Sun and fixed stars are the two primary astronomical sensing objects. Attitude sensors are critical components for the survival and knowledge improvement of spacecraft. Of these, sun sensors are the most common and important sensor for spacecraft attitude determination. The sun sensor measures the Sun vector in spacecraft coordinates.
The sun sensor calibration process is particularly difficult due to the complex nature of the uncertainties involved. The uncertainties are small, difficult to observe, and vary spatio-temporally over the lifecycle of the sensor. In addition, the sensors are affected by numerous sources of uncertainties, including manufacturing, electrical, environmental, and interference sources. This motivates the development of advanced calibration algorithms to minimize uncertainty over the sensor lifecycle and improve accuracy.
Although modeling and calibration techniques for sun sensors have been explored extensively in the literature over the past two decades, there is currently no resource that consolidates and systematically reviews this body of work. The present review proposes a systematic mapping of sun sensor modeling and calibration algorithms across a breadth of sensor configurations. It specifically provides a comprehensive survey of each methodology, along with an analysis of research gaps and recommendations for future directions in sun sensor modeling and calibration techniques. 
\end{abstract}



\maketitle\newpage

\nomenclature{A/D}{Analog/Digital}
\nomenclature{ACM}{Assembly Compensation Model}
\nomenclature{AEGNN}{Asynchronous Event-Based Graph Neural Networks}
\nomenclature{AKF}{Adaptive Kalman Filter}
\nomenclature{ANN}{Artificial Neural Network}
\nomenclature{BCM}{Basic Centroiding Method}
\nomenclature{BCTM}{Basic Centroiding Thresholding Method}
\nomenclature{BSCM}{Black Sun Centroiding Method}
\nomenclature{CCD}{Charge-Coupled Device}
\nomenclature{CHT}{Circle Hough Transform}
\nomenclature{CKF}{Cubature Kalman Filter}
\nomenclature{CMOS}{Complementary Metal-Oxide Semiconductor}
\nomenclature{CNN}{Convolutional Neural Network}
\nomenclature{CSS}{Coarse Sun Sensor}
\nomenclature{DBCM}{Double Balance Centroiding Method}
\nomenclature{DNN}{Deep Neural Network}
\nomenclature{DSS}{Digital Sun Sensor}
\nomenclature{ECM}{Electrical Compensation Model}
\nomenclature{EKF}{Extended Kalman Filter}
\nomenclature{ESCM}{Event Sensor Centroiding Method}
\nomenclature{FEIC}{Feature Extraction Image Correlation}
\nomenclature{FMMS}{Fast Multi-Point MEANSHIFT}
\nomenclature{FOV}{Field Of View}
\nomenclature{FSS}{Fine Sun Sensor}
\nomenclature{FZP}{Fresnel Zone Plate}
\nomenclature{HT}{Hough Transform}
\nomenclature{ICM}{Interference Compensation Model}
\nomenclature{IFM}{Image Filtering Method}
\nomenclature{KFFNNS}{Kalman Filter Family Neural Network In\\ Succession}
\nomenclature{LCE}{Lensless Compound Eye}
\nomenclature{LPD}{Linear Photodiode}
\nomenclature{LSQ}{Least Squares}
\nomenclature{LSI}{Laser Signal Injection}
\nomenclature{LTF}{Light-To-Frequency}
\nomenclature{LUT}{Look Up Table}
\nomenclature{MCAM}{Multiple Centroid Averaging Method}
\nomenclature{MMAE}{Multiple Model Adaptive Estimation}
\nomenclature{MT-ACM}{Multiple-Threshold Averaging Centroiding Method}
\nomenclature{NN}{Neural Network}
\nomenclature{NNTKFF}{Neural Network Trained Kalman Filter Family}
\nomenclature{OCM}{Optical Compensation Model}
\nomenclature{OI}{Opto-Isolator}
\nomenclature{PD}{Peak Detection}
\nomenclature{PDA}{Photodiode Array}
\nomenclature{PM}{PixelMax}
\nomenclature{PPE}{Peak Position Estimate}
\nomenclature{PSD}{Position Sensitive Device}
\nomenclature{PVA}{Photovoltaic Array}
\nomenclature{QPD}{Quadrant Photodiode}
\nomenclature{RKF}{Robust Kalman Filter}
\nomenclature{ROI}{Region Of Interest}
\nomenclature{SECM}{Space Environment Compensation Model}
\nomenclature{SPM}{Standard Projection Model}
\nomenclature{SSCNN}{Sparse Submanifold Convolutional Neural Network}
\nomenclature{STF-SNN}{Spatio-Temporal Fusion Spiking Neural Network}
\nomenclature{TM}{Template Method}
\nomenclature{TRL}{Technology Readiness Levels}
\nomenclature{UKF}{Unscented Kalman Filter}



\begin{table*}[!t]   
    \begin{framed}
        \printnomenclature
    \end{framed}
\end{table*}

\let\clearpage\origclearpage

\section{Introduction}


Attitude sensors determine the spacecraft attitude through the sensing of an astronomical object, field or other phenomena. The Sun and fixed stars are the two primary astronomical sensing objects. Attitude sensors are critical components for the survival and state knowledge of spacecraft. The sun sensor, magnetometer, star sensor, and Earth sensor make up this category. Of these, sun sensors are the most common and important sensor for small satellite attitude determination \cite{Zhu2022}. Nearly all low-Earth orbiting small satellites employ sun sensors as part of the attitude sensor package, which determine the satellite attitude by measuring the Sun vector relative to the satellite coordinates \cite{KeQiang2020}. Sun sensors have also proven essential in small satellite operations, where larger, and more resource-intensive sensors, may be impractical. In addition to their applications for satellite attitude determination, sun sensors are also used for terrestrial and non-terrestrial applications such as rovers. The two main categories of sun sensors are analog and digital.

\textbf{The need for calibration.} The main challenge for using sun sensors for attitude estimation is sensor errors. These errors limit the overall achievable attitude estimation accuracy. During development and in-orbit operation, the sensors are affected by numerous sources of uncertainties, including manufacturing, environmental, interference sources, or uncertainties inherent to the sensor architecture. Calibrating sun sensors is particularly challenging due to the complex nature of the associated uncertainties. These uncertainties are often small, difficult to detect, and may vary throughout the operational lifecycle of the sensor, necessitating in-flight calibration to maintain accuracy over time.

These challenges have motivated the development of enhanced calibration algorithms to improve upon the current state of the art in sun sensor performance and minimize uncertainty over the sensor lifecycle. In particular, they have been addressed through a number of approaches including novel sensor architectures, model representations, and feature extraction techniques to improve sensor performance. Some common model representations include algebraic, geometric, and physics-based relations.

\textbf{Challenges in calibration algorithms.} However, there is no one size fits all solution for sensor calibration as it typically involves trade-offs to achieve specific performance requirements, such as improving accuracy, reducing latency, or increasing reliability. Furthermore, sun sensor model representations are often tightly coupled to the sensor architecture, including the detector and mask configurations.

In some cases, existing calibration algorithms for sun sensor modeling and feature extraction are approaching their performance limits for specific applications and demand further innovation. Sun sensor model representations often lack generality and are unable to achieve configuration-agnostic calibration. They also tend to be inflexible in adapting to temporal errors throughout the sensor’s lifecycle. Furthermore, achieving high performance and minimizing model uncertainty requires a deep understanding of the sensor’s underlying physical behavior.

The feature extraction process is limited by sensor noise floor challenges and image processing techniques. In addition, feature extraction techniques are often shallow and reduce the problem to single or multi-point detection. As a result, the richness of the full feature space and levels of feature abstraction are lost.

Finally, the harshness of the space environment leads to more risk-averse operational conditions. As a result, spacecraft navigation differs from terrestrial applications due to high reliability requirements and a lack of comprehensive calibration datasets. This motivates the need for publicly available calibration datasets for the training and testing of sun sensor algorithms to enable a deeper understanding of sensor uncertainties.

\textbf{Motivation.} The longstanding interest in calibration and recent innovations in feature representation techniques have become increasingly relevant to the space domain with applications to sun sensors, star trackers \cite{Zapevalin2022,Mastrofini2023,Zhao2024}, Earth sensors \cite{Koizumi2018,Kikuya2023}, rendezvous \cite{Phisannupawong2020}, hazard avoidance \cite{Tomita2022}, and pose estimation \cite{Sharma2019,Sharma2020,Cassinis2023}. The varied number of calibration algorithms, application-specific approaches to sun sensor calibration, and the large body of research papers in this field motivates a systematic survey.

\textbf{Relevance of the proposed survey.} Only a few surveys related to sun sensors have been written so far. These include: A review by \citet{Conrado2018} details the current state of terrestrial sun position sensors and their architectures. For analog sun sensors, the study by \citet{Salazar2024} compares different architecture designs.

To the best of the authors’ knowledge, no systematic survey currently exists on modeling and calibration algorithms for sun sensors. This work addresses that gap by building on previous surveys and offering a structured categorization of the models and feature representations used in sun sensor calibration. It presents a comprehensive overview based on an analysis of 128 studies focused on the development and application of novel sun sensor architectures, modeling methods, and feature extraction techniques. Additionally, the paper outlines potential directions for future research and innovation in the field.

\textbf{Expected contributions.} This paper offers several key contributions. First, it introduces a taxonomy that categorizes sun sensor modeling and calibration algorithms, encompassing sensor architectures, feature extraction techniques, modeling representations, and their integration within the calibration framework. A major focus is placed on classifying different modeling representations and their roles in addressing specific types of errors. Second, the paper provides a detailed review of calibration methods available in the literature, along with an in-depth discussion on the use of various model representations and feature extraction techniques—such as non-physical, geometric, physics-informed, neural network-based, and centroid detection methods—in the calibration process. Third, it evaluates each calibration algorithm and offers recommendations tailored to specific application requirements. Finally, the paper identifies promising future directions for advancing calibration algorithms, drawing insights from both domain-specific and cross-domain literature. 

\textbf{Target audience.} The goal of this work is to equip potential new users of sun sensor calibration algorithms with established and successful methods. In addition, this work intends to equip practitioners with explanations of application-specific implementations and recommendations. For researchers, directions to improve upon the current state of the art are provided.

Given the wide range of calibration algorithms covered, this work does not delve into the detailed implementation of each method. Instead, its primary goal is to describe and analyze a diverse set of techniques to highlight the key research directions in sun sensor modeling and calibration algorithms.

\begin{figure*}[ht]
\noindent\rule{\textwidth}{1.25pt} 
\centerline{\includegraphics[width=\textwidth]{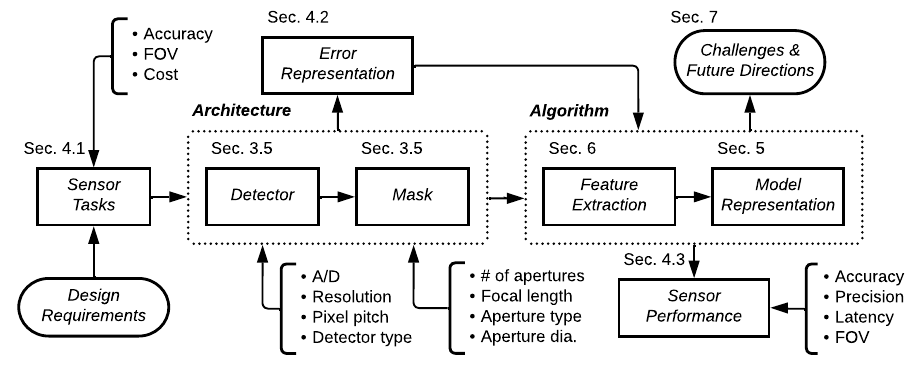}}
\noindent\rule{\textwidth}{1.25pt} 
\caption{Decision flow in sun sensor selection process.}
\label{fig:decisionflow}
\end{figure*}

\textbf{Organization of the paper.} The  organization of the paper follows the decision flow of a typical sun sensor selection process, as shown in Figure \ref{fig:decisionflow}. Section \ref{sec:ssmeasmod} presents a formulation of the concept of sun sensor operation. Section \ref{sec:resmethod} describes the literature selection process and presents a taxonomy of the field. Section \ref{sec:attribanalysis} presents a deeper analysis of the relationships between sensor attributes and provides key insights for sensor selection. Section \ref{sec:ssmodrep} discusses model representation techniques in more detail with a focus on model formulations. Section \ref{sec:fetech} discusses feature extraction techniques in detail with a focus on implementation and evaluation. 
Section \ref{sec:futuredir} outlines current challenges, offers practical guidance for practitioners, and recommends future research directions to advance the state of the art in sun sensor algorithms. Section \ref{sec:conc} offers concluding remarks. The dataset collected for the survey is referenced in Appendix \ref{sec:appendixa} with links to the Tableau Public and Zenodo repositories.

\section{Sun sensor measurement model} \label{sec:ssmeasmod}

\begin{figure}[ht]
	\centering
		\includegraphics[width=\colfigwidth]{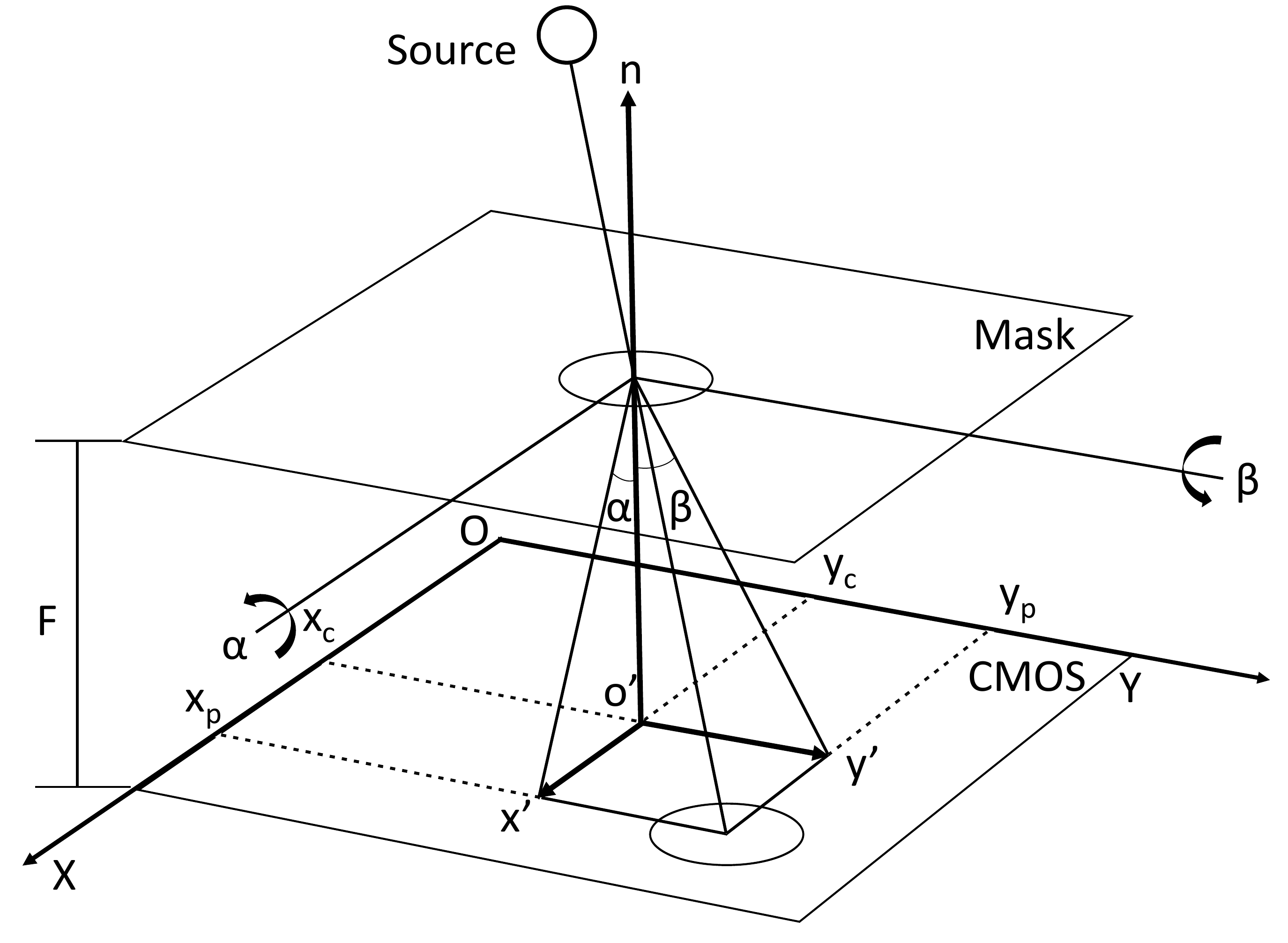}
	\caption{Ideal single-aperture sensor operating principle. The two sun angles $\alpha$ and $\beta$ are the values estimated in this study.}
	\label{fig:operbw}
\end{figure}

This section provides an overview of the sun sensor concept, its operating principles, and its architectural components. The sun sensor is employed to detect the satellite’s attitude angle with respect to the sun. Its working principle is illustrated in Figure \ref{fig:operbw}. Using the detector illumination information, the satellite’s orientation in space can be determined through a dedicated algorithm. 

The conventional sun sensor operates based on the following principle: mounted on the satellite, the sensor features a thin mask positioned above its chip surface, with a pinhole at the top. Sunlight passes through this pinhole, creating a spot of light known as a sun spot on the image sensor array. As the light enters through the hole, it forms a nearly circular spot on the detector’s sensing surface.

The distance F between the focal plane and the mask, which defines the system’s focal length, causes the position of the sun spot to shift as the incident angle of incoming sunlight changes. This shift allows the Sun vector to be measured based on the spot’s center position. The detector within the sensor package captures the location of the sun spot on the detector plane, and using this information, a calibration algorithm calculates the satellite’s attitude angle relative to the Sun.

\section{Research methodology} \label{sec:resmethod}

The goal of this study is develop a taxonomy of sun sensor calibration algorithms, explore relationships among key sensor attributes, present various algorithm implementations, and recommend future directions to advance the state of the art. This is achieved through a scoping review and a comprehensive survey of methodologies from existing literature.

\subsection{Research questions}

The following research questions have been formulated to guide a clear and effective exploration of the literature for both newcomers and experienced practitioners. The study begins with a high-level overview to map the landscape and examine key correlations. It then shifts focus to the detailed, low-level implementations of sensor algorithms. Finally, the study concludes by outlining recommended future directions based on the insights gained throughout the analysis.

\begin{description}\label{rq:rq1}
\item[RQ 1 \textit{Sensor attribute cross-analysis}:] \textit{How do the sensor main attributes affect algorithm selection?}
\end{description}

In \hyperref[rq:rq1]{RQ1}, the trends of sensor attributes on algorithm selection are investigated with the goal of providing a more informed selection approach. A summary of findings is provided in \hyperref[rqsum:rqsum1]{RQ 1 Summary}.

\begin{description}\label{rq:rq2}
\item[RQ 2 \textit{Model representation analysis}:] \textit{Which model representations are being used for sun sensor calibration algorithms and how are they formulated?}
\end{description}

In \hyperref[rq:rq2]{RQ2}, the model representations are classified and the formulations of key case-studies are presented. A summary of findings is provided in \hyperref[rqsum:rqsum2]{RQ 2 Summary}.

\begin{description}\label{rq:rq3}
\item[RQ 3 \textit{Feature extraction analysis}:] \textit{Which feature extraction methods are being used for sun sensor calibration algorithms and how are they implemented?}
\end{description}

In \hyperref[rq:rq3]{RQ3}, the feature extraction approaches are classified, key case-studies are presented, and the methods are compared. The goal of this research question is to assist practitioners in the selection and implementation of feature extraction algorithms for sun sensor calibration. A summary of findings is provided in \hyperref[rqsum:rqsum3]{RQ 3 Summary}.

\begin{description}\label{rq:rq4}
\item[RQ 4 \textit{Gaps in the literature}:] \textit{What are the current gaps and challenges in the field of sun sensor calibration algorithms?}
\end{description}

Finally, in \hyperref[rq:rq4]{RQ4}, challenges in the field are assessed and future research directions are recommended based on insights from this study. A summary of findings is provided in \hyperref[rqsum:rqsum4]{RQ 4 Summary}.

\subsection{Literature search and selection}

To carry out the systematic study, a two-step process was used for literature search and selection. First, an initial set of papers was identified through a primary database search. This was followed by iterative backward and forward snowballing to expand and refine the study set.

The initial set of papers was gathered through a single database search using the Georgia Tech Library collection (Ex Libris Primo). This set was further expanded through backward and forward snowballing of each included paper using Google Scholar.

To identify the initial set, the following search string was used in the primary database:\\


\textit{sun sensor calibration}\\

During the screening process, the first twenty pages of search results were evaluated based on paper titles. The resulting set of candidate papers was then filtered according to the study’s inclusion criteria. After twenty pages, the search was concluded due to saturation, as few additional papers met the criteria. At that point, the initial search phase was completed, and the second phase was initiated.

In the second step of the process backward and forward snowballing was used to extend the initial set of collected papers. The reference lists of each selected paper were screened for further relevant studies. In addition, the "\textit{cited by}" feature on Google Scholar was used to screen for relevant works that had cited each selected paper. This process was iterated over the collected literature until no further relevant papers could be found.

\subsection{Inclusion criteria}

In order to ensure that the collected body of literature is within the scope of the study, the following inclusion criteria are defined:

\begin{itemize}
    \item \textbf{I1:} Must be published in 2002 or later. This cutoff year was selected because, in 2002, the Jet Propulsion Laboratory developed a prototype micro digital sun sensor using micro-\\electromechanical systems (MEMS) technology and a CMOS sensor \cite{Liebe2002}. Additionally, the introduction of the CubeSat standard in 2003 further increased the demand for compact and precise sun sensor technologies. 
    \item \textbf{I2:} Must be written in English.
    \item \textbf{I3:} Must include and provide a detailed description of the implementation of a model representation or feature extraction algorithm.
    \item \textbf{I4:} Must explicity report the specific sensor task being addressed, the final performance metrics of the sensor or algorithm, and the architectural configuration of the sensor used. 
    \item \textbf{I5:} Must be focused on sun sensor calibration algorithms. Published works on the physical design and architecture of sun sensors were not reviewed.
\end{itemize}

I1 was chosen to ensure that the selected papers are relevant to modern sensors and the CubeSat standard. I3-I4 require that the methods are sufficiently reported to enable a thorough systematic mapping study. I5 ensures topical relevance of the review. Papers were required to satisfy all five inclusion criteria to be selected for the study.

\subsection{Compiled studies dataset}

The literature search and selection process yielded a dataset of 128 papers. These papers are categorized by model representation in Table \ref{tab:mrlit} and by feature extraction method in Table \ref{tab:felit}. The complete sun sensor calibration literature dataset compiled for this study is publicly available on the open research repository Zenodo \cite{Herman2025a}. Additionally, an interactive data visualization of the dataset is accessible via the Tableau Public platform \cite{Herman2025b}. We reference links to the associated data repositories in Appendix \ref{sec:appendixa}. 

\begin{figure}[h!]
	\centering
        \fbox{\includegraphics[width=\colfigwidth]{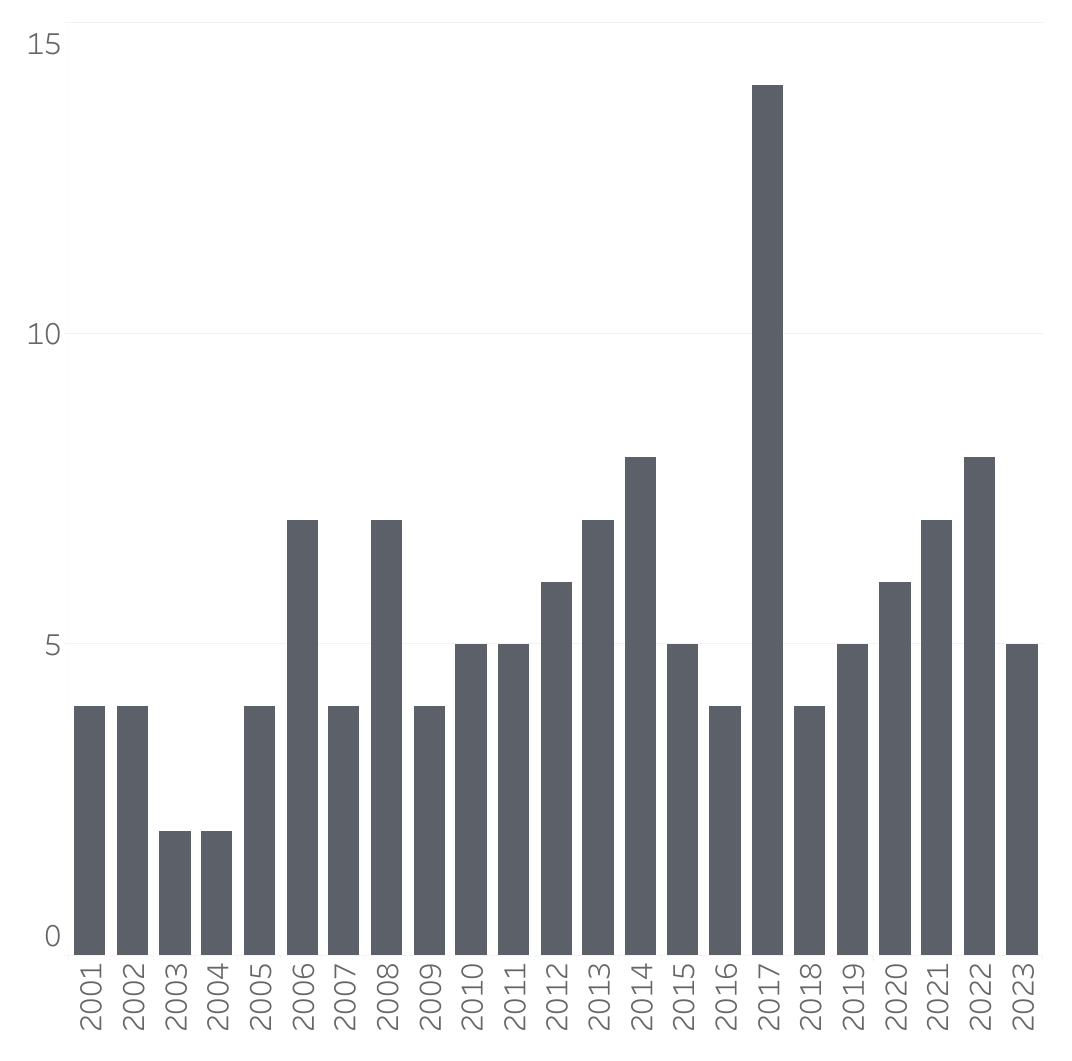}}
	\caption{Number\ of publications per year. \cite{Xie2008,Saleem2017,He2005,Coutinho2022,Massari2004,Wei2017,Post2013,deBoer2013,Rufino2005,Zhuo2020,Xie2014,Ortega2010,Farian2018,Delgado2010,Jing2016,Chang2006,Xing2008,deBoom2003,You2011,Lu2017a,Chum2003,Wang2019,Bardallo2017b,Kim2021,Saleem2019,Delgado2012b,Arshad2018,Enright2006b,Enright2007a,Xie2011,Xie2012,Fan2015,Wei2013,Chen2007,Buonocore2005,Bolshakov2020,Boslooper2017,Farian2022,Allgeier2009,TrebiOllennu2001,Rodrigues2006,Maji2015,He2013,Enright2008a,Bahrampoor2018,Rufino2004a,Rhee2012,ONeill2020,Maqsood2010,Antonello2018,Chang2008,Kim2005,Chen2006,Rufino2012,KeQiang2020,Li2012,Fan2013,Delgado2011,Yousefian2017,Mobasser2001,Chang2007a,Pelemeshko2020,Zhang2021,Kohli2019,Abhilash2014,Diriker2021,Delgado2012a,Wei2011,Rufino2017,Ali2011,Xie2013,Sozen2021,Mobasser2003,Liebe2001,Ortega2009,Fan2016,deBoom2006,Liebe2002,Xie2010,Farian2015,deBoer2017,Enright2007b,Rufino2009b,Wei2014,Miao2017,Godard2006,Faizullin2017,Alvi2014,Enright2008c,Richie2015,Wang2015,Zhang2022a,Rufino2008,Rufino2009a,Delgado2013,Liebe2004,Nurgizat2021,Frezza2022,Bardallo2017a,Lin2016,Lu2017b,Enright2008b,Hales2002,Strietzel2002,Tsuno2019,Kondo2024,Liu2016,Minor2010,Barnes2014,Merchan2021,Leijtens2020,Sun2023,Hermoso2022a,Merchan2023,Hermoso2022b,Zhang2022b,Zhang2023,Pentke2022,Soken2023,OKeefe2017,Radan2019,Hermoso2021,Fan2017,Wu2001,Wu2002,Springmann2014,Duan2006,Mafi2023,Pita2014,Herman2025a}}
	\label{fig:pubstotal}
\end{figure}

To assess research interest in the field, it is helpful to examine the chronological distribution of publications. Figure \ref{fig:pubstotal} shows the number of publications per year, revealing a steady growth in interest in sun sensor calibration since 2001. Notably, publication activity peaked in 2017 and again in 2022, with a slight decline observed in 2023. For a more detailed breakdown of publication frequency by academic journals and conference venues, this information is provided in the accompanying Zenodo and Tableau Public repositories. 

\subsection{Systematic mapping results}

To analyze the data and address the research questions, a calibration taxonomy was developed. The resulting systematic mapping is presented in Table \ref{tab:sysmap}. The main categories include model representation, sensor goals, feature extraction, architecture, and performance. \textit{Model representation} encompasses both the general model type and the specific modeling approach. \textit{Sensor goals} capture the targeted sensor tasks and associated error types. \textit{Feature extraction} lists the techniques used to process sensor data. \textit{Architecture} includes details on the sensor’s detector and mask design, while \textit{performance} categorizes reported sensor accuracy into defined bins. Each model representation and feature extraction method listed in Table \ref{tab:sysmap} is cross-referenced with the section where it is discussed in further detail.

\begingroup
\renewcommand{\arraystretch}{1.5} 
\begin{table*}[htb]
\caption{Overview of mapping sun sensor calibration.}
\label{tab:sysmap}
\scriptsize
\centering
\newcount\numcols
\numcols=11
\newlength\q
\setlength\q{\dimexpr (\textwidth - 2\tabcolsep*\numcols)/\numcols - 1pt\relax}
\noindent\begin{tabular}{>{\rowmac}p{\q}>{\rowmac}p{\q}>{\rowmac}p{\q}>{\rowmac}p{\q}>{\rowmac}p{\q}>{\rowmac}p{\q}>{\rowmac}p{\q}>{\rowmac}p{\q}>{\rowmac}p{\q}>{\rowmac}p{\q}>{\rowmac}p{\q}<{\clearrow}}
\hline
\multicolumn{4}{l}{\setrow{\bfseries}Model Representation} & \multicolumn{2}{l}{\setrow{\bfseries}Sensor Goals} & \multicolumn{2}{l}{\setrow{\bfseries}Feature Extraction} & \multicolumn{2}{l}{\setrow{\bfseries}Architecture} & \multicolumn{1}{l}{\setrow{\bfseries}Performance}\\
\cmidrule(lr){1-4}\cmidrule(lr){5-6}\cmidrule(lr){7-8}\cmidrule(lr){9-10}\cmidrule(lr){11-11}
\setrow{\bfseries}Category & \multicolumn{3}{c}{\setrow{\bfseries}Approaches} & Task & Error & \multicolumn{2}{c}{\setrow{\bfseries}Techniques} & Detector & Masks & Accuracy\\
\Xhline{1.25pt}  
\hyperref[model:lut]{LUT} & \hyperref[model:lut]{LUT} & \hyperref[model:mslit]{Multi-Slit} & \hyperref[model:assem]{ACM} & Accuracy & Alignment & \hyperref[fe:direct]{Direct} & \hyperref[fe:escm]{ESCM} & PD & Single-Ap. & CSS \\
\hyperref[model:npr]{Non-Physical} & \hyperref[model:linear]{Linear} & \hyperref[model:vslit]{V-Slit} & \hyperref[model:optic]{OCM} & Cost & Manufa-cturing & \hyperref[fe:vb]{Voltage balance} & \hyperref[fe:bscm]{BSCM} & QPD & Multi-Ap. & FSS \\
\hyperref[model:geo]{Geometric} & \hyperref[model:poly]{Poly-nomial} & \hyperref[model:nslit]{N-Slit} & \hyperref[model:mnc]{ECM/ Noise} & Power & Optical & \hyperref[fe:peak]{PD} & \hyperref[fe:ht]{HT} & PDA & Slit & VFSS \\
\hyperref[model:pi]{Physics-Informed} & \hyperref[model:trig]{Trig} & \hyperref[model:cam]{Camera} & \hyperref[model:nonlin]{ECM/ Angular loss} & FOV & Electrical & \hyperref[fe:ppe]{PPE} & \hyperref[fe:fmms]{FMMS} & PSD & L-Slit & UFSS \\
\hyperref[model:multiplex]{Multiplex} & \hyperref[model:sigmoid]{Sigmoid} & \hyperref[model:base]{Basic} & \hyperref[model:albedo]{ICM/ Albedo} & Latency & Inter-ference & \hyperref[fe:bcm]{BCM} & \hyperref[fe:ifm]{IFM} & CCD & N-Slit & \\
\hyperref[model:nn]{Neural network} & \hyperref[model:fourier]{Fourier} & \hyperref[model:solarpanel]{Solar panel} & \hyperref[model:shad]{ICM/ Shadow} & Volume & Environ-mental & \hyperref[fe:bctm]{BCTM} & \hyperref[fe:tm]{TM} & CMOS & V-Slit & \\
None & \hyperref[model:spm]{SPM} & \hyperref[model:pyra]{Pyramid} & \hyperref[model:periodic]{Periodic} & Precision & None & \hyperref[fe:mcam]{MCAM} & \hyperref[fe:feic]{FEIC} & Event & Multi-Slit & \\
& \hyperref[model:lsq]{LSQ} & \hyperref[model:pano]{Pano-ramic} & \hyperref[model:code]{Coded} & & & \hyperref[fe:dbcm]{DBCM} & \hyperref[fe:linearphase]{Linear-phase} & OI & Encoded & \\
& \hyperref[model:qpd]{QPD} & \hyperref[model:solar]{SECM/ Solar} & \hyperref[model:ann]{ANN} & & & \hyperref[fe:mtacm]{MT-ACM} & \hyperref[fe:eigena]{Eigen-analysis} & LTF & Lens & \\
& \hyperref[model:slit]{Slit} & \hyperref[model:temp]{SECM/ Temper-ature} & \hyperref[model:dnn]{DNN} & & & \hyperref[fe:pm]{PM} & None & PVA & Maskless & \\
\Xhline{1.25pt}
\end{tabular}
\begin{flushleft}
Abbreviations: A/D, analog/digital; ACM, assembly compensation model; ANN, artificial neural network; BCM, Basic Centroiding Method; BCTM, Basic Centroiding Thresholding Method; BSCM, Black Sun Centroiding Method; CCD, charge-coupled device; CMOS, complementary metal-oxide semiconductor; CSS, coarse sun sensor; DBCM, Double Balance Centroiding Method; DNN, deep neural network; ECM, electrical compensation model; ESCM, Event Sensor Centroiding Method; FEIC, Feature extraction image correlation; FMMS, Fast Multi-Point MEANSHIFT; FOV, field of view; FSS, fine sun sensor; HT, Hough transform; ICM, interference compensation model; IFM, Image filtering method; LPD, linear photodiode; LTF, light-to-frequency; LSQ, least squares; LUT, look up table; MCAM, Multiple Centroid Averaging Method; MT-ACM, Multiple-Threshold Averaging Centroiding Method; OCM, optical compensation model; OI, opto-isolator; PD, Peak Detection; PDA, photodiode array; PM, PixelMax; PPE, Peak Position Estimate; PSD, position sensitive device; PVA, photovoltaic array; QPD, quadrant photodiode; SECM, space environment compensation model; SPM, standard projection model; TM, Template method; UFSS, ultra fine sun sensor; VFSS, very fine sun sensor.
\end{flushleft}
\end{table*}
\endgroup

\section{Sensor attribute analysis} \label{sec:attribanalysis}

This section analyzes the systematic mapping using Sankey diagrams to visualize the relationships between key sensor attributes. By tracing the flow between attributes, the diagrams highlight research trends for newcomers and provide deeper guidance for practitioners. Specifically, each primary qualitative sensor attribute is illustrated using three-level Sankey diagrams, including separate diagrams for error types and performance levels. Following this, a cross-analysis is conducted across five sensor attributes to further explore interdependencies and patterns in the literature.

The compiled data was processed and visualized using Tableau software to generate Sankey diagrams. Additional analyses are available in the Tableau Public visualization \cite{Herman2025b}, which includes a task-oriented Sankey diagram, a traceable cross-analysis Sankey, an aggregate version of the cross-analysis Sankey, and a parallel coordinates plot for quantitative performance evaluation. The following subsections present and discuss the results of each primary attribute diagram in detail.

\begin{figure}[t]
	\centering
		\includegraphics[width=\colfigwidth]{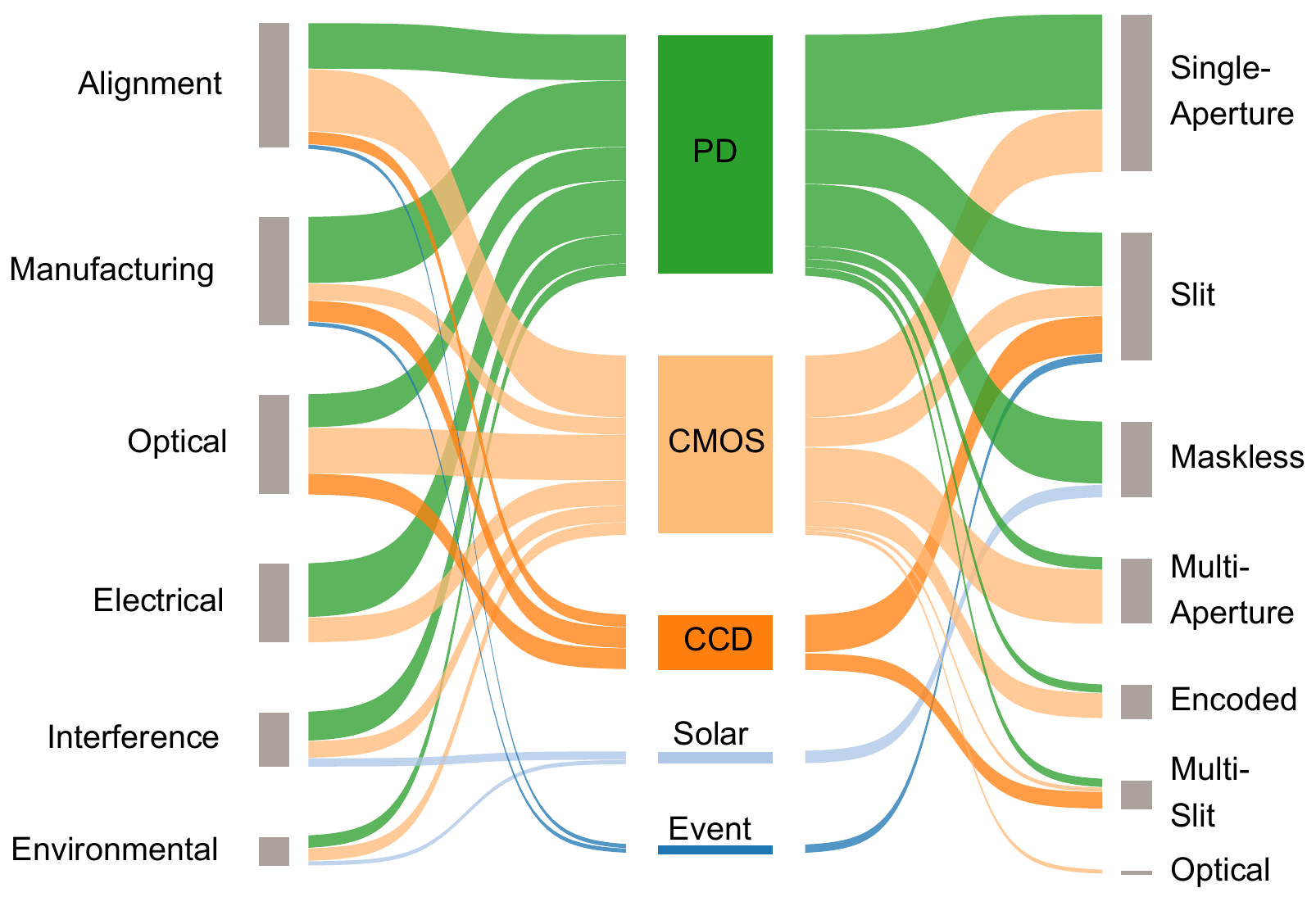}
	\caption{Error Sankey. \cite{Xie2008,Saleem2017,He2005,Coutinho2022,Massari2004,Wei2017,Post2013,deBoer2013,Rufino2005,Zhuo2020,Xie2014,Ortega2010,Farian2018,Delgado2010,Jing2016,Chang2006,Xing2008,deBoom2003,You2011,Lu2017a,Chum2003,Wang2019,Bardallo2017b,Kim2021,Saleem2019,Delgado2012b,Arshad2018,Enright2006b,Enright2007a,Xie2011,Xie2012,Fan2015,Wei2013,Chen2007,Buonocore2005,Bolshakov2020,Boslooper2017,Farian2022,Allgeier2009,TrebiOllennu2001,Rodrigues2006,Maji2015,He2013,Enright2008a,Bahrampoor2018,Rufino2004a,Rhee2012,ONeill2020,Maqsood2010,Antonello2018,Chang2008,Kim2005,Chen2006,Rufino2012,KeQiang2020,Li2012,Fan2013,Delgado2011,Yousefian2017,Mobasser2001,Chang2007a,Pelemeshko2020,Zhang2021,Kohli2019,Abhilash2014,Diriker2021,Delgado2012a,Wei2011,Rufino2017,Ali2011,Xie2013,Sozen2021,Mobasser2003,Liebe2001,Ortega2009,Fan2016,deBoom2006,Liebe2002,Xie2010,Farian2015,deBoer2017,Enright2007b,Rufino2009b,Wei2014,Miao2017,Godard2006,Faizullin2017,Alvi2014,Enright2008c,Richie2015,Wang2015,Zhang2022a,Rufino2008,Rufino2009a,Delgado2013,Liebe2004,Nurgizat2021,Frezza2022,Bardallo2017a,Lin2016,Lu2017b,Enright2008b,Hales2002,Strietzel2002,Tsuno2019,Kondo2024,Liu2016,Minor2010,Barnes2014,Merchan2021,Leijtens2020,Sun2023,Hermoso2022a,Merchan2023,Hermoso2022b,Zhang2022b,Zhang2023,Pentke2022,Soken2023,OKeefe2017,Radan2019,Hermoso2021,Fan2017,Wu2001,Wu2002,Springmann2014,Duan2006,Mafi2023,Pita2014,Herman2025a}}
	\label{fig:errorsankey}
\end{figure}

\textbf{Error analysis.} Figure \ref{fig:errorsankey} illustrates the relationship between error sources and sensor architectures. Specifically, it maps the associated sensor detectors and their respective masks to the corresponding sources of error. The purpose of this diagram is to guide practitioners in identifying and correcting expected errors for each sensor architecture. To that end, we trace each step of the error analysis and summarize the key findings below. The most commonly addressed error sources in the literature are alignment, manufacturing, and optical errors, while interference and environmental errors are the least frequently considered.

Alignment error correction is most frequently associated with CMOS detectors paired with multi-aperture mask architectures in the literature. This trend likely stems from the heightened accuracy requirements of this configuration and its susceptibility to minor misalignments introduced during assembly. Manufacturing errors, on the other hand, are most commonly addressed in systems using photodiode detectors with single-aperture mask designs—likely because these simpler, low-cost setups are more susceptible to manufacturing and tolerance-related faults. 

Optical error correction tends to be most associated with CMOS detectors and single or multi-aperture masks. The literature suggests that this correlation is related to diffraction effects from small apertures that must be limited via design optimization or accounted for in the calibration model. Electrical error mitigation strategies are primarily focused on photodiode detectors with both single aperture and maskless designs. We find that these errors stem from Fresnel reflection losses on the photodiode interface that must be corrected via calibration modeling, especially at large angles of incidence.

Interference errors are primarily addressed in systems utilizing photodiode and solar cell detectors, across both masked and maskless architectures. These errors are largely attributed to Earth albedo interference affecting analog sensors, while digital sensors tend to be relatively immune \cite{Xie2013}. Environmental error mitigation, meanwhile, appears distributed across photodiode, CMOS, and solar cell detectors, with no evident correlation to mask design. Coverage of environmental error sources in the literature is limited and mainly focuses on temperature effects, which can influence the performance of most detector types.

\begin{figure}[t]
	\centering
		\includegraphics[width=\colfigwidth]{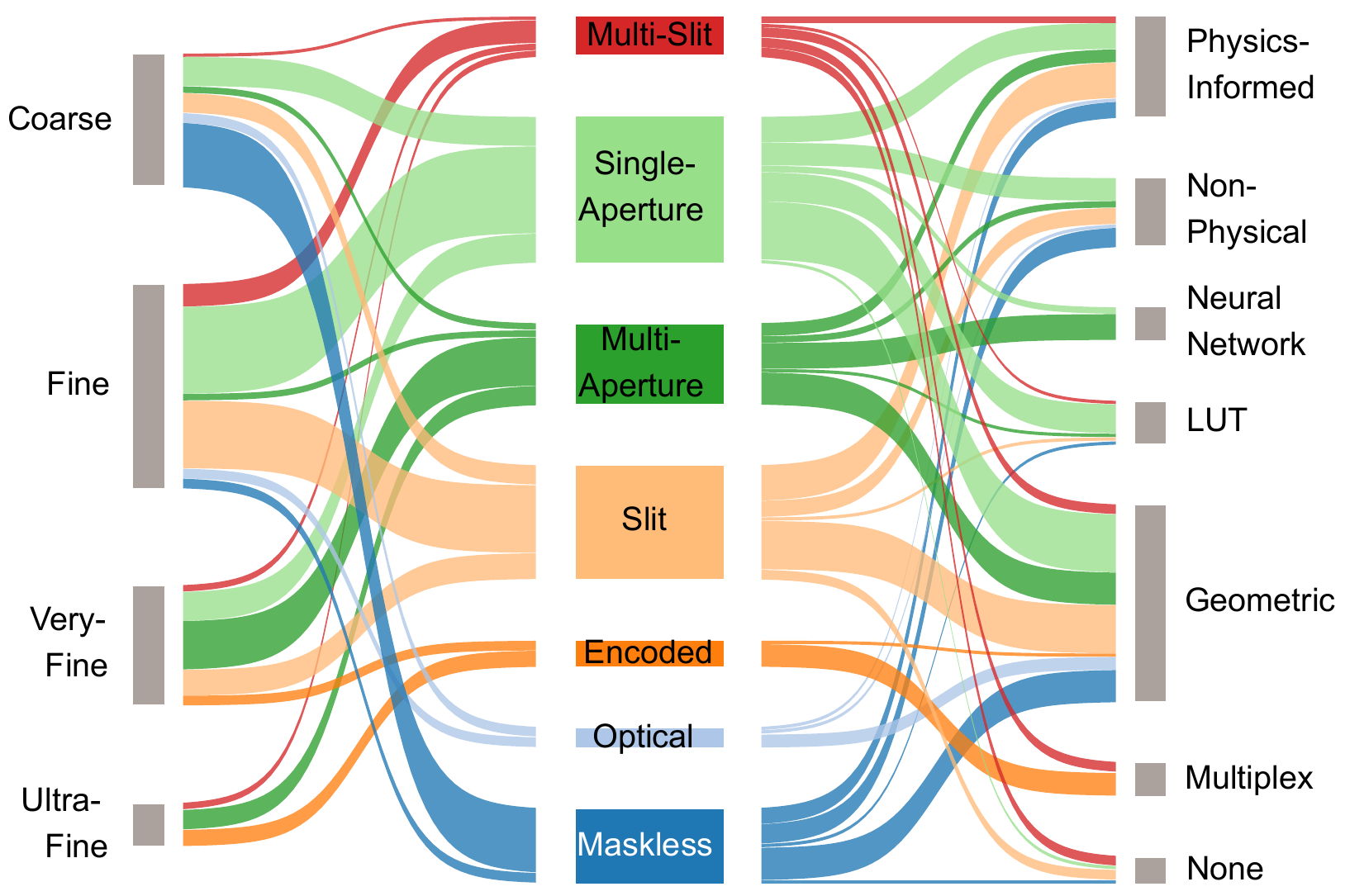}
	\caption{Performance Sankey. \cite{Xie2008,Saleem2017,He2005,Coutinho2022,Massari2004,Wei2017,Post2013,deBoer2013,Rufino2005,Zhuo2020,Xie2014,Ortega2010,Farian2018,Delgado2010,Jing2016,Chang2006,Xing2008,deBoom2003,You2011,Lu2017a,Chum2003,Wang2019,Bardallo2017b,Kim2021,Saleem2019,Delgado2012b,Arshad2018,Enright2006b,Enright2007a,Xie2011,Xie2012,Fan2015,Wei2013,Chen2007,Buonocore2005,Bolshakov2020,Boslooper2017,Farian2022,Allgeier2009,TrebiOllennu2001,Rodrigues2006,Maji2015,He2013,Enright2008a,Bahrampoor2018,Rufino2004a,Rhee2012,ONeill2020,Maqsood2010,Antonello2018,Chang2008,Kim2005,Chen2006,Rufino2012,KeQiang2020,Li2012,Fan2013,Delgado2011,Yousefian2017,Mobasser2001,Chang2007a,Pelemeshko2020,Zhang2021,Kohli2019,Abhilash2014,Diriker2021,Delgado2012a,Wei2011,Rufino2017,Ali2011,Xie2013,Sozen2021,Mobasser2003,Liebe2001,Ortega2009,Fan2016,deBoom2006,Liebe2002,Xie2010,Farian2015,deBoer2017,Enright2007b,Rufino2009b,Wei2014,Miao2017,Godard2006,Faizullin2017,Alvi2014,Enright2008c,Richie2015,Wang2015,Zhang2022a,Rufino2008,Rufino2009a,Delgado2013,Liebe2004,Nurgizat2021,Frezza2022,Bardallo2017a,Lin2016,Lu2017b,Enright2008b,Hales2002,Strietzel2002,Tsuno2019,Kondo2024,Liu2016,Minor2010,Barnes2014,Merchan2021,Leijtens2020,Sun2023,Hermoso2022a,Merchan2023,Hermoso2022b,Zhang2022b,Zhang2023,Pentke2022,Soken2023,OKeefe2017,Radan2019,Hermoso2021,Fan2017,Wu2001,Wu2002,Springmann2014,Duan2006,Mafi2023,Pita2014,Herman2025a}}
	\label{fig:perfsankey}
\end{figure}

\textbf{Performance analysis.} Figure \ref{fig:perfsankey} illustrates the relationship between qualitative sensor performance and relevant system attributes. Sensor performance is categorized into four accuracy bins, which are then traced through the associated sensor mask types and model representations found in the literature. The purpose of this Sankey diagram is to help practitioners identify optimal configurations based on specific sensor accuracy requirements. Most sensors reported in the literature fall within the fine accuracy bin, followed by those in the coarse bin. Sensors achieving ultra-fine accuracy are the least commonly implemented and most difficult to produce.

A coarse sun sensor is defined as having an accuracy of 0.5° or worse, typically achieving around 1° of accuracy \cite{Delgado2012b}. In the literature, coarse sensors are most commonly associated with maskless designs, followed by single-aperture masks. These configurations generally utilize simpler representations—such as geometric models, lookup tables (LUTs), or non-physical models—due to their reliance on analog signal mapping. In contrast, a fine sun sensor is characterized by an accuracy better than 0.5°, with some achieving precision as high as 0.01° (approximately one arcminute) \cite{Delgado2012b}. Fine sensors typically employ single-aperture or slit masks, as they often rely on the detection and mapping of a single centroid feature. This approach generally requires geometric or physics-informed models to accurately represent the feature space.

A very-fine sun sensor is defined as having an accuracy better than one arcminute but not reaching arcsecond-level precision. These sensors typically employ multi-aperture and encoded masks to capture more complex feature patterns simultaneously. As a result, more sophisticated models are used in the literature—such as neural networks for multi-aperture systems and multiplexing techniques for encoded masks.
At the highest level of precision, ultra-fine sun sensors achieve accuracies equal to or better than arcsecond level \cite{Zhang2021}. These sensors predominantly use encoded and multi-aperture masks, which enable the capture of a richer feature space. Encoded mask configurations are often paired with advanced multiplexing models, such as coding rule-based approaches, to accurately map the feature space.

\begin{figure*}[ht]
	\centering
		\includegraphics[width=\textwidth]{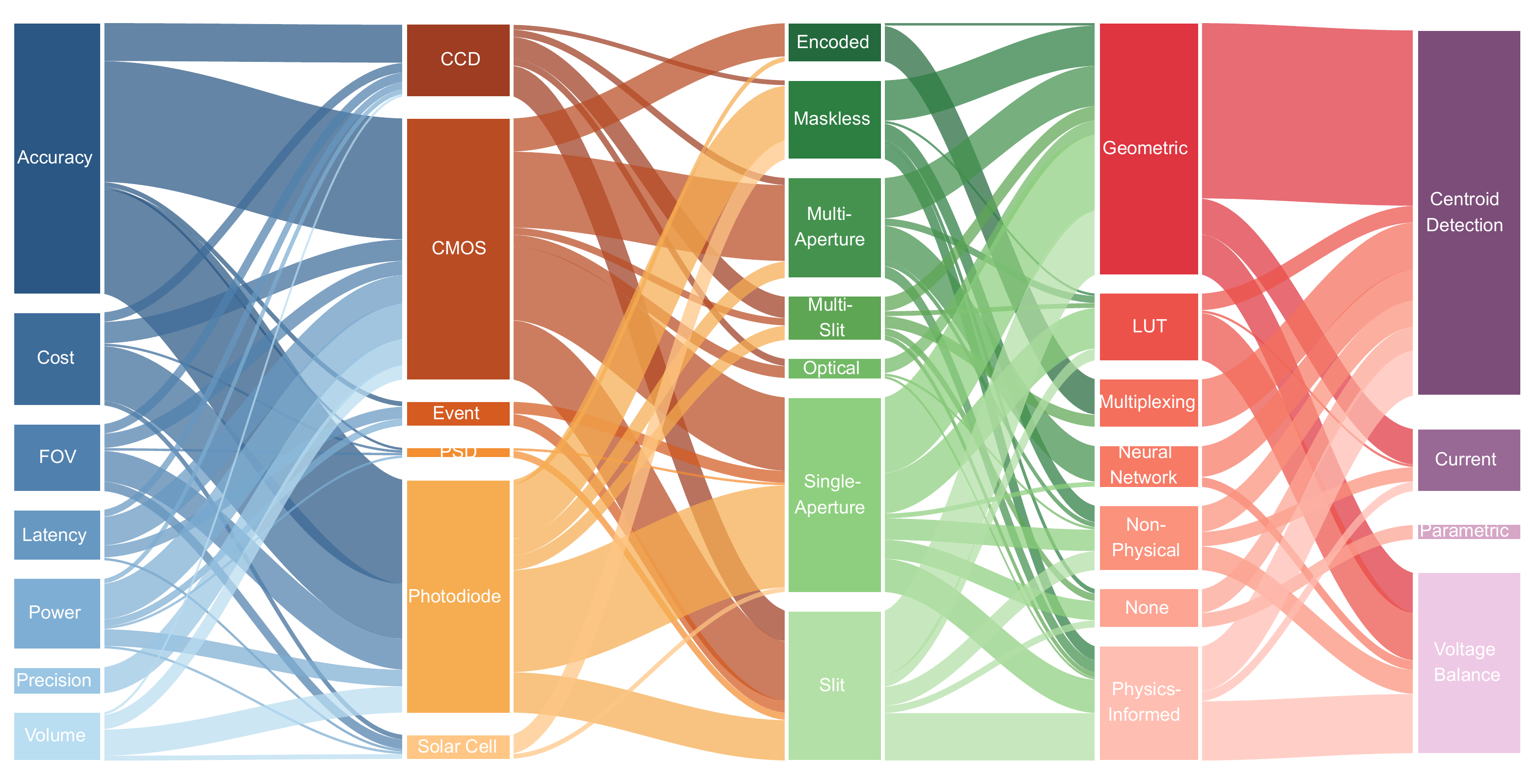}
	\caption{Cross-Analysis Sankey. \cite{Xie2008,Saleem2017,He2005,Coutinho2022,Massari2004,Wei2017,Post2013,deBoer2013,Rufino2005,Zhuo2020,Xie2014,Ortega2010,Farian2018,Delgado2010,Jing2016,Chang2006,Xing2008,deBoom2003,You2011,Lu2017a,Chum2003,Wang2019,Bardallo2017b,Kim2021,Saleem2019,Delgado2012b,Arshad2018,Enright2006b,Enright2007a,Xie2011,Xie2012,Fan2015,Wei2013,Chen2007,Buonocore2005,Bolshakov2020,Boslooper2017,Farian2022,Allgeier2009,TrebiOllennu2001,Rodrigues2006,Maji2015,He2013,Enright2008a,Bahrampoor2018,Rufino2004a,Rhee2012,ONeill2020,Maqsood2010,Antonello2018,Chang2008,Kim2005,Chen2006,Rufino2012,KeQiang2020,Li2012,Fan2013,Delgado2011,Yousefian2017,Mobasser2001,Chang2007a,Pelemeshko2020,Zhang2021,Kohli2019,Abhilash2014,Diriker2021,Delgado2012a,Wei2011,Rufino2017,Ali2011,Xie2013,Sozen2021,Mobasser2003,Liebe2001,Ortega2009,Fan2016,deBoom2006,Liebe2002,Xie2010,Farian2015,deBoer2017,Enright2007b,Rufino2009b,Wei2014,Miao2017,Godard2006,Faizullin2017,Alvi2014,Enright2008c,Richie2015,Wang2015,Zhang2022a,Rufino2008,Rufino2009a,Delgado2013,Liebe2004,Nurgizat2021,Frezza2022,Bardallo2017a,Lin2016,Lu2017b,Enright2008b,Hales2002,Strietzel2002,Tsuno2019,Kondo2024,Liu2016,Minor2010,Barnes2014,Merchan2021,Leijtens2020,Sun2023,Hermoso2022a,Merchan2023,Hermoso2022b,Zhang2022b,Zhang2023,Pentke2022,Soken2023,OKeefe2017,Radan2019,Hermoso2021,Fan2017,Wu2001,Wu2002,Springmann2014,Duan2006,Mafi2023,Pita2014,Herman2025a}}
	\label{fig:casankey}
\end{figure*}

\textbf{Attribute cross-analysis.} We conclude this section with a cross-analysis of key sensor attributes using a five-level Sankey diagram, shown in Figure \ref{fig:casankey}. The purpose of this figure is to assist both newcomers and experts in navigating the sun sensor selection process, using insights drawn from the surveyed literature and the decision flow outlined in Figure \ref{fig:decisionflow}. The five categories represented from left to right are: task, detector type, mask type, model representation, and feature extraction method.
Unsurprisingly, accuracy is the most demanded requirement for sun sensor algorithms. Following the decision flow for an accuracy requirement, the literature commonly recommends selecting a CMOS detector paired with either a multi-aperture or encoded mask. For a multi-aperture mask, geometric models or neural networks combined with centroiding algorithms are suggested. In contrast, for an encoded mask, a multiplexing model coupled with centroiding is the preferred approach.

When tracing the decision flow based on cost considerations, a strong trend emerges favoring photodiode detectors paired with either single-aperture masks or maskless architectures. For single-aperture masks, geometric, LUT, and physics-informed models combined with voltage balance algorithms are commonly recommended. In the case of maskless designs, geometric models with either direct or voltage balance methods are typically selected.
For FOV requirements, the decision flow often points toward the use of photodiode detectors or solar cells without a mask, or CMOS detectors with encoded masks. In maskless configurations, geometric models paired with current or voltage balance algorithms are frequently applied. In contrast, for encoded masks, multiplexing models combined with centroiding techniques are the preferred choice.
Sensor designs driven by latency requirements tend to favor event-based or CMOS detectors with single-aperture or slit masks. These configurations are commonly supported by geometric models coupled with centroiding algorithms.

When considering power constraints, the decision flow shows a strong preference for CMOS and event-based detectors paired with slit masks. In these configurations, geometric models combined with centroiding algorithms are the most commonly adopted design approach.
For precision-focused designs, the decision flow consistently favors CMOS detectors with multi-aperture masks. These setups are typically supported by geometric or neural network models in conjunction with centroiding techniques.
Finally, for designs driven by volume constraints, the flow trends toward the selection of photodiode or solar cell detectors without a mask. These configurations are most often paired with geometric models utilizing current or voltage-based feature extraction algorithms. \phantomsection\label{rqsum:rqsum1}

\begin{description}
\item[RQ 1 Summary] \textit{Algorithm selection is dependent on the downstream decision flow of sensor design requirements.} (See \hyperref[rq:rq1]{RQ 1})
\end{description} 

\section{Sun sensor model representations} \label{sec:ssmodrep}

Offline calibration refers to the process of tuning a sensor model within a controlled experimental environment. This process may involve a single static test or a series of calibration scenarios focused on characterizing uncertainties. For sun sensors, offline calibration typically represents the initial step in the overall calibration strategy, with online calibration applied throughout the sensor’s lifecycle to compensate for accuracy degradation over time. It is important to note that not all sun sensors require a fine-tuned calibration; this process is generally reserved for fine and ultra-fine sun sensors that demand high-performance state estimation. Nevertheless, most sun sensors use some form of calibration model representation for attitude estimation. 

\begin{figure*}[h]
	\centering
		\includegraphics[width=\textwidth]{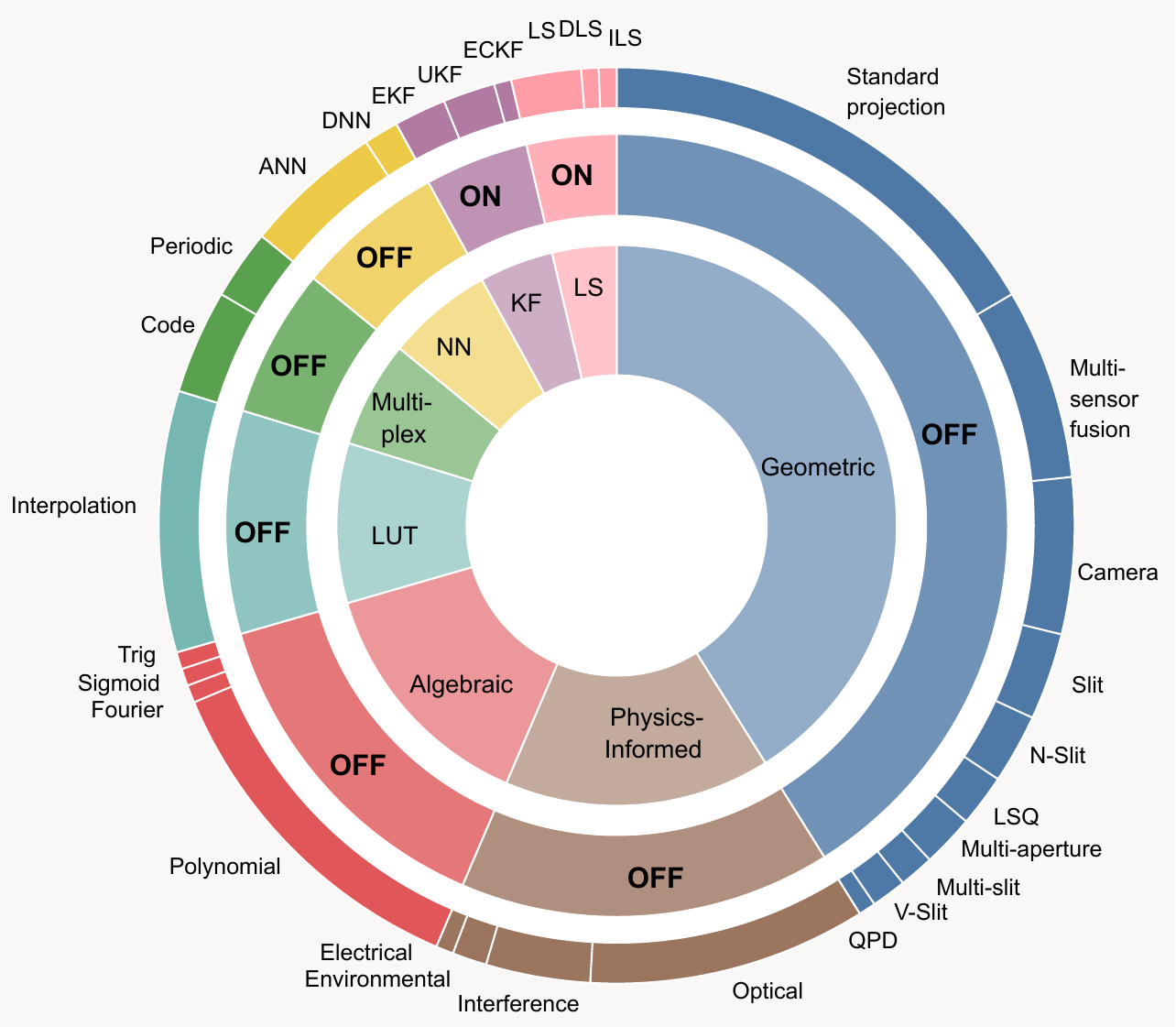}
	\caption{Model Representation Sunburst. \cite{Xie2008,Saleem2017,He2005,Coutinho2022,Massari2004,Wei2017,Post2013,deBoer2013,Rufino2005,Zhuo2020,Xie2014,Ortega2010,Farian2018,Delgado2010,Jing2016,Chang2006,Xing2008,deBoom2003,You2011,Lu2017a,Chum2003,Wang2019,Bardallo2017b,Kim2021,Saleem2019,Delgado2012b,Arshad2018,Enright2006b,Enright2007a,Xie2011,Xie2012,Fan2015,Wei2013,Chen2007,Buonocore2005,Bolshakov2020,Boslooper2017,Farian2022,Allgeier2009,TrebiOllennu2001,Rodrigues2006,Maji2015,He2013,Enright2008a,Bahrampoor2018,Rufino2004a,Rhee2012,ONeill2020,Maqsood2010,Antonello2018,Chang2008,Kim2005,Chen2006,Rufino2012,KeQiang2020,Li2012,Fan2013,Delgado2011,Yousefian2017,Mobasser2001,Chang2007a,Pelemeshko2020,Zhang2021,Kohli2019,Abhilash2014,Diriker2021,Delgado2012a,Wei2011,Rufino2017,Ali2011,Xie2013,Sozen2021,Mobasser2003,Liebe2001,Ortega2009,Fan2016,deBoom2006,Liebe2002,Xie2010,Farian2015,deBoer2017,Enright2007b,Rufino2009b,Wei2014,Miao2017,Godard2006,Faizullin2017,Alvi2014,Enright2008c,Richie2015,Wang2015,Zhang2022a,Rufino2008,Rufino2009a,Delgado2013,Liebe2004,Nurgizat2021,Frezza2022,Bardallo2017a,Lin2016,Lu2017b,Enright2008b,Hales2002,Strietzel2002,Tsuno2019,Kondo2024,Liu2016,Minor2010,Barnes2014,Merchan2021,Leijtens2020,Sun2023,Hermoso2022a,Merchan2023,Hermoso2022b,Zhang2022b,Zhang2023,Pentke2022,Soken2023,OKeefe2017,Radan2019,Hermoso2021,Fan2017,Wu2001,Wu2002,Springmann2014,Duan2006,Mafi2023,Pita2014,Herman2025a}}
	\label{fig:modelsunburst}
\end{figure*}

The primary offline calibration techniques for sun sensors include voltage balance, lookup tables (LUT), non-physical models, geometric models, physics-informed models, and neural networks. These model representation techniques are visualized in Figure \ref{fig:modelsunburst} using a sunburst diagram. Each concentric ring in the diagram represents a different level of categorization: the innermost ring shows general model types, the middle ring indicates whether the model is used for offline or online calibration, and the outermost ring identifies the specific model. As illustrated, geometric model representations are the most widely implemented in the literature. 

In this section, each calibration approach is discussed in detail, including relevant case studies, as well as the strengths and limitations of each technique. The supporting literature for these calibration methods is summarized in Table \ref{tab:mrlit} at the end of the section.

\subsection{Lookup table model}\label{model:lut}

The error table approach stores the sensor output response or the associated fitted coefficients in a lookup table for later interpolation. This calibration approach is the simplest method and also the fastest, however it has limited accuracy in practice. 
It is especially inefficient as the size of the calibration table becomes very large. However, the methodology can be useful when the sensor response is difficult to model with a typical geometric or physics-based representation due to complex error factors \cite{He2013}. As such, this calibration method is usually limited to coarse sun sensors.

The work by He et al. \cite{He2013} is reviewed as a case-study for the LUT model representation. In the following study, error compensation and calibration are modeled using a coefficient table of two-dimensional error with a 1$^{\circ}$ interval. The error curve is then fitted using a polynomial fitting from the table. The polynomial order used is of the 8th degree from -64$^{\circ}$ to 64$^{\circ}$ over the full FOV. A LUT model was selected due to the difficulty in otherwise modeling the complex nature of the errors, which include geometric offsets, manufacturing errors, and refraction.

The LUT needed further refinement to meet the accuracy goals, however increasing the size of the LUT would require a large number of data points. This would greatly increase the computational load of the algorithm and inhibit in-orbit operations. As such, the authors took the compensation values as the weighted sum of two neighboring polynomials \cite{He2013}. An overview of the LUT method body of literature is presented in Table \ref{tab:mrlit}.

\subsection{Non-physical model}\label{model:npr}

The non-physical model calibration process involves the fitting of algebraic or transcendental regression models to measured sensor observations. These techniques are simplified representations of the underlying behavior of the sensor that lacks physical interpretability of the system from the model parameters. In the sections below, a number of non-physical representations are introduced including the linear fit, polynomial, trigonometric, Fourier series, and sigmoid model representations. Furthermore, the associated model strengths are compared, and example implementations are presented.

\subsubsection*{Linear fit}\label{model:linear}

The implementation of linear-fit model representations is rare in sun sensor calibration since most of the sensor responses and errors are highly-complex and non-linear in nature. However, linear-fit regression models have been used in analog photodiode applications to some effect. In particular, the response curves of quadrant photodiode output signal ratios \cite{Faizullin2017} and the illumination curves of photodiodes exhibit linear behaviors within certain regimes \cite{Hermoso2022a}. The technique is limited in accuracy, however the speed of the algorithm is useful in low-latency critical on-board applications.

The work by Faizullin et al. \cite{Faizullin2017} is reviewed as a case-study for the linear fit model representation. In the following study, the response curve of a quadrant photodiode is fitted with a linear fit. The line was fitted over three FOVs, each with a larger coverage to the full FOV range. The highest accuracy was achieved over the smallest FOV range, as the photodiode response curve is non-linear at the edges of the FOV. It was found that the model has utility over a very limited FOV, while the full FOV fit had large angular error. An overview of the linear-fit method body of literature is presented in Table \ref{tab:mrlit}.

\subsubsection*{Polynomial}\label{model:poly}

The polynomial model is the most widely implemented and extensively studied approach for sun sensor calibration. This technique uses polynomial curve fitting to model the relationship between the measured sensor response and the incident angle of sunlight on each axis. The sensor response is typically represented by analog voltage balance, digital intensity centroid values, or direct angle measurements.
Key advantages of this algebraic modeling approach include its simplicity, compactness, and suitability for fast on-board processing. However, it has several limitations: the polynomial coefficients lack physical interpretability, higher-order polynomials can lead to overfitting, and the model struggles to accurately represent highly complex systems. In practice, increasing the polynomial order beyond the 11th provides little to no performance improvement for sun sensor calibration \cite{Chang2006}.

The study by \citet{You2011} is reviewed as a case study for the polynomial model representation. In this work, calibration is performed using a sixth-degree polynomial to model each rotation angle across the sensor’s field of view. Separate polynomial equations are derived for each of the two sensor axes, with the non-rotating axis held fixed during calibration. Model parameters are trained using turntable ground-truth data in conjunction with sensor measurements for each independent axis. Once the model is fitted, it defines the relationship between the sensor’s measured angle and the estimated angle \cite{You2011}. An overview of the polynomial method body of literature is presented in Table \ref{tab:mrlit}.

\subsubsection*{Trigonometric}\label{model:trig}

The trigonometric model technique uses trigonometric functions to fit the sensor's scaled measurement data. The curve-fit sets can be combined for an averaged curve fit or the functions can be incorporated into a direct relationship of the scaled measurement. This approach is commonly used for fitting to analog sensor response curves, where high accuracy is less of a requirement. One limitation of using trigonometric model representations is that they require a quadrant check when solving for the inverse functions to resolve ambiguity \cite{Richie2015}.

The work by Richie et al. \cite{Richie2015} is reviewed as a case-study for the trigonometric model representation. In the following study, two trigonometric techniques are presented: an averaged combined panel curve-fit method and a direct relation of the scaled measurement approach. The averaged combined panel curve-fit method uses trigonometric functions to fit each sensor's scaled measurements. A quadrant check is then used to resolve estimate ambiguity, thereby resulting in four estimates of the slew angle. The lit sensor sides are used to determine the best match to the slew angle estimate by merging measurements from opposing sensors into averaged sets \cite{Richie2015}.

The second method presented in the study is a direct relation of the scaled measurement approach. This method fits the measurement data to a trigonometric function of the slew angle that is mathematically similar to the behavior of the sensor-scaled curves. This allows for a more compact relation that functionally includes a self-contained quadrant check. It is found that the direct approach is more accurate than the averaged combined approach \cite{Richie2015}. An overview of the trigonometric method body of literature is presented in Table \ref{tab:mrlit}.

\subsubsection*{Fourier series}\label{model:fourier}

The Fourier series approach is a non-linear regression method that uses a sum of trigonometric functions to express a periodic function. The calibration error of sun sensors often exhibits periodic behavior, especially from the signal processing of periodically coded apertures. The largest advantage of using a Fourier series representation is to better capture trigonometric effects in the sensor measurement data. However, the method is limited by increased computational demands and offset errors \cite{Pentke2022}.

One such example, the work by Fan et al. \cite{Fan2013}, is reviewed as a case-study for the Fourier series model representation. In the following study, a periodically encoded sun sensor is modeled that contains both coarse-code and fine-code rows. The incident angle is estimated from the sum of coarse-code and fine-code outputs. Four fine-code rows ($F1-F4$) in Equation \ref{eq:fourier} exhibit the same pattern, however a phase difference exists between two adjacent rows.

\begin{equation} \label{eq:fourier}
\begin{split}
F1=\frac{a_{0}}{2}-\sum_{n=1}^{\infty}a_{n}\cos\left ( n\frac{2\pi}{\theta_{0}}\alpha \right )\\
F2=\frac{a_{0}}{2}-\sum_{n=1}^{\infty}a_{n}\sin\left ( n\frac{2\pi}{\theta_{0}}\alpha \right )\\
F3=\frac{a_{0}}{2}+\sum_{n=1}^{\infty}a_{n}\cos\left ( n\frac{2\pi}{\theta_{0}}\alpha \right )\\
F4=\frac{a_{0}}{2}+\sum_{n=1}^{\infty}a_{n}\sin\left ( n\frac{2\pi}{\theta_{0}}\alpha \right )
\end{split}
\end{equation}

For Equation \ref{eq:fourier}, $F$ is the output current of the fine-code rows, $a_{0}$ is the amplitude of the fine-code output current, $\theta_{0}$ is the period of the fine-code rows, and $\alpha$ is the incident angle of sunlight. The primary error contribution in the sensor is due to the processing error of the fine-code apertures. The ideal output current of the four fine-code rows does not match the actual output expressions, but rather the true output is a periodic function expressed by a Fourier series. Therefore, the fine-code algorithm error of the sensor can be estimated from the Fourier series modeled phase error of fine-code output current. An overview of the Fourier series method body of literature is presented in Table \ref{tab:mrlit}.

\subsubsection*{Sigmoid}\label{model:sigmoid}

The sigmoid model representation is a non-linear regression method that uses a sigmoid function to fit the sensor measurement data. The most commonly used form in sun sensor calibration is a modified logistic function. In particular, the response curves of analog photodiode calibration are well suited to fitting the shape of the sigmoid function \cite{Pentke2022}. 

The study by Pentke et al. \cite{Pentke2022} is reviewed as a case study for the sigmoid model representation. In this work, a modified sigmoid function is used to calibrate a quadrant photodiode array by correcting for nonlinear errors. The calibration model consists of a combination of a sigmoid function, first-order Fourier terms, and a constant offset. While the sigmoid function effectively captures the characteristic response curve of the photodiode, the approach introduces a higher computational load compared to polynomial-based methods \cite{Pentke2022}. A summary of the relevant literature on sigmoid-based calibration methods is provided in Table \ref{tab:mrlit}.

\subsection{Geometric model}\label{model:geo}

Geometric model calibration involves fitting model representations whose parameters are physically meaningful and reflect the actual geometry of the sensor within the inverse function. These models range from general incident angle projection models to more complex, architecture-specific formulations that account for errors unique to a given sensor design. The following sections introduce several geometric model types, including the standard projection model, the least squares (LSQ) model, and a variety of architecture-specific models tailored to different mask and sensor configurations. For each approach, we highlight any strengths and weaknesses and provide example formulations. 

\subsubsection*{Standard projection model}\label{model:spm}

The standard projection model is the most common geometric model representation used in sun sensor calibration. In fact, it is so common that it is also the most widely implemented modeling technique for sun sensors in general. The principle of operation behind the model relies on the accurate knowledge of the center of the projected incident light spot on the detector. From this information, a simple inverse relation between the sensor focal length, pixel size, and spot centroid location is obtained to estimate the incident angle \cite{deBoer2013}. The popularity of the approach mostly stems from the algorithmic simplicity, generalizability to a number of sensor configurations, and the straightforward physical interpretability. However, the model does have limited accuracy due to not accounting for geometric misalignment in the model parameters. 

The work by de Boer et al. \cite{deBoer2013} is reviewed as a case-study for the standard projection model representation. The model is implemented for a digital sun sensor with a pinhole aperture. The centroid of intensity of the projected spot is used to estimate the two-axis incident angle. The standard projection model is defined in Equation \ref{eq:stdproj}.

\begin{equation} \label{eq:stdproj}
    \begin{split}
        \alpha=\arctan\left ( Y\cdot \frac{s}{h} \right ) \\
        \beta=\arctan\left ( X\cdot \frac{s}{h} \right )
    \end{split}
\end{equation}

For Equation \ref{eq:stdproj}, ($\alpha$, $\beta$) are the angular estimates, ($X$, $Y$) are the projected spot centroid locations, $s$ is the pixel size, and $h$ is the focal length. In addition, the underlying geometry of this formulation is illustrated in Figure \ref{fig:operbw}. The geometric model enables accurate calibration without the need for additional correction through lookup tables (LUTs) or temperature compensation. However, the primary source of error in this idealized approach stems from aperture misalignment. A summary of the literature on the standard projection method is provided in Table \ref{tab:mrlit}. 

\subsubsection*{Least Squares (LSQ) geometric model}\label{model:lsq}

The Least Squares (LSQ) geometric model is a refined calibration approach based on the standard projection model \cite{Rufino2004a}. While the standard projection model provides a basic framework for sun vector estimation, it falls short in achieving high accuracy due to discrepancies between the projected feature centroid and the actual sun vector intersection on the detector. These discrepancies result from both systematic and random errors, including feature deformation under varying light incidence, alignment imperfections, and machining tolerances. The LSQ model was developed specifically to compensate for these limitations and improve calibration accuracy over the standard projection approach.

The study by de Rufino et al. \cite{Rufino2004a} is reviewed as a case study for the LSQ geometric model representation. This method was developed for an APS using a pinhole mask configuration. Its core principle involves applying least-squares optimization to refine the physical calibration parameters of the system. The primary source of error in the baseline model is attributed to inaccuracies in the geometric configuration parameters. In this case, the calibration focuses on adjusting the focal length and mask tilt. The LSQ geometric model is defined in Equation \ref{eq:lsq}.

\begin{equation} \label{eq:lsq}
  \begin{aligned}
    \alpha_{m} & =-\alpha_{0}+\tan^{-1}\Biggl[ \frac{1}{F+\Delta F}\\
    & \times \cos\left ( \tan^{-1}\frac{x_{zp}-x+(F+\Delta F)\tan\beta_{0}}{F+\Delta F} \right )\\
    & \times \left ( y_{zp}-y\frac{(F+\Delta F)\tan\alpha_{0}}{\cos\beta_{0}} \right ) \Biggr]\\
    \beta_{m}& =-\beta_{0}+\tan^{-1}\frac{x_{zp}-x+(F+\Delta F)\tan\beta_{0}}{F+\Delta F}
  \end{aligned}
\end{equation}

The model is expressed in terms of the spot center coordinates ($x_{zp},y_{zp}$) at zero boresight offset. The parameters $\alpha_{m}$ and $\beta_{m}$ represent the measured sun vector rotations. The LSQ optimized parameters are then estimated for calibration via the least-squares process. The three terms that are calculated include the focal length deviation $\Delta F$ and the two boresight offsets ($\alpha_{0}$, $\beta_{0}$). An overview of the LSQ model representation body of literature is presented in Table \ref{tab:mrlit}.

\subsubsection*{Architecture-specific model}\label{model:asm}

An architecture-specific model is a sub-type of geometric model that is developed for a specific sensor mask or detector configuration. The following methods are presented without a comparative discussion of their strengths, as each is specifically optimized for its intended system and excels within that context. In the sections that follow, several architecture-specific model representations are introduced, including an analog sensor gap compensation model, various slit-based approaches, a camera-based model, and multiple multi-sensor fusion techniques. For each method, the underlying model is described, and example implementations are provided.

\textbf{QPD model.}\label{model:qpd} Light that falls in between the gaps of a quadrant photodiode (QPD) detector can lead to error in the expected sun sensor response. As such, the error can be compensated for by characterizing the photodiode gap geometry in a model representation. Generally, this requires a mathematical model that subtracts the gap intensity that would have fallen on the detector from the idealized model output.

The work by Faizullin et al. \cite{Faizullin2017} is reviewed as a case-study for the gap compensation model representation. This work focuses on sun angle estimation using a quadrant photodiode with a pinhole mask configuration. The gap error is accounted for by modeling the expected signal lost due to the quadrant photodiode gap geometry. The gap error is modeled as a function of both the pinhole aperture diameter and the dimensions of the gap itself. To correct for this, the light spot center is estimated based on the balance of output signals from the photodiodes and the area of the gap between them, which is proportional to the lit regions within the quadrant detector space. The gap compensation model is expressed in Equation \ref{eq:gapcomp} in the form of a gap corrected photodiode signal balance \cite{Faizullin2017}.

\begin{equation}\label{eq:gapcomp}
G_{i}=\frac{S_{G_{i}}}{k_{G}}\frac{A+B+C+D}{S_{A}+S_{B}+S_{C}+S_{D}}
\end{equation}

In the above equation, the output signal balance is $G_{i}$, the gap area between the photodiodes is $S_{G_{i}}$, the calibration coefficient is $k_{G}$, the areas of the quadrant photodiode covered by a light spot is ($S_{A},S_{B},S_{C},S_{D}$), and the  output signals from the photodiodes are ($A,B,C,D$). An overview of the gap compensation body of literature is presented in Table \ref{tab:mrlit}.

\textbf{Slit model.}\label{model:slit} The Slit model representation is an\\ architecture-specific method that accounts for the geometric errors embedded in a system with a slit mask and digital sensor configuration. The focus of the model is on correcting installation and manufacturing errors within the system. Specifically, the errors accounted for in the model include: the offset between the slit mask and linear array plane, the angular error between the slit mask and x-axis of the linear array, and the error derived from the slit mask non-perpendicularity with linear array \cite{KeQiang2020}.

The work by Ke-Qiang et al. \cite{KeQiang2020} is reviewed as a case-study for the Slit model representation. The Slit model is presented below in Equation \ref{eq:slit}.

\begin{align}\label{eq:slit}
    \begin{split}
        \tan{\alpha}'=\frac{{f}'}{f}\Biggl[ \tan\alpha+\tan\beta\cdot \tan\Big(\\
        \arctan\left ( \frac{\tan\alpha\tan\delta}{1-\tan\beta\tan\delta} \right )-\theta   \Big) \Biggr]
    \end{split}
\end{align}

For Equation \ref{eq:slit}, the corrected angular estimate is $\alpha'$, the angular measurements are ($\alpha$, $\beta$), $f$ is the theoretical focal length, $f'$ is the actual focal length, the 
installation deflection angular error is $\theta$, and the installation inclination angular error is $\delta$. An overview of the Slit model representation body of literature is presented in Table \ref{tab:mrlit}.

\textbf{Multi-slit model.}\label{model:mslit} The Multi-slit model representation is an architecture-specific approach designed to capture the periodic behavior of the mask geometry in systems using a multiple-slit mask combined with a linear photodiode array. This design enhances error reduction by detecting multiple features simultaneously and is particularly effective at low angles of incidence \cite{Bolshakov2020}.
The work by Bolshakov et al. \cite{Bolshakov2020} is reviewed as a case-study for the Multi-slit model representation. In this study a 5-slit aperture mask is used in which the center slit is the primary feature mapping source, while the other four apertures are used to reduce measurement error in the system. The Multi-slit model is presented in Equations \ref{eq:mslit1} \& \ref{eq:mslit2}.

\begin{equation}\label{eq:mslit1}
\theta_{i}=\tan^{-1}\left ( \frac{X_{i}-d_{i}}{h} \right ) \textup{ for } i=2,4,6
\end{equation}

\begin{equation}\label{eq:mslit2}
\theta_{i}=\tan^{-1}\left ( \frac{X_{i}-d_{i}}{h+t} \right ) \textup{ for } i=1,3,5
\end{equation}

In the above equations, the mask thickness is denoted as $t$, the focal length is $h$, the distance from the mask edge to the aperture is $X_{i}$, the location of the periodic pixel intensity shift is $d_{i}$, which shifts from $0\to 1$ for odd indices and $1\to0$ for even indices. An overview of the Multi-slit model representation body of literature is presented in Table \ref{tab:mrlit}.

\textbf{V-slit model.}\label{model:vslit} The V-slit model representation is an architecture-specific method that predicts the spot intersections of the mask geometry of a V-slit type mask with a linear digital detector. The V-slit type mask configuration is composed of a vertical slit and a tilted slit with a 45$^{\circ}$ angle between. The two-axis incident angle is estimated by processing the two light spots which are projected onto the detector array. The focus of the compensation model is to correct for the two primary error sources for a V-slit mask configuration, which include structural and refraction errors.

The studies by Fan et al. \cite{Fan2015, Fan2016} are reviewed as a case study for the V-slit model representation. The proposed mathematical model includes an error compensation framework based on eight intrinsic calibration parameters required for accurate sun angle estimation. Of these, six parameters account for structural errors, while the remaining two address refraction effects. During operation, the sun sensor uses a two-axis angle model to compute the estimated sun angle \cite{Fan2015, Fan2016}. The V-slit model is defined in Equations \ref{eq:vslit1} and \ref{eq:vslit2}.

\begin{flalign}\label{eq:vslit1}
\tan\alpha=\frac{a_{11}(T_{1}+y\tan\delta-X_{2})+a_{21}T_{2}+a_{31}T_{3}}{a_{13}(T_{1}+y\tan\delta-X_{2})+a_{23}T_{2}+a_{33}T_{3}}&&
\end{flalign}

\begin{flalign}\label{eq:vslit2}
\tan\beta =\myfrac{\begin{matrix*}[l]
y\tan\delta+(a_{11}-a_{12}\tan\delta-a_{13}\tan\alpha)T_{1} &\\
+(a_{21}-a_{22}\tan\delta-a_{23}\tan\alpha)T_{2} &\\
+(a_{31}-a_{32}\tan\delta-a_{33}\tan\alpha)T_{3} &\\
+(a_{11}-a_{12}\tan\delta-a_{13}\tan\alpha) &\\
\times (y\tan\delta-X_{1})
\end{matrix*}}
{\begin{matrix*}[l]
\bigl[ a_{13}(y\tan\delta-X_{1})+a_{13}T_{1} &\\
+a_{23}T_{2}+a_{33}T_{3} \bigr]\tan\delta
\end{matrix*}}&&
\end{flalign}

The error rotation matrix $\mathbf{R}$ consists of the calibration parameters $a_{11}$ through $a_{33}$. Additionally, the error displacement vector $\mathbf{T}$ is expressed as $[T_{1}, T_{2}, T_{3}]$. The angle $\delta$ represents the angle between the vertical axis and the tilted slit configuration. The light spot intersection points without accounting for refraction are denoted as $X_{1}$ and $X_{2}$; however, these must be estimated indirectly using a refraction model. Finally, the length of the vertical slit is denoted by $y$ \cite{Fan2015, Fan2016}. An overview of the V-slit model representation body of literature is presented in Table \ref{tab:mrlit}.

\textbf{N-slit model.}\label{model:nslit} The N-slit model representation is an architecture-specific approach designed to predict the light spot intersections produced by an N-shaped slit mask in combination with a linear digital array. At any given moment, three light spots are formed on the detector array, which are used to estimate the two-axis incident angle of incoming light. When the incident light projects at an angle $\alpha$ in the Z-Y plane, all three spots shift by an equal distance. In contrast, when the light projects at an angle $\beta$ in the Z-X plane, only the left and right spots shift. Under a composite incident angle, the center spot shifts solely due to $\alpha$, while the left and right spots are influenced by both $\alpha$ and $\beta$ \cite{Maji2015}.

The work by Maji et al. \cite{Maji2015} is reviewed as a case-study for the N-slit model representation. The N-slit model is presented in Equations \ref{eq:nslit1} \& \ref{eq:nslit2}.

\begin{equation}\label{eq:nslit1}
\alpha=\tan^{-1}\left ( \frac{C_{m}-C_{r}}{h}\cdot K \right )
\end{equation}

\begin{align}\label{eq:nslit2}
    \begin{split}
        \beta =\tan^{-1}&\Biggl[ \Big( \frac{(L_{m}-L_{r})+(R_{m}-R_{r})}{2h\tan(\delta)}\\
        &-\frac{C_{m}-C_{r}}{h\tan(\delta)} \Big) \cdot K \Biggr]
    \end{split}
\end{align}

The reference centroids are defined as $L_{r}$ for the left centroid, $C_{r}$ for the center centroid, and $R_{r}$ for the right centroid. In addition, the shifted centroids are defined as $L_{m}$ for the left centroid, $C_{m}$ for the center centroid, and $R_{m}$ for the right centroid. Moreover, the focal length is $h$, the angle formed between the central and diagonal slit is $\delta$, and $K$ is the refraction error correction scale factor \cite{Maji2015}. An overview of the N-slit model representation body of literature is presented in Table \ref{tab:mrlit}.

\textbf{Camera model.}\label{model:cam} The camera model representation is an architecture-specific method that estimates the two-axis incident angle from a lensed imaging system with a digital sensor. The method uses the intrinsic parameters of the calibrated camera and the captured spot centroid coordinates for the state estimate. The work by Saleem et al. \cite{Saleem2019} is reviewed as a case-study for the camera model representation. The camera model is presented in Equation \ref{eq:cameq}.

\begin{align}\label{eq:cameq}
\begin{split}
\phi=&\tan^{-1}\left ( \frac{u-p_{x}}{f} \right )\\
\theta=&\tan^{-1}\left ( \frac{-(v-p_{y})}{f} \right ) 
\end{split}
\end{align}

In the equations above, the focal length is $f$, the principal point is $(p_{x}, p_{y})$, and the image spot coordinates are $(u, v)$. An overview of the camera model representation body of literature is presented in Table \ref{tab:mrlit}.

\subsubsection*{Multi-sensor fusion}\label{model:msf}

A multisensor fusion model is a subtype of geometric model that is developed to synthesize observations from multiple sensors to improve the estimation accuracy or extend the capabilities of a system. In the following sections, the multisensor methods are introduced including: the base model, solar panel, pyramidal, and hemispherical approaches. For each, the associated models are introduced and example implementations are presented.

\textbf{Basic.}\label{model:base} The basic model representation is a multisensor fusion method that uses the synthesis of information from sensors on each face of the satellite body to compute the incident light vector. Typically, this method uses body-mounted analog sensors resulting in a coarse but large FOV estimate.

The work by Allegeier et al. \cite{Allegeier2009} is reviewed as a case-study for the basic multisensor fusion model representation. The incident angle is calculated by measuring and processing the incident solar flux along each face of the satellite body. The intensities incident on the six faces of the satellite are measured and are used to form the intensities $I_{a}$, $I_{b}$, $I_{c}$ below. These intensities make up the components of the sun vector to be calculated for the combined sensor estimate $^{b}\underline{\hat{S}}$ \cite{Allegeier2009}. The basic model is presented below in Equation \ref{eq:msfstd}.

\begin{equation}\label{eq:msfstd}
^{b}\underline{\hat{S}}=\begin{bmatrix}
\frac{\pm I_{a}}{\sqrt{I_{a}^{2}+I_{b}^{2}+I_{c}^{2}}}\\ 
\frac{\pm I_{b}}{\sqrt{I_{a}^{2}+I_{b}^{2}+I_{c}^{2}}}\\ 
\frac{\pm I_{c}}{\sqrt{I_{a}^{2}+I_{b}^{2}+I_{c}^{2}}}
\end{bmatrix}\textup{, where: }\left\{\begin{matrix}
I_{a}=\max(I_{x-},I_{x+})\\ 
I_{b}=\max(I_{y-},I_{y+})\\ 
I_{c}=\max(I_{z-},I_{z+})
\end{matrix}\right.
\end{equation}

It is assumed that the sensors exhibit a cosine response in the above equations. Furthermore, each sensor must have a minimum FOV of 110$^{\circ}$ to ensure full $4\pi$ sr coverage. One limitation of this model is the lack of correction for albedo effects \cite{Allegeier2009}. An overview of the basic model representation body of literature is presented in Table \ref{tab:mrlit}.

\textbf{Solar panel.}\label{model:solarpanel} The solar panel model representation is a multisensor fusion approach that utilizes data from the satellite’s solar panels to estimate the incident light vector. While algorithmically similar to the standard model described above, this method eliminates the need for dedicated photodiode sensors. Instead, analog currents or voltages from each of the six panel faces are processed to compute the incident angle. The estimation relies on the assumption that the current generated by incident light is proportional to the cosine of the incidence angle. However, this assumption can introduce errors, as the output current may be influenced by factors such as signal interference, Earth albedo, and solar panel degradation over time \cite{Nurgizat2021}.

The work by Nurgizat et al. \cite{Nurgizat2021} is reviewed as a case-study for the solar panel multisensor fusion model representation. The solar panel model is presented in Equation \ref{eq:msfsc}.

\begin{equation}\label{eq:msfsc}
\overrightarrow{n}_{ns}=\frac{\sum_{i=1}^{6}\overrightarrow{n}_{i}I_{0}}{\sum_{i=1}^{6}I_{i}^{2}}
\end{equation}

The incident vector $\overrightarrow{n}_{ns}$ is determined using the above calculation, where $\overrightarrow{n}_{i}$ represents the normal vector of the $i$-th exposed panel, $I_{0}$ is the reference current produced by a panel under full sunlight, and $I_{i}$ is the measured current output from the $i$-th panel \cite{Nurgizat2021}. An overview of the solar panel model representation body of literature is presented in Table \ref{tab:mrlit}.

\textbf{Pyramidal.}\label{model:pyra} The pyramidal model representation is a multisensor fusion approach based on a pyramidal variant of the non-planar sun sensor. This sensor estimates the solar incident angle using photodiodes or solar cells mounted on the lateral faces of a regular pyramid structure. Non-planar sun sensors offer the advantage of wide FOV coverage without requiring a large number of individual sensors, which helps reduce the overall size and mass of the satellite. However, they typically suffer from coarse accuracy due to signal interference. The pyramidal model was introduced to address this issue by compensating for both constant and zero-mean interference effects inherent in non-planar configurations \cite{Wang2019}.

The work by Wang et al. \cite{Wang2019} is reviewed as a case-study for the pyramidal multisensor fusion model representation. The pyramidal sensor in the study is developed for a sensor array mounted on the surfaces of a regular M-pyramid. The angular geometry of the pyramid is defined by $\alpha_{i}$ for the azimuth and $\gamma$ for the elevation angles. The pyramidal model is presented  in Equations \ref{eq:msfpyra1} \& \ref{eq:msfpyra2}. 

\begin{equation}\label{eq:msfpyra1}
\begin{matrix*}[l]
\mathbf{b}_{x}=(\sin\alpha_{1}\cos\gamma\cdots\sin\alpha_{M}\cos\gamma)^{\top}\\ 
\mathbf{b}_{y}=(\cos\alpha_{1}\cos\gamma\cdots\cos\alpha_{M}\cos\gamma)^{\top}\\ 
\mathbf{b}_{z}=(\sin\gamma\cdots\sin\gamma)^{\top}
\end{matrix*}
\end{equation}

The equation for the sun vector solution $\mathbf{s}$ is shown in Equation \ref{eq:msfpyra2}, where $\xi$ is the measurement coefficient, $\mathbf{e}$ is the measurement vector, and $(\mathbf{b}_{x},\mathbf{b}_{y},\mathbf{b}_{z})$ are the respective components of the unit normal vectors of all M illuminated pyramidal surfaces.

\begin{equation}\label{eq:msfpyra2}
\mathbf{s}=\xi \begin{pmatrix}
\mathbf{b}_{x}^{\top}/\mathbf{b}_{x}^{\top}\mathbf{b}_{x} \\[2pt]
\mathbf{b}_{x}^{\top}/\mathbf{b}_{x}^{\top}\mathbf{b}_{x} \\[2pt]
\mathbf{b}_{x}^{\top}/\mathbf{b}_{x}^{\top}\mathbf{b}_{x}
\end{pmatrix}\mathbf{e}
\end{equation}

An overview of the pyramidal model representation body of literature is presented in Table \ref{tab:mrlit}.

\textbf{Panoramic.}\label{model:pano} The panoramic model representation is a multisensor fusion method that operates on the spherically arranged variant of a non-planar sun sensor \cite{Zhang2022b}. The primary advantage of such a configuration is an enhanced FOV with full spherical coverage. 

The study by Zhang et al. \cite{Zhang2022b} is reviewed as a case study for the panoramic multisensor fusion model representation. In this work, 991 solar cells are mounted on the surface of a spherical satellite, with 16 selected for use in the panoramic sensor measurements. These sensors are strategically distributed to provide full $4\pi$ sr FOV coverage, ensuring continuous sun detection regardless of satellite attitude. The proposed model estimates the incident angle using measurements from N solar cells with known installation vectors. To account for varying contributions among sensors, a weighted approach is applied, as not all measurements equally influence the final estimate. The panoramic model is detailed in Equation \ref{eq:msfpano}.

\begin{equation}\label{eq:msfpano}
\overrightarrow{s}=\left ( A^{\top}WA \right )^{-1}WA^{\top}\begin{bmatrix}
\frac{I_{ph,1}}{I_{\max,T_{0}}-K(T_{1}-T_{0})}\\ 
\frac{I_{ph,2}}{I_{\max,T_{0}}-K(T_{2}-T_{0})}\\ 
\vdots \\ 
\frac{I_{ph,N}}{I_{\max,T_{0}}-K(T_{N}-T_{0})}
\end{bmatrix}
\end{equation}

In the above formulation, the sensor outputs are assumed to follow a standard cosine response, and measurements taken at large angles of incidence are excluded. The incident angle estimate, denoted by $\overrightarrow{s}$, is computed using several parameters: $A$ represents the set of installation vectors, $W$ is the corresponding weight matrix, $I_{ph,i}$ is the current measured by the $i$-th solar cell, $T_{i}$ is its temperature, $I_{\max,T_{0}}$ is the reference current under perpendicular sunlight at a baseline temperature, and $K$ is the temperature compensation coefficient. A summary of the literature on the panoramic model representation is provided in Table \ref{tab:mrlit}.

\subsection{Physics-informed model}\label{model:pi}

The physics-informed model calibration process involves developing compensation models to correct for errors arising from physical phenomena inherent to the sensor configuration. These errors may stem from various sources, including environmental influences, manufacturing and assembly imperfections, optical distortions, electrical effects, and sensor interference. For a clearer understanding of the relationship between error sources and sensor architectures, refer to Figure \ref{fig:errorsankey}. In the following section, physics-informed model representations are discussed in more detail below.

\subsubsection*{Environmental}\label{model:enviro}

The space environment compensation model (SECM) is a subtype of the physics-informed approach that accounts for the influence of space environmental factors on sensor performance. Examples of such factors include variations in solar characteristics, temperature fluctuations, and radiation-induced damage.

\textbf{Solar compensation.}\label{model:solar} 
Inaccurate modeling of solar characteristics can introduce errors in the final sensor estimate. The study by Antonello et al. \cite{Antonello2018} is examined as a case study for the solar compensation model representation. In this work, the model accounts for the effects of non-parallel incident light on the detector, which alters the shape of the projected light spot and results in a translational bias of the spot center. An overview of the solar compensation body of literature is presented in Table \ref{tab:mrlit}.

\textbf{Temperature compensation.}\label{model:temp} Analog sun sensor responses are highly sensitive to temperature variations in the space environment. To account for this effect, a temperature compensation coefficient can be incorporated into the mathematical model governing the photodiode’s current output. 

The study by Lu et al. \cite{Lu2017a} is reviewed as a case study for the temperature compensation model representation. In this work, a mathematical model is developed to describe the short-circuit current of a photodiode while accounting for temperature effects. The incident angle can be estimated when the current outputs from at least three coplanar photodiodes are known. The sun vector is calculated using Equation \ref{eq:tempcomp}.

\begin{flalign}\label{eq:tempcomp}
&\left\{ \begin{aligned}
    x &= \frac{kI_{sc1}}{1+C(T_{1}-T_{0})}\\
    y &= \frac{kI_{sc2}}{1+C(T_{2}-T_{0})}\\ 
    z &= \frac{kI_{sc3}}{1+C(T_{3}-T_{0})}
\end{aligned} \right.&&\\
&\left\lbrace \begin{aligned}
    (I_{sc1},I_{sc2},I_{sc3}) &= \max(I_{scX})\\
    k &= \left\{\begin{matrix}
        1 & (X=a,c,e)\\ 
        -1 & (X=b,d,f)
    \end{matrix}\right.
\end{aligned} \right\rbrace &&\nonumber
\end{flalign}

In Equation \ref{eq:tempcomp}, $(x,y,z)$ are the 3D coordinates of the solar vector, $I_{sc}$ is the short-circuit current, $T$ is the photodiode temperature, $C$ is the temperature coefficient of the short-circuit current, and $(a, b, c, d, e, f)$ represent the six installed photodiodes. From this formulation, the measurement error due to temperature variations is mitigated through the compensation model. An overview of the temperature compensation body of literature is presented in Table \ref{tab:mrlit}.

\subsubsection*{Optical}\label{model:optic}

The optical compensation model is a sub-type of physics-informed model that accounts for errors due to wave properties such as refraction, reflection, and diffraction. Error from refraction occurs when the incident light is deflected as it passes through the interface medium of the sensor mask. The interface is often a protective glass covering to protect the detector from debris and contamination. This error can be mitigated through a refraction error compensation scheme. However, effective correction requires a thorough understanding of the refractive indices of the interface materials. 

The work by Wei et al. \cite{Wei2011} is reviewed as a case-study for the refraction compensation model representation. In this study, an N-shaped slit mask with a linear CCD array configuration is implemented. The detector is protected by a glass layer through which the incident light diffracts, which introduces the primary error source for the sun sensor. To address this, the authors developed an iterative refraction compensation method to mitigate the refraction error.

The correction coefficient $k$ is introduced to measure the refraction error in the model. The most critical step in correcting the refraction error is to properly calculate the value of $k$. However, this is not trivial since $k$ is a function of the incident angle. Hence, an iterative algorithm is proposed to estimate the incident angle $\theta$. Once the incidence angle is found, the correction coefficient, and thereby sun angle relations, can be solved. The correction coefficient is described by Equation \ref{eq:refcomp1}. The sun angle relations are described by Equation \ref{eq:refcomp2}. Lastly, the basic iterative equation as described by Equation \ref{eq:refcomp3} is used to solve for the incident angle $\theta$.

\begin{flalign}\label{eq:refcomp1}
k =\myfrac{(h_{2}+h_{3}+h_{4})\tan\theta}
{\begin{matrix*}[l]
\Big(h_{2}\tan\theta +h_{3}\tan\left [ \arcsin\frac{n_{1}\sin\theta}{n_{2}} \right ]&\\
+h_{4}\tan\left [\arcsin\frac{n_{1}\sin\theta}{n_{3}} \right ]\Big)
\end{matrix*}}&&
\end{flalign}

\begin{flalign}\label{eq:refcomp2}
    & \mu = \arctan\left ( k\frac{y_{1m}-\overline{y}_{1}}{h} \right )&&\nonumber\\
    & \nu =\gamma=\arctan\left ( k\frac{(y_{2m}-\overline{y}_{2})-(y_{1m}-\overline{y}_{1})}{h} \right )&&\\ 
    & \beta = \arctan\left ( k\frac{y_{1m}-\overline{y}_{1}}{\sqrt{\left [ (y_{2m}-\overline{y}_{2})-(y_{1m}-\overline{y}_{1}) \right ]^{2}k^{2}+h^{2}}} \right )\nonumber
\end{flalign}

\begin{flalign}\label{eq:refcomp3}
(h_{2}+h_{4})\tan\theta+h_{3}\tan\left [ \arcsin\frac{\sin\theta}{n_{2}} \right ]-l=0&&
\end{flalign}

For Equations \ref{eq:refcomp1}-\ref{eq:refcomp3}, $k$ is the correction coefficient, $l$ is the refraction error, and ($\mu$, $\nu$) denote the sunray horizontal and azimuth orientations, respectively. Furthermore, $\overline{y}_{1}$ is the initial distance between the central sun spot and the origin, $\overline{y}_{2}$ is the initial distance between the sideways sun spot and the origin, and ($y_{1m}$, $y_{2m}$) are the measurement distances of ($\overline{y}_{1}$, $\overline{y}_{2}$), respectively. Finally, $h$ is the focal length, ($n_{1}$, $n_{2}$, $n_{3}$) denote the light refractive index of the vacuum, and $\theta$ is the estimated incident angle. An overview of the refraction compensation body of literature is presented in Table \ref{tab:mrlit}.

\subsubsection*{Electrical}\label{model:elec}

Electrical error compensation methods rely on model-based representations to address issues such as sensor non-linearity, noise, limited dynamic range, dark current, and measurement repeatability \cite{Jerram2020}. These effects typically need to be filtered out during real-time operation to enable effective correction. 

\textbf{Measurement noise compensation.}\label{model:mnc} Noise errors are derived from the noise floor during low signal levels. To compensate for sun angle estimation errors caused by random measurement noise, various online iterative calibration techniques—such as Kalman filtering—can be employed. For effective compensation of both sensor noise and model inaccuracies, the measurement noise is typically modeled as an independent, zero-mean Gaussian random variable \cite{OKeefe2017}.

The study by O’Keefe et al. \cite{OKeefe2017} is examined as a case study demonstrating a measurement noise compensation model. In this work, a coarse sun sensor is calibrated on-orbit using a consider Kalman filter approach. The proposed model is based on Lambert’s cosine law and incorporates the effects of surface albedo and field-of-view (FOV) limitations. The measurement noise model for the sun sensor is presented in Equation \ref{eq:mnoise}, expressed in terms of the photodiode output voltage \cite{OKeefe2017}.

\begin{flalign}\label{eq:mnoise}
    & V=C(V_{d}+V_{\alpha}+\nu_{V})&&\nonumber\\
    & V_{d}=\left\{\begin{matrix}
n^{\top }\frac{s}{\left \| s \right \|} & \text{ if } \left ( n^{\top }\frac{s}{\left \| s \right \|}\geq \cos\psi \right )\wedge (B\notin S)\\ 
0 & \text{ if } \left ( n^{\top }\frac{s}{\left \| s \right \|}<  \cos\psi \right )\vee  (B\in S)
\end{matrix}\right.&&\\ 
    & V_{\alpha}=\left\{\begin{matrix}
\begin{matrix*}[l]
-\frac{1}{\pi}\iint_{A}\frac{\alpha}{\left \| \mathbf{r}_{AB}^{2} \right \|}
\left ( n^{\top}_{A}\frac{\mathbf{s}_{\bigoplus}}{\left \| \mathbf{s}_{\bigoplus} \right \|} \right )\\ 
\times\left ( n^{\top}_{A}\frac{\mathbf{r}_{AB}}{\left \| \mathbf{r}_{AB} \right \|} \right )
\left ( n^{\top}\frac{\mathbf{r}_{AB}}{\left \| \mathbf{r}_{AB} \right \|} \right )dA
\end{matrix*} & \text{ if } B\notin S\\ 
0 & \text{ if } B\in S
\end{matrix}\right.\nonumber
\end{flalign}

In the above equation, the calibration factor is $C$, the zero-mean Gaussian random variable to account for sensor noise and model error is $\nu_{V}$, the sun vector in the body frame is $\mathbf{s}$, the half angle of the sensor FOV is $\psi$, the surface of the Earth visible to the spacecraft and the sun is $\mathbf{A}$, the normalized differential area of $\mathbf{A}$ is $n_{A}$, the vector from the Earth to the sun is $\mathbf{s}_{\bigoplus}$, the vector from $dA$ to the spacecraft body is $\mathbf{r}_{AB}$, the albedo of $dA$ is $\alpha$, the spacecraft position is $B$, and the part of the spacecraft orbit under Earth shadow is $S$ \cite{OKeefe2017}. An overview of the measurement noise compensation body of literature is presented in Table \ref{tab:mrlit}.

\textbf{Angular loss compensation.}\label{model:nonlin} The photovoltaic angular loss correction or Kelly Cosine method is an adaptation of the Cosine law to correct for the inexact modeling of photodiode output in photovoltaic mode, especially at large incidence angles. For small incident angles, the photodiode response follows the Cosine law. However, the Cosine law is not valid for large incident angles due to the inhibition of absorbed light from reflections. Moreover, the Cosine law yields large errors at incident angles beyond 50$^{\circ}$ and a vanishing output for angles beyond 85$^{\circ}$ \cite{Yousefian2017}.

The work by Yousefian et al. \cite{Yousefian2017} is reviewed as a case-study for the angular loss compensation model representation. In this study, a non-flat sun sensor configuration composed of six photodiodes is implemented with a FOV of 110$^{\circ}$. The Kelly Cosine law is a more accurate non-linear model representation for the full range of expected incident angles during photodiode operation \cite{Yousefian2017}. The Kelly Cosine law is described by Equation \ref{eq:kellycos}.

\begin{align}\label{eq:kellycos}
    \begin{split}
        I_{i}&=\alpha_{s}I_{0,i}\bigl( -0.369\cos^{3}\theta+0.637\cos^{2}\theta\\
        &+0.750\cos\theta - 0.015 \bigr)+\eta_{i}
    \end{split}
\end{align}

The measured output current of the photodiode is $I_{i}$, the ratio of sunlight relative to a test reference is $\alpha_{s}$, the output of the photodiode at the test reference is $I_{0,i}$, and the measurement error is $\eta_{i}$. An overview of the angular loss compensation body of literature is presented in Table \ref{tab:mrlit}.

\subsubsection*{Assembly}\label{model:assem}

Assembly error compensation methods differ from traditional geometric methods since they account for structural deformations and assembly offsets during real-time operations. Therefore, this physics-informed method typically uses online iterative calibration techniques such as Kalman filtering for deformation error correction. Since each assembly method model is unique to a given sensor configuration, we will not present a specific case-study formulation. Nevertheless, it is still of use to briefly describe an example case-study implementation for completeness.

In the study by Rahdan et al. \cite{Radan2019}, the author's account for the installation error, offsets of the central point of the CCD array, errors in the filter thickness, and sensor misalignment. This is accomplished through the Levenberg–Marquardt algorithm for offline calibration and the Extended Kalman Filter (EKF) approach for online calibration. An overview of the assembly error compensation body of literature is presented in Table \ref{tab:mrlit}.

\subsubsection*{Interference}\label{model:interf}

Sensor interference errors arise from both external and internal sources. External sources include factors such as albedo, stray illumination, and shadowing or reflections from the satellite body. Internal sources involve issues like sensor self-shadowing, internal reflections, and light leakage. To mitigate these effects, interference compensation models can be employed. These models characterize the interference behavior as a physics-informed process, thereby enabling correction of both external and internal interference errors. 

\textbf{Albedo compensation.}\label{model:albedo} Analog sun sensors are prone to external errors that lower their accuracy, in particular from the effects of Earth albedo. This is due to the fact that it is not possible to differentiate between light from the Sun and other spurious sources \cite{Frezza2022}. The effects of albedo depend on many factors, including the reflectivity of the Earth’s surface, the satellite attitude, and the position of the Sun. In some circumstances, error from albedo can reduce the accuracy from analog sensors by up to 20$^{\circ}$ \cite{Soken2023}. Authors have proposed a few different approaches to mitigate the effects of albedo, which will be explored below.

The first method reviewed is the classical method of maximum currents \cite{Frezza2022}. This is the simplest albedo effect mitigation strategy, where the highest current is selected for each pair of photodiodes and used to create the sun vector. The formulation for this method is shown in Equation \ref{eq:albedo1}.

\begin{equation}\label{eq:albedo1}
s_{meas}=\begin{bmatrix}
\max\left \{ I_{+x},I_{-x} \right \}\\ 
\max\left \{ I_{+y},I_{-y} \right \}\\ 
\max\left \{ I_{+z},I_{-z} \right \}
\end{bmatrix}
\end{equation}

Another method is the Summarized Sun and Earth (SSE) algorithm proposed by Bhanderi et al. \cite{Bhanderi2006}. Here, the albedo is included in the reference vector as the sum of Sun and albedo vectors. The albedo is estimated with a model of the visible radiation environment at the satellite altitude without directly estimating the sun vector. The sum of the combined vectors is then compared with the sensor measurements. This method is limited by the accuracy of the albedo model, local weather, direct solar reflection, and the use of a single vector for the albedo \cite{Frezza2022}.

\begin{equation}\label{eq:albedo2}
s\mathbf{a}_{body}=\begin{bmatrix}
I_{+x}-I_{-x}\\ 
I_{+y}-I_{-y}\\ 
I_{+z}-I_{-z}
\end{bmatrix}
\end{equation}

The Sun and Albedo Inputs Estimation (SAIE) model, proposed by Frezza et al. \cite{Frezza2022}, provides a closed-form solution for simultaneously estimating Sun and albedo inputs when up to five satellite surfaces are illuminated by albedo. This method utilizes measurements from both photodiodes and a three-axis magnetometer to derive the Sun vector estimate. The algorithm defines four distinct problem types, based on the overlap of sensor illumination by the Sun, albedo, or both. For the following equations, $\mathbf{s}$ is the sun vector, $\mathbf{a}$ is the albedo vector, and $I$ are the measured photodiode currents. In Problem Type 0, where no overlap occurs, an example solution is presented in Equation \ref{eq:albedo3a} below. 

\begin{align}\label{eq:albedo3a}
    \begin{split}
        \mathbf{s}=&\begin{bmatrix}
        0 & I_{+y} & -I_{-z}
        \end{bmatrix}\\
        \mathbf{a}=&\begin{bmatrix}
        I_{+x} & -I_{-y} & 0
        \end{bmatrix}
    \end{split}
\end{align}

For Problem Type 1, in which there is one overlap, an example solution is shown in Equation \ref{eq:albedo3b} below.

\begin{align}\label{eq:albedo3b}
    \begin{split}
        \mathbf{s} &= \begin{bmatrix}
        \sqrt{1-(I_{+y}^{2}+I_{-z}^{2})} & I_{+y} & -I_{-z}
        \end{bmatrix}\\
        \mathbf{a} &= \begin{bmatrix}
        I_{+x} & -I_{-y} & 0
        \end{bmatrix}
    \end{split}
\end{align}

For Problem Type 2, in which there are two overlaps, magnetometer measurements are required to restrict the possible solutions of the sun vector on a cone around the measured magnetic field vector. For Problem Type 3, in which there are three overlaps, no solution exists \cite{Frezza2022}.

Finally, a Deep Neural Network (DNN) approach to albedo error measurement correction was proposed by Sozen et al. \cite{Sozen2021,Soken2023}. The primary advantage of this approach is that it does not require an additional albedo or sensor model for the error compensation process. The network is trained on analog sensor output voltages, the Sun reference directions, and the satellite attitude over a small number of orbital periods. The training can be done offline with simulated data or online with measurements from another on-board sensor. For the final estimate, only voltage inputs from the analog sensor are required for corrected measurements \cite{Soken2023}. An overview of the albedo compensation body of literature is presented in Table \ref{tab:mrlit}.

\textbf{Shadowing compensation.}\label{model:shad} Self-occultation or shadowing errors occur when incoming radiation is partially blocked by the sensor mask or other structural components, leading to changes in sensor response. The most common source of this error is the thickness of the mask, which can alter the shape of the illumination pattern projected onto the detector. A shadowing compensation model mitigates these effects by accounting for the geometry of the mask and its potential to cause self-shadowing \cite{Antonello2018}. 

The study by Antonello et al. \cite{Antonello2018} is reviewed as a case study demonstrating the application of a shadowing compensation model. In this work, an enhanced model is proposed for a pinhole mask combined with a digital array sensor to account for self-shadowing effects resulting from mask thickness. Under ideal conditions, where the mask thickness is negligible, the projected light spot on the detector maintains the same shape as the mask. However, in non-ideal conditions, the finite thickness of the mask distorts the shape of the light spot, introducing shadowing errors. The center of this modified light spot is determined by the projected centers of the upper and lower surfaces of the mask. The shadowing compensation model is described by Equation \ref{eq:shadcomp}  \cite{Antonello2018}. 

\begin{flalign}\label{eq:shadcomp}
&\left\{\begin{matrix*}[l]
R_{C0}=(t+h)\tan\left ( \frac{\pi}{2}-\theta \right )\\ 
R_{C1}=h\tan\left ( \frac{\pi}{2}-\theta \right )\\ 
\Delta C=h\tan\left ( \frac{\pi}{2}-\alpha \right )
\end{matrix*}\right.&&\\
&\left\{\begin{matrix}
R^{2}_{C0}-2R_{C0}(x\cos\Phi+y\sin\Phi)+x^{2}+y^{2}=d^{2}/4\\ 
\text{for } R_{C0}:-\frac{d}{2}<r<R_{C0}+\frac{\Delta C}{2}\\ 
R^{2}_{C1}-2R_{C1}(x\cos\Phi+y\sin\Phi)+x^{2}+y^{2}=d^{2}/4\\ 
\text{for } R_{C1}:-\frac{\Delta C}{2}<r<R_{C1}+\frac{d}{2}
\end{matrix}\right.\nonumber&&
\end{flalign}

In the equation above, the distance of the upper and lower mask points from the reference frame origin is ($R_{C0}, R_{c1}$), respectively. The distance between the two mask arc centers is $\Delta C$. Finally, the piecewise equations of the projected light spot shape are presented. The focal length is $h$, the mask thickness is $t$, the azimuth angle is $\Phi$, the elevation angle is $\theta$, and the pinhole aperture diameter is $d$. An overview of the shadowing compensation body of literature is presented in Table \ref{tab:mrlit}.

\subsection{Multiplexing model}\label{model:multiplex}

The multiplexing model representation uses coding pattern rules related to the mask configuration to enable sun vector estimation with very high accuracy and a large field of view. The associated sensor configuration is usually comprised of hundreds or more of apertures and tens of sub-FOVs to extend the full sensor FOV coverage \cite{Wang2015}. To robustly estimate the incident angle within a given sub-FOV, a unique and unambiguous mask pattern is required. 
The two main categories of multiplexing model representations are the periodic aperture and coded aperture variants. Both of these approaches are discussed in detail in the following sections, their implementations presented and merits compared. 

\subsubsection*{Periodic Aperture}\label{model:periodic}

The periodic aperture model representation is a variant of the multiplexing method that uses a periodic slit pattern for the mask configuration and an associated pattern inverse model for sun angle estimation. The periodic pattern that is projected is captured and the signal phase of the projected image is retrieved. Thereafter, the obtained phase is processed with a correlation function to estimate the sun angle. This methodology is sensitive to phase errors in the captured signal and therefore requires the careful implementation of a compensation model to ensure accurate results.

Next, three distinct periodic aperture model representations proposed in the literature are discussed, followed by the presentation of a selected case study implementation. The first error model and calibration approach for periodic sun sensors to be discussed is the Optical Vernier Measuring Principle proposed by Chen et al. \cite{Chen2006}. This methodology differs from the others presented in that it uses the projected image position information rather than the phase to estimate the sun angle. The mask design consists of an optical vernier movable ruler projective part and an optical vernier fixed ruler. The movable ruler part consists of narrow periodic slits that modulate the incident light, thereby shifting the projected invariable gap pattern by the incidence angle on the CCD detector fixed ruler \cite{Chen2006}. The sun vector is then estimated from the primary position information and processing the pinpoint position information from pattern codes.

For the second model, Tsuno et al. \cite{Tsuno2019,Kondo2024} propose an error model and calibration approach for periodic UFSS systems. The sensor configuration features incident light passing through a periodic reticle with 16 uniformly spaced slits, projecting onto a linear CCD array. Unlike previous methods, this design utilizes multiple identical slits, leveraging the principle that accuracy improves by averaging multiple independent observations. Specifically, if sensor errors are independent, accuracy improves by a factor of for $N^{-1/2}$, while the error from a single slit can be reduced by a factor of $N^{-1} I$. Instead of detecting the position of individual slit projections, this approach estimates the sun angle by capturing the phase of the reticle current signal. A signal correlation algorithm is then used to solve for and compensate the resulting phase error \cite{Tsuno2019,Kondo2024}. 

Lastly, Fan et al. \cite{Fan2013} propose an error model and calibration approach for periodic encoded sun sensors. The proposed sensor configuration involves incident light passing through a semi-cylindrical lens, into an etched entrance slit, and then projected onto a code dial with photocells underneath. The photocells receive incident light through the code dial, which is converted into a current signal. The encoded rows on the code dial are composed of both coarse-code and fine-code patterns for improved accuracy. However, the uncalibrated sensor accuracy is limited by signal and geometric errors. The periodic aperture compensation model accounts for structural errors such as offset and tilt, while the measured current signal is processed with a signal correlation function to correct the phase error of fine-code output current \cite{Fan2013}.

The work by Fan et al. \cite{Fan2013} is reviewed as a case-study for the periodic aperture calibration approach. The compensation model accounts for structural errors due to offset and tilt as well as fine-code algorithm error due to phase errors in the output current. The compensation model is presented in Equation \ref{eq:pamulti}.

\begin{flalign}\label{eq:pamulti}
&\alpha_{comp}=\tan^{-1} \left ( \frac{b-d\tan\left ( \alpha_{s} \right )}{c\tan\left ( \alpha_{s}\right )-a } \right )  \textup{, where: }&&\\
    &\alpha_{s}=\alpha_{1}+\overline{\alpha}_{2} \textup{ and } \overline{\alpha}_{2}=\alpha_{2}+k\sin\left ( 4\pi\alpha+t \right ) &&\nonumber
\end{flalign}

For Equation \ref{eq:pamulti}, $\alpha_{comp}$ is the compensated output sun angle, $\alpha_{1}$ is the output of the coarse-code, $\alpha_{2}$ is the output of the fine-code, $\overline{\alpha}_{2}$ is output angle of the compensated fine-code, ($a, b, c, d$) are conversion coefficients between the sunlight incidence plane and the code dial reference frame, $\alpha$ is the measured incident 
angle of sunlight, $k$ is the amplitude of the fine-code error, and $t$ is the phase of the fine-code error. An overview of the periodic aperture body of literature is presented in Table \ref{tab:mrlit}.

\subsubsection*{Coded Aperture}\label{model:code}

The coded aperture model representation is a variant of the multiplexing method that uses a coded pattern for the mask configuration and an associated pattern coding ruleset for sun angle estimation. Specifically, the mask configuration is arranged as an array of apertures whose projection can be unambiguously mapped over a series of sub FOVs. However, there is no standardized mask pattern and a number of studies have investigated various mask coding rule approaches, as discussed below. The coded aperture method enables very high accuracy sun angle estimation, while also maintaining a large operating FOV.

In 2013 Wei et al. \cite{Wei2013} proposed a multiplexing sensor based on ERS imaging mode of an APS detector. The mask configuration is composed of both periodic and positioning apertures arranged in 7x9 sub-FOVs to obtain a 105$^{\circ}$x105$^{\circ}$ FOV. The positioning apertures, which are used to determine the sub-FOV, are grouped in sets of three apertures over the mask plane to convey distance information. The periodic apertures are periodically arranged in a diagonal fashion over the mask plane and are used to process the sun angle after the sub-FOV has been determined.

In 2014 Wei et al. \cite{Wei2014} proposed a multiplexing sensor architecturally similar to the previous work but with only positioning apertures.
The apertures are arranged in 13x13 sub-FOVs to obtain a 105$^{\circ}$x105$^{\circ}$ FOV and an accuracy of 5 arcsec. The positioning apertures are grouped over the mask plane in sets of three apertures, however the distance between the three apertures is varied to ensure a unique feature mapping.

In 2015 Wang et al. \cite{Wang2015} presented an expansion of the previous 2014 work \cite{Wei2014}, where a multiplexing sensor is proposed with a mask configuration of varying and coded apertures. The mask configuration is arranged in 13x13 sub-FOVs to obtain a 120$^{\circ}$x120$^{\circ}$ FOV and an accuracy of 1.32 arcsec. As previously, the positioning apertures are grouped over the mask plane in sets of three apertures, however the size of the apertures and the distance between them is varied to ensure a unique feature mapping. In addition, the apertures are varied in size over the sub-FOVs according to diffraction theory optimal parameters. This change allows for improved accuracy over the full FOV range.

In 2017 Wei et al. \cite{Wei2017} proposed a multiplexing sensor applied to a wireless digital sun sensor. The mask configuration is composed of asymmetric coded apertures arranged in 7 sub-FOVs to obtain a 100$^{\circ}$ conical FOV with an accuracy of 0.01$^{\circ}$. Specifically, the apertures are arranged obliquely over the mask plane to ensure the projections can be uniquely identified and located for sun angle estimation.

Lastly, Zhang et al. \cite{Zhang2021,Zhang2022a} proposed a multiplexing sensor applied to a LCE digital sun sensor. The LCE is composed of a circular pattern of 1026 apertures with varying sizes. The mask configuration is arranged in 37 sub-FOVs to obtain a 120$^{\circ}$x120$^{\circ}$ FOV and an accuracy of 0.0023$^{\circ}$. Similar to the 2015 work by Wang et al. \cite{Wang2015}, the apertures are changed in shape according to diffraction theory. The sub-FOVs and spot-aperture correlations can be determined by processing the coding information embedded in the acquired spot images. Since the coding rules vary over the mask region, a maximum correlation peak is used to identify the matching sub-region of the acquired image. Thereafter, the sun orientation can be determined from the associated matched aperture and light spot positions. An overview of the coded aperture body of literature is presented in Table \ref{tab:mrlit}.

\subsection{Neural network-based model}\label{model:nn}

The neutral network-based model is a machine learning method that uses interconnected artificial neurons in a layered structure to mimic the structure and function of the human brain. This model representation is powerful since it is able to capture relationships between large amounts of measured sensor data. Furthermore, the methodology is particularly well suited to sun sensor calibration since the sensor responses are often complex, non-linear, high-dimensional, and noisy. The two primary variants of this approach are the artificial neural network (ANN) and deep neural network (DNN), where the ANN has one or two hidden layers, while the DNN consists of multiple layers between the input and output layers. Implementations of the two variants are presented in the following sections.

\subsubsection*{Artificial neural network}\label{model:ann}

The artificial neural network (ANN) based model representations enable accurate modeling of complex systems without the need to manually design complicated compensation models unique to a given sensor architecture. In fact, since ANNs are universal function approximators, they are capable of generalizing to any sensor configuration given sufficient training. The incident angle mapping of the sensor is constructed without any a priori knowledge of the centroid-to-sun-line transformation and is developed from collected sensor data alone \cite{Rufino2009b}.

The work by Rufino et al. \cite{Rufino2009b} is reviewed as a case-study for the ANN compensation model representation. For this implementation, a network is trained on the sensor configuration of a multi-aperture mask with 100 apertures and a digital sensor array. The network architecture selected is a multilayer feed-forward ANN with a sigmoid activation function in the hidden layer and linear output neurons. The network structure is relatively simple with only one hidden layer. Two separate ANNs are used to independently compute each sun angle \cite{Rufino2009b}.

While several network input variables were compared, the spot centroid coordinates were selected due to higher accuracy over the FOV and lower computational demand. The average centroid of the sun spots is computed and used as an input to the network. In S-FOV mode, the number of spots averaged is constant, resulting in a smaller FOV, whereas in X-FOV mode, the number of spots averaged is variable, thereby increasing the effective FOV. Nevertheless, the average accuracy is worse in S-FOV mode, especially in larger off-boresight angles, since a larger input space is mapped \cite{Rufino2009b}. An overview of the ANN body of literature is presented in Table \ref{tab:mrlit}.

\subsubsection*{Deep neural network}\label{model:dnn}

The deep neural network (DNN) based model representation shares all the properties of the previously presented ANN model, but with with many hidden layers. The work by Sozen et al. \cite{Sozen2021,Soken2023} is reviewed as a case-study for the DNN compensation model representation. In this study, the DNN model is trained to correct errors induced from Earth albedo effects on an analog sensor. The primary advantage of the DNN implementation is the lack of need for an underlying albedo model to achieve error compensation. The network is trained on voltage measurements from the analog sensor, the Sun reference direction, and the satellite attitude \cite{Soken2023}.

Training can be performed either on synthetic data generated on the ground or using in-orbit data from another onboard sensor. The DNN architecture includes four input nodes corresponding to the photodiode voltage readings, two output nodes representing the two-axis sun angle errors, and two hidden layers. The network uses a log-sigmoid activation function in the input layer, a hyperbolic tangent sigmoid in the hidden layers, and a linear activation function in the output layer \cite{Soken2023}. An overview of relevant DNN literature is provided in Table \ref{tab:mrlit}. \phantomsection\label{rqsum:rqsum2} 

\begin{description}
\item[RQ 2 Summary] \textit{Geometric and non-physical model representations are the most commonly implemented in the literature, however these models are inflexible and often tied to a specific architecture. Multiplexing models improve performance through coding rules, however the mapping requires a complex mask pattern. Physics-informed models are able to account for more uncertainties, however they require a deep knowledge of error behaviors. Neural network techniques are a promising direction in both model flexibility and uncertainty capture.} (See \hyperref[rq:rq2]{RQ 2})
\end{description}

\begingroup
\renewcommand{\arraystretch}{1.5} 
\begin{table*}[htb]
\caption{Overview of model representation literature.}
\label{tab:mrlit}
\scriptsize
\centering
\newlength\qmr
\setlength\qmr{\dimexpr .077\textwidth -2\tabcolsep}
\noindent\begin{tabular}{>{\rowmac}p{12mm}>{\rowmac}p{5.75mm}>{\rowmac}p{12mm}>{\rowmac}p{5.75mm}>{\rowmac}p{12mm}>{\rowmac}p{5.75mm}>{\rowmac}p{12mm}>{\rowmac}p{5.75mm}>{\rowmac}p{12mm}>{\rowmac}p{5.75mm}>{\rowmac}p{12mm}>{\rowmac}p{5.75mm}>{\rowmac}p{12mm}<{\clearrow}}
\Xhline{1.25pt}
\multicolumn{1}{l}{\setrow{\bfseries}LUTs} & \multicolumn{2}{l}{\setrow{\bfseries}Non-Physical} & \multicolumn{4}{l}{\setrow{\bfseries}Geometric} & \multicolumn{2}{l}{\setrow{\bfseries}Physics-Informed} & \multicolumn{2}{l}{\setrow{\bfseries}Multiplexing}  & \multicolumn{2}{l}{\setrow{\bfseries}Neural Network}\setrow{}\\
\cmidrule(lr){1-1}\cmidrule(lr){2-3}\cmidrule(lr){4-7}\cmidrule(lr){8-9}\cmidrule(lr){10-11}\cmidrule(lr){12-13}
\cite{Bolshakov2020,He2013,Maqsood2010,Frezza2022,Ortega2010,Delgado2010,Delgado2012b,Boslooper2017,Delgado2011,Delgado2012a,Ortega2009,Alvi2014,Delgado2013,Strietzel2002,Pentke2022} & Poly-nomial & \cite{Chang2006,Rodrigues2006,Pentke2022,Kim2005,Zhang2023,deBoom2003,Pita2014,Li2012,Wu2001,Wu2002,You2011,Pelemeshko2020,He2013,Faizullin2017,Merchan2021,Hermoso2022a,Merchan2023,Hermoso2022b,Hermoso2021} & SPM & \cite{Boslooper2017,Yousefian2017,deBoer2013,Xie2014,Farian2018,Jing2016,Chum2003,Bardallo2017b,Arshad2018,Xie2011,Xie2012,Farian2022,Bahrampoor2018,Rhee2012,ONeill2020,Maqsood2010,Chang2008,Chang2007a,Diriker2021,Xie2013,deBoom2006,Xie2010,Farian2015,deBoer2017,Miao2017,Bardallo2017a,Lin2016,Leijtens2020} & N-Slit & \cite{Maji2015,Kohli2019,Wei2011,Post2013} & \hyperref[model:optic]{OCM} & \cite{Delgado2012b,Enright2007a,He2013,Li2012,Delgado2011,Ortega2009,Delgado2013,You2011,Xing2008,Fan2015,Maji2015,Kohli2019,Delgado2012a,Wei2011,Fan2016,Hales2002} & Peri-odic & \cite{Chen2006,Fan2013,Tsuno2019,Kondo2024} & ANN & \cite{Rufino2005,Rufino2004a,Rufino2012,Rufino2017,Sozen2021,Rufino2009b,Rufino2008,Rufino2009a}\\
& Linear & \cite{Hermoso2022a,Faizullin2017} & LSQ & \cite{Rufino2017,Buonocore2005} & Cam-era & \cite{Saleem2017,Saleem2019,TrebiOllennu2001,Mobasser2001,Mobasser2003,Liebe2001,Liebe2002,Liebe2004,Liu2016} & \hyperref[model:elec]{ECM} & \cite{Yousefian2017,Sozen2021,Faizullin2017,Sun2023,Soken2023,OKeefe2017,Wu2001,Wu2002,Springmann2014,Mafi2023,Chen2007} & Code  & \cite{Wei2017,Wei2013,Zhang2021,Wei2014,Wang2015,Zhang2022a} & DNN & \cite{Sun2023,Soken2023}\\
& Trig & \cite{Richie2015} & QPD & \cite{Faizullin2017} & Basic & \cite{Allgeier2009} & \hyperref[model:enviro]{SECM} & \cite{Antonello2018,Lu2017a} & & & &\\
& Fourier & \cite{Pentke2022} & Slit & \cite{Chang2006,He2013,KeQiang2020,Abhilash2014,Ali2011} & Solar panel & \cite{Nurgizat2021,Lu2017b} & \hyperref[model:assem]{ACM} & \cite{Radan2019,Fan2017,Wu2001,Wu2002,Duan2006,Mafi2023} & & & &\\
& Sig-moid & \cite{Pentke2022} & Multi-Slit & \cite{Fan2013,Enright2007a,Bolshakov2020} & Pyra-midal & \cite{Zhang2023,Mafi2023,Zhuo2020,Wang2019,Yousefian2017,Pita2014} & \hyperref[model:interf]{ICM} & \cite{Sozen2021,Soken2023,Frezza2022,Chang2006,Antonello2018} & & & &\\
& & & V-Slit & \cite{Fan2015,Fan2016} & Pano-ramic & \cite{Barnes2014,Zhang2022b} & & & & & &\\
\Xhline{1.25pt}
\end{tabular}
\begin{flushleft}
Abbreviations: ACM, assembly compensation model; ANN, artificial neural network; DNN, deep neural network; ECM, electrical compensation model; ICM, interference compensation model; LUT, look up table; LSQ, least squares; OCM, optical compensation model; QPD, quadrant photodiode; SECM, space environment compensation model; SPM, standard projection model.
\end{flushleft}
\end{table*}
\endgroup

\section{Feature extraction techniques} \label{sec:fetech}

Feature extraction is the first step of the sun sensor state estimation process. The primary goal of feature extraction is to unambiguously detect features in the image space for later mapping to the incident sun angle. This process is done through the capture and processing of features such as sun image centroids, edge detection, pattern matching, and other features of interest. In general, the more unique features that can be extracted from an image, the higher the accuracy of the final state estimate.

The feature extraction processes presented here are classified into four primary techniques: photodiode processing, centroid detection, encoded methods, and parametric approaches. These methods can be further generalized into two classes: conventional and parametric methods \cite{Enright2006b}. Conventional methods (e.g. photodiode processing, centroid detection, encoded) rarely use information about the expected shape of the projected light spot. However, parametric methods derive advantage from a priori knowledge of the expected projected light spot shape \cite{Enright2006b}.

In the following sections, each of the aforementioned feature extraction techniques is briefly introduced. A case study is provided for each technique, along with the corresponding pseudocode. The advantages of each method are evaluated, and the section concludes with an overview of the feature extraction literature in Table \ref{tab:felit} and a comparative analysis of the techniques in Table \ref{tab:fesummary}.

\subsection{Photodiode processing}

Photodiode processing is a feature extraction technique that uses the analog sun sensor’s measured response to incident sunlight as its primary feature. It is the second most commonly implemented feature extraction technique, and second only to the centroid detection method. This approach includes two main classifications: the direct and voltage balance techniques.

\subsubsection*{Direct}\label{fe:direct}

Direct processing is the simplest of the feature extraction techniques. The process involves the collection of incident light from a single photodiode into a current, which is then converted to a voltage with a transimpedance amplifier (TIA) circuit. The incident light on the photodiode is proportional to the projected irradiance on the detector. The voltage signal is used as a feature representation for each associated angle. An overview of the direct processing body of literature is presented in Table \ref{tab:felit}.

\subsubsection*{Voltage balance technique}\label{fe:vb}

For this method a quadrant photodiode array detector collects incident light and converts it to a current. The current is then converted to a voltage with a transimpedance amplifier (TIA) circuit. The voltages are measured from four photosensitive elements of the quadrant photodiode array and processed through a voltage balance for feature extraction. The four output signals are directly proportional to the light falling on each quadrant of the detector array.

A normalization relation is then implemented to process the quadrant signals and the incidence solar angles. The voltage balance technique approximates the center of the projected light spot on the detector plane through the balance of voltages over the array as the primary feature. An overview of the voltage balance body of literature is presented in Table \ref{tab:felit}.

The work by Boslooper et al. \cite{Boslooper2017} is reviewed as a case-study for the voltage balance calibration approach. In the following study, the four photodiode currents ($I_{1},I_{2},I_{3},I_{4}$) are converted to voltages ($Q_{1},Q_{2},Q_{3},Q_{4}$) via a TIA circuit. The feature representations are then extracted by the balancing of the four detector array voltages through the normalization functions $S_{A}$ for the sensor x-axis and $S_{B}$ for the sensor y-axis. These two normalization functions are proportional to the angular state estimates and the maximum solar aspects angles. The solar max angles are a function of the sensor geometry, including the aperture diameter $d$ and focal length $h$. The pseudocode for the technique is presented in Algorithm \ref{alg:VB}.

\begin{algorithm}[h]
    \caption{Voltage Balance \cite{Boslooper2017}}
    \label{alg:VB}
    \SetAlgoLined
    \SetKwInOut{Input}{input}
    \SetKwInOut{Output}{output}
    \DontPrintSemicolon

    \Input{Photodiode currents $I_{1},I_{2},I_{3},I_{4}$.}
    
    \Output{Normalization functions $(S_{A},S_{B})$.}

    Convert photodiode current to voltage with transimpedance amplifier (TIA) $Q_{1},Q_{2},Q_{3},Q_{4}\gets I_{1},I_{2},I_{3},I_{4}$

    // Calculate x-axis normalization function\\
    
    $\displaystyle
    S_{A}=\frac{Q_{2}+Q_{3}-Q_{1}-Q_{4}}{Q_{1}+Q_{2}+Q_{3}+Q_{4}}=\frac{\tan\left ( \alpha \right )}{\tan\left ( \alpha_{MAX} \right )}
    $\\

    // Calculate y-axis normalization function\\
    
    $\displaystyle
    S_{B}=\frac{Q_{1}+Q_{2}-Q_{3}-Q_{4}}{Q_{1}+Q_{2}+Q_{3}+Q_{4}}=\frac{\tan\left ( \beta \right )}{\tan\left ( \beta_{MAX} \right )}
    $\\

    where, $\alpha_{MAX}=\beta_{MAX}=\frac{d}{2h}$
    
    \KwRet{$(S_{A},S_{B})$}

\end{algorithm}

\subsection{Centroid detection}

 The centroid detection method is the most common feature extraction technique in sun sensor image processing. It has been extensively studied, resulting in a vast number of unique implementations. Typically, a digital sensor array captures image data from the projected light spot. These methods generally share a common premise: processing the pixel positions and intensity values from the detector to determine the centroid coordinates. 

\subsubsection*{Peak Detection (PD)} \label{fe:peak}

The peak detection method is the baseline algorithm of the centroid detection methods. The main goal of the technique is to search for the brightest pixel in the image space as the feature representation. While the method is computationally simple, it is useful to quickly compute the centroid of the image along a single profile.

Two algorithmic approaches to the peak detection techniques exist and include single peak and multiple peaks. The single peak method simply searches for the maximum intensity over the image space, while the multiple peaks method searches for local maximum intensities associated with a known pattern of peak offsets. An overview of the peak detection body of literature is presented in Table \ref{tab:felit}.

The work by Enright et al. \cite{Enright2006b} is reviewed as a case-study for the peak detection approach with a single peak in the image space. In their study, the captured image is imported and the associated pixel matrix $I=I[n]$ is generated . The processed image space is then converted into grayscale and a known pixel spacing is established $\Delta X$.

The algorithm then searches for the value of the pixel with the maximum intensity $n_{peak}$ over the full image space $n=-\frac{N}{2},\cdots ,\frac{N}{2}$. Thereafter, the centroid location $\tilde{\tau}$ is calculated from the known pixel spacing. The pseudocode for the technique is presented in Algorithm \ref{alg:PDsp} \cite{Enright2006b}.

\begin{algorithm}[h]
    \caption{Peak Detection (Single Peak) \cite{Enright2006b}, \texttt{PD-SP}}
    \label{alg:PDsp}
    \SetAlgoLined
    \SetKwInOut{Input}{input}
    \SetKwInOut{Output}{output}
    \DontPrintSemicolon

    \Input{Captured Image.}
    
    \Output{Peak Coordinate $\tilde{\tau}$.}
    
    // Initialization

    \quad \quad \quad Import image and generate pixel matrix, $I=I[n]$ 

    \quad \quad \quad Convert to grayscale
    
    \quad \quad \quad Pixel spacing, $\Delta X$

    $\displaystyle
    n_{peak}=\argmaxA_n (I[n]),\quad n=-\frac{N}{2},\cdots ,\frac{N}{2}
    $\\

    $\displaystyle
        \tilde{\tau}=n_{peak}\Delta X
    $\\
    
    \KwRet{$\tilde{\tau}$}
    
\end{algorithm}

\begin{algorithm}[h]
    \caption{Peak Detection (Multiple Peaks) \cite{Enright2006b}, \texttt{PD-MP}}
    \label{alg:PDmp}
    \SetAlgoLined
    \SetKwInOut{Input}{input}
    \SetKwInOut{Output}{output}
    \DontPrintSemicolon

    \Input{Captured Image.}
    
    \Output{Peak Coordinate $\tilde{\tau}$.}
    
    // Initialization

    \quad \quad \quad Import image and generate pixel matrix, $I=I[n]$ 

    \quad \quad \quad Convert to grayscale
    
    \quad \quad \quad Pixel spacing, $\Delta X$

    \quad \quad \quad Number of peaks, $N_{P}$

    Describe pattern by vector of peak offsets as
    $D=\{d_{1},d_{2},\cdots,d_{N_{P}}\}$

    // Locate the pattern in the image as maxi-min

        $\displaystyle
        n_{peak}=\argmaxA_n \left (  \argminA_i (I[n+d_{i}])\right ),\linebreak n=-\frac{N}{2},\cdots ,\frac{N}{2} \text{ and } \quad i=1,\cdots ,N_{P}
        $\\

    $\displaystyle
        \tilde{\tau}=n_{peak}\Delta X
    $\\
    
    \KwRet{$\tilde{\tau}$}
    
\end{algorithm}

Next, the work by Enright et al. \cite{Enright2006b} is reviewed as a case-study for the peak detection approach with multiple peaks in the image space. In their study, the captured image is imported and the associated pixel matrix $I=I[n]$ is generated. The processed image space is then converted into grayscale, a known pixel spacing is established $\Delta X$, and the number of peaks $N_{P}$ is known.

The pattern of known peak offsets is described by the vector $D$ over $N_{P}$ elements. Afterward, the algorithm locates and matches the pattern in the image as a maxi-min operation. The value of the pixel with the maximum intensity $n_{peak}$ is computed over the full image space $n=-\frac{N}{2},\cdots ,\frac{N}{2}$ for each peak $i=1,\cdots ,N_{P}$. Finally, the centroid location $\tilde{\tau}$ is calculated for each peak location and the associated pixel spacing. The pseudocode for the technique is presented in Algorithm \ref{alg:PDmp} \cite{Enright2006b}.

\subsubsection*{Peak Position Estimate (PPE)} \label{fe:ppe}

The Peak Position Estimate method builds on the observation that more information can be extracted from an image than just the locations of the brightest pixels. Rather than being concentrated in a single feature, state estimate information is distributed across the image plane. As a result, the extracted data should capture the underlying structure of the image. In this method, the peaks formed by the intersections of multiple apertures are expected to follow a known pattern. This prior knowledge of the image structure enhances the understanding of the feature space and improves the accuracy of the final centroid estimate \cite{Enright2007a}. 

The method requires a specific mask configuration of multiple cross slits to operate. The intersections of the projected sun images are matched to a known pattern during processing. The extraction process is complicated by the need to find distinct peaks, which can be difficult when the peaks overlap or are very close to each other . Therefore, the algorithm’s extraction process is divided into distinct cases based on the location of the peak pattern. An overview of the peak detection body of literature is presented in Table \ref{tab:felit}.

The work by Enright et al. \cite{Enright2007a} is reviewed as a case-study for the Peak Position Estimate approach. In their study, the captured image is imported and the associated pixel matrix $I_{s}\left [ n \right ]$ is generated . The processed image space is then converted into grayscale. The known pixel spacing $\Delta X$ and peak spacing $L_{x}, L_{y}$ is initialized. Finally, the exit conditions are established as the tolerance $tol$ and maximum number of iterations $n_{max}$. The iteration count is then initialized as $n = 0$.

First, the analytical model of sensor illumination is represented with $I_{mdl_{1}}(u)$. The image model is the ratio of polynomials, chosen for its speed in embedded application. The coefficients are chosen based on a fit to the peaks in sample images. Other possible selections for $I_{mdl_{1}}(u)$ include a diffraction model from physical principals and a simple Gaussian model.

The net illumination at the detector is then calculated as $I_{mdl_{1}}(u)$. The functions $A$, $B$, $C$, and $D$ are introduced subsequently. The model includes six parameters to be estimations, represented as $\boldsymbol{\lambda} = \vect{a_{1}, a_{2}, b_{1}, b_{2}, \tilde{m}_{x_{1}}, \tilde{m}_{y_{1}}}^\top$, comprising four amplitude values and two displacement terms.

The algorithm continues with a typical NLSQ implementation. The observed error $d\beta_{n}$ is defined as the error between the real captured image $I_{s}\left [ n \right ]$ and the modeled image $I_{mdl}(\tau)$. Next, the matrix $\mathbf{Q}$ is formed as the partial derivatives of the image model $I_{mdl}(\tau)$ with respect to the model parameters. The component partial derivatives of $I_{mdl_{1}}$ are then defined as $E(\tau)$, $F(\tau)$, $G(\tau)$, and $H(\tau)$. Thereafter, the matrix $\mathbf{Q}$ can be written as a series of column vectors.

\SetKw{Break}{break}
\SetKw{KwRet}{return}
\begin{algorithm}[htbp]
    \caption{Peak Position Estimate \cite{Enright2007a}, \texttt{PPE}}
    \label{alg:PPE}
    \SetAlgoLined
    \SetKwInOut{Input}{input}
    \SetKwInOut{Output}{output}
    \DontPrintSemicolon
    \algnewcommand\algorithmicswitch{\textbf{switch}}
    \algnewcommand\algorithmiccase{\textbf{case}}
    \SetKwFunction{FMain}{GetQ}
    \makeatother

    \Input{Captured Image.}
    
    \Output{Centroid Coordinates $C=(\tilde{m}_{x_{1}}, \tilde{m}_{y_{1}})$.}
    
    // Initialization

    \quad \quad \quad Import image, $I_{s}\left [ n \right ]$ 

    \quad \quad \quad Convert to grayscale
    
    \quad \quad \quad Maximum iter, $n_{max}$; Initialize iter, $n = 0$

    \quad \quad \quad Tolerance, $tol$

    \quad \quad \quad Pixel spacing, $\Delta X$; Peak spacing, $L_{x}, L_{y}$

    $\displaystyle
    I_{mdl_{1}}(u)=\frac{c_{1}u^{4}+c_{2}u^{2}+c_{3}}{c_{4}u^{4}+c_{5}u^{2}+c_{6}}
    $ // Image model\\

    $\displaystyle
    I_{mdl}(\tau)=a_{1}A(\tau)+a_{2}B(\tau)+b_{1}C(\tau)+b_{2}D(\tau) \linebreak
    A=I_{mdl_{1}}(\tau-\tilde{m}_{x_{1}}) \linebreak
    B=I_{mdl_{1}}(\tau-\tilde{m}_{x_{1}}-2L_{x}/\Delta X) \linebreak
    C=I_{mdl_{1}}(\tau-\tilde{m}_{y_{1}}) \linebreak
    D=I_{mdl_{1}}(\tau-\tilde{m}_{y_{1}}-2L_{y}/\Delta X)
    $\\
        
    $\displaystyle
    \boldsymbol{\lambda} = \vect{a_{1}, a_{2}, b_{1}, b_{2}, \tilde{m}_{x_{1}}, \tilde{m}_{y_{1}}}^\top
    $\\
        
        \For{$n = 0$ \textbf{to} $n_{max}$} {

            $n \gets n+1$

            $\displaystyle
            d\beta_{n}=I_{s}\left [ n \right ]-I_{mdl}(\tau_{n}, a_{1}, a_{2}, b_{1}, b_{2}, \tilde{m}_{x_{1}}, \tilde{m}_{y_{1}})
            $\\
            
            $E(\tau)=-\diffp{}{u}(mdl_{1}(u))\Biggr|_{u=\tau-m_{x_{1}}}$
            
            $F(\tau)=-\diffp{}{u}(mdl_{1}(u))\Biggr|_{u=\tau-m_{x_{1}}-2L_{x}/\Delta X}$
            
            $G(\tau)=-\diffp{}{u}(mdl_{1}(u))\Biggr|_{u=\tau-m_{y_{1}}}$ 
            
            $H(\tau)=-\diffp{}{u}(mdl_{1}(u))\Biggr|_{u=\tau-m_{y_{1}}-2L_{y}/\Delta X}$

            \Switch{Image Type}{
            \textbf{when} $0\Longrightarrow \mathbf{Q}=\linebreak\vect{\mathbf{A},\mathbf{B},\mathbf{C},\mathbf{D}, (a_{1}\mathbf{E}+a_{2}\mathbf{F}), (b_{1}\mathbf{G}+b_{2}\mathbf{H})}$\;
            \textbf{when} $1\Longrightarrow \mathbf{Q}=\linebreak\vect{(\mathbf{A}+\mathbf{C}),\mathbf{B},\mathbf{D}, (a_{1}\mathbf{E}+a_{2}\mathbf{F}), (b_{1}\mathbf{G}+b_{2}\mathbf{H})}$\;
            \textbf{when} $2\Longrightarrow \mathbf{Q}=\linebreak\vect{\mathbf{A},(\mathbf{B}+\mathbf{C}),\mathbf{D}, (a_{1}\mathbf{E}+a_{2}\mathbf{F}), (b_{1}\mathbf{G}+b_{2}\mathbf{H})}$\;
            \textbf{when} $3\Longrightarrow \mathbf{Q}=\linebreak\vect{(\mathbf{A}+\mathbf{D}),B,C, (a_{1}\mathbf{E}+a_{2}\mathbf{F}), (b_{1}\mathbf{G}+b_{2}\mathbf{H})}$\;
            \textbf{when} $4\Longrightarrow \mathbf{Q}=\linebreak\vect{\mathbf{A},(\mathbf{B}+\mathbf{D}),\mathbf{C}, (a_{1}\mathbf{E}+a_{2}\mathbf{F}), (b_{1}\mathbf{G}+b_{2}\mathbf{H})}$\;
            }

            $\displaystyle
            \boxed{\text{Solve }\mathbf{S} d\boldsymbol{\lambda}=b} \text{, where }\linebreak
            \mathbf{S}=\mathbf{Q}^\top \mathbf{Q} \text{ and } b=\mathbf{Q}^\top d\beta
            $\\

            \If {$\left |\frac{\Delta\boldsymbol{\lambda}}{\boldsymbol{\lambda}}  \right |<tol$}{
    
                \Break
    
            }
            
        }

    \KwRet{($\tilde{m}_{x_{1}}, \tilde{m}_{y_{1}}$)}
    
\end{algorithm}

The matrix $\mathbf{Q}$ is determined based on five distinct cases, each corresponding to specific characteristics of peak behavior in the image space. These cases are implemented as a switch statement in the algorithm. Image Type 0 is defined when all four physical peaks are distinct. Image Type 1 is defined when A and C overlap. Image Type 2 is defined when B and C overlap. Image Type 3 is defined when A and D overlap. Lastly, Image Type 4 is defined when B and D overlap.

In each iteration, the parameter update $d\boldsymbol{\lambda}$ is determined by solving the linear system $\mathbf{S} d\boldsymbol{\lambda}=b$, which consists of six equations with six unknowns. The algorithm terminates either when the maximum number of iterations $n_{max}$ is reached  or when the change in $\lambda$ falls below the tolerance $tol$. The pseudocode for this technique is presented in Algorithm \ref{alg:PPE} \cite{Enright2007a}.

\subsubsection*{Basic Centroid Method (BCM)} \label{fe:bcm}

The Basic Centroid Method uses the image of the incident light on the detector without any pre-processing or data manipulation to extract features. The intensity of the light across the image space is calculated as a weighted average to find the centroid feature. This is the simplest algorithmic form under the centroid method class \cite{Chang2008}.

The strength of this method lies in its ease of implementation and the lack of any pre-processing requirements. Furthermore, it is capable of achieving sub-pixel resolution. However, its accuracy is generally lower than that of other centroid-based methods, primarily due to the lack of pre-processing filters. This limitation reduces its ability to compensate for noise, making it more susceptible to noise-induced errors. An overview of the Basic Centroid Method body of literature is presented in Table \ref{tab:felit}.

The work by Chang et al. \cite{Chang2008} is reviewed as a case-study for the Basic Centroid Method approach. In their study, the captured image is imported and the associated pixel matrix $I(x,y)$ is generated. The processed image space is then converted into grayscale. The number of pixel rows and columns are then established.

The centroid algorithm proceeds with the calculation of the total intensity $I_{Tot}$ over the image space as the sum over each pixel. The centroid coordinates ($x_{c},y_{c}$) for each axis are then calculated as the weighted average over the image space. The pseudocode for the technique is presented in Algorithm \ref{alg:BCM} \cite{Chang2008}.

\begin{algorithm}[h]
    \caption{Basic Centroid Method \cite{Chang2008}, \texttt{BCM}}
    \label{alg:BCM}
    \SetAlgoLined
    \SetKwInOut{Input}{input}
    \SetKwInOut{Output}{output}
    \DontPrintSemicolon

    \Input{Captured Image.}
    
    \Output{Centroid Coordinates $C=(x_{c},y_{c})$.}
    
    // Initialization

    \quad \quad \quad Import image and generate pixel matrix, $I=I(x,y)$ 

    \quad \quad \quad Convert to grayscale
    
    \quad \quad \quad Set number of pixel matrix rows, $N_{r}$

    \quad \quad \quad Set number of pixel matrix columns, $N_{c}$
    
    \ForEach{pixel \textbf{I}$(x,y)$} {

        $\displaystyle
        I_{Tot}=\sum_{i=1}^{N_{r}}\sum_{j=1}^{N_{c}}I\left ( i,j \right )
        $\\
        
        $\displaystyle
        x_{c}=\frac{1}{I_{Tot}}\sum_{i=1}^{N_{r}}\sum_{j=1}^{N_{c}}x_{i}I\left ( i,j \right )
        $\\

        $\displaystyle
        y_{c}=\frac{1}{I_{Tot}}\sum_{i=1}^{N_{r}}\sum_{j=1}^{N_{c}}y_{j}I\left ( i,j \right )
        $\\

    }
    \KwRet{($x_{c},y_{c}$)}
    
\end{algorithm}

\subsubsection*{Basic Centroid Threshold Method (BCTM)} \label{fe:bctm}

The Basic Centroid Threshold Method builds on top of the foundational Basic Centroiding Method to improve the weakness to noise by adding pre-processing filtering to the algorithm. The algorithm works by establishing a reference threshold level and applying it to filter the intensity of the image space. 

Pixel values that are above the threshold level are preserved or set to a fixed value, while those that are below are set to zero. If the image space contains noise, then a threshold level is selected to mitigate the noise error. Therefore, the reference threshold level affects the accuracy of the centroid method. However, it is difficult to select an optimal threshold value that minimizes the noise in the image space \cite{He2005}. 

The main advantages of the method are the addition of the image filtering to reduce noise and the improvement in feature extraction accuracy. 
Although incorporating pre-processing slightly increases the algorithm’s complexity and latency, it significantly enhances estimation accuracy. As a result, this approach is the most commonly used centroid method in practice, offering an good balance between accuracy and simplicity. 

A notable drawback of the method, however, is its significant reduction in accuracy when the noise level exceeds a certain threshold. An overview of the Basic Centroid Threshold Method body of literature is presented in Table \ref{tab:felit}. 

The work by He et al. \cite{He2005} is reviewed as a case-study for the Basic Centroid Threshold Method approach. In their study, the captured image is imported and the associated pixel matrix $I(x,y)$ is generated. The processed image space is then converted into grayscale. The number of pixel rows and columns are then established. Finally, the threshold value $\mu$ is set.

The image space is first pre-processed by applying the established threshold over each pixel. Each pixel is subtracted by a threshold fraction of the maximum intensity, where any values lower than the threshold are processed as zero. The centroid algorithm then proceeds identically to the Basic Centroid Method.

Therefore, the total intensity $I_{Tot}$ is first calculated over the image space as the sum over each pixel. The centroid coordinates ($x_{c},y_{c}$) for each axis are then calculated as the weighted average over the image space. The pseudocode for the technique is presented in Algorithm \ref{alg:BCTM} \cite{He2005}.

\begin{algorithm}[h]
    \caption{Basic Centroid Threshold Method \cite{He2005}, \texttt{BCTM}}
    \label{alg:BCTM}
    \SetAlgoLined
    \SetKwInOut{Input}{input}
    \SetKwInOut{Output}{output}
    \SetKwProg{GetCentroid}{GetCentroid}{}
    \DontPrintSemicolon

    \Input{Captured Image.}
    
    \Output{Centroid Coordinates $C=(x_{c},y_{c})$.}
    
    // Initialization

    \quad \quad \quad Import image and generate pixel matrix, $I=I(x,y)$ 

    \quad \quad \quad Convert to grayscale
    
    \quad \quad \quad Set number of pixel matrix rows, $N_{r}$

    \quad \quad \quad Set number of pixel matrix columns, $N_{c}$
    
    \quad \quad \quad Set threshold value, $\mu$

    Apply threshold to image as $I^{\prime} = I - \mu I_{max}$, where $I > 0$

    All data lower than the threshold is processed as 0
    
    \ForEach{pixel \textbf{I}$(x,y)$} {

        $\displaystyle
        I_{Tot}=\sum_{i=1}^{N_{r}}\sum_{j=1}^{N_{c}}I^{\prime}\left ( i,j \right )
        $\\
        
        $\displaystyle
        x_{c}=\frac{1}{I_{Tot}}\sum_{i=1}^{N_{r}}\sum_{j=1}^{N_{c}}x_{i}I^{\prime}\left ( i,j \right )
        $\\

        $\displaystyle
        y_{c}=\frac{1}{I_{Tot}}\sum_{i=1}^{N_{r}}\sum_{j=1}^{N_{c}}y_{j}I^{\prime}\left ( i,j \right )
        $\\

    }
    \KwRet{($x_{c},y_{c}$)}

\end{algorithm}

\subsubsection*{Multiple Centroid Averaging Method (MCAM)} \label{fe:mcam}

The Multiple Centroid Averaging Method is an extension of the Basic Centroid Threshold Method for multiple pinhole aperture masks. Specifically, the mask consists of multiple small holes arranged in an array pattern. The algorithm uses multiple spot images as extracted features to improve the centroid estimation process. In particular, the technique of averaging multiple sun images is used to improve the accuracy and precision of the feature extraction estimate.

The core principle of MCAM is the simultaneous capture of multiple sun images on the image plane, each of which is used to compute a separate centroid measurement. By averaging these multiple measurements, a more accurate estimate of the centroid features is achieved. This approach effectively reduces measurement error through averaging, resulting in improved precision and reliability compared to the Basic Centroid Thresholding Method. 

The random noise components in each single image are filtered out by averaging over multiple sun images. Assuming the image noise contributions are uncorrelated, the increase to the extraction precision is a function of the square root of the number of projected sun features used in the averaging operation. The increase in precision enables ranges from very-fine to ultra-fine performance to be achieved. In addition, the increase in number of apertures allows for more reliable feature extraction, since a variable number of spots can be processed \cite{Rufino2009b}.

However, these improvements come with some limitations, including additional latency to process multiple spots and a reduced sensor FOV due to more image area being taken by the multiple sun spots. 
This limitation can be mitigated by implementing FOV segmentation across aperture zones. An overview of the Multiple Centroid Averaging Method body of literature is presented in Table \ref{tab:felit}.

The work by Rufino et al. \cite{Rufino2009b} is reviewed as a case-study for the Multiple Centroid Averaging Method approach. In their study, the captured image is imported and the associated pixel matrix $I(x,y)$ is generated. The processed image space is then converted into grayscale. The number of pixel rows and columns are established. Thereafter, the number of expected apertures and threshold value $\mu$ are set. Finally, the average centroid coordinates are initialized.

The image space is pre-processed by applying the established threshold over each pixel. Each pixel is subtracted by a threshold fraction of the maximum intensity, where any values lower than the threshold are processed as zero. The centroid algorithm then proceeds by iterating over each aperture.

In each iteration, the n-th aperture is segmented as $I_{n}(x,y)$, and processed by calculating the centroid of the associated sun image as the weighted sum from the Basic Centroid Threshold Method. Each centroid coordinate estimate is summed over the iteration loop until all of the apertures have been processed. The final averaged centroid coordinates ($x_{avg},y_{avg}$) are then calculated as the sum of all the centroid estimates divided by the total number of apertures $N_{aper}$. The pseudocode for the technique is presented in Algorithm \ref{alg:MCAM} \cite{Rufino2009b}.

\begin{algorithm}[h]
    \caption{Multiple Centroid Averaging Method \cite{Rufino2009b}, \texttt{MCAM}}
    \label{alg:MCAM}
    \SetAlgoLined
    \SetKwInOut{Input}{input}
    \SetKwInOut{Output}{output}
    \SetKwProg{GetCentroid}{GetCentroid}{}
    \DontPrintSemicolon

    \Input{Captured Image.}
    
    \Output{Centroid Coordinates $C=(x_{avg},y_{avg})$.}
    
    // Initialization

    \quad \quad \quad Import image and generate pixel matrix, $I=I(x,y)$ 

    \quad \quad \quad Convert to grayscale
    
    \quad \quad \quad Set number of pixel matrix rows, $N_{r}$

    \quad \quad \quad Set number of pixel matrix columns, $N_{c}$

    \quad \quad \quad Set number of apertures, $N_{aper}$
    
    \quad \quad \quad Set threshold value, $\mu$

    \quad \quad \quad Initialize $x_{avg} = 0$ and $y_{avg} = 0$

    Apply threshold to image as $I_{n}^{\prime} = I_{n} - \mu I_{max}$, where $I_{n} > 0$

    All data lower than the threshold is processed as 0
    
    \For{$n=1$ \KwTo $N_{aper}$} {

        Segment the $n^{\text{th}}$ aperture image as $I_{n}(x,y)$ 

        \ForEach{pixel $I_{n}(x,y)$} {

        $\displaystyle
        I_{nTot}=\sum_{i=1}^{N_{r}}\sum_{j=1}^{N_{c}}I_{n}^{\prime}\left ( i,j \right )
        $\\
        
        $\displaystyle
        x_{n}=\frac{1}{I_{nTot}}\sum_{i=1}^{N_{r}}\sum_{j=1}^{N_{c}}x_{i}I_{n}^{\prime}\left ( i,j \right )
        $\\

        $\displaystyle
        y_{n}=\frac{1}{I_{nTot}}\sum_{i=1}^{N_{r}}\sum_{j=1}^{N_{c}}y_{j}I_{n}^{\prime}\left ( i,j \right )
        $\\

        }

        Sum each aperture centroid \linebreak $x_{avg} \gets x_{avg} + x_{n}$ and $y_{avg} \gets y_{avg} + y_{n}$

    }

    Calculate the average aperture centroid \linebreak $x_{avg} \gets x_{avg} / N_{aper}$ and $y_{avg} \gets y_{avg} / N_{aper}$

    \KwRet{($x_{avg},y_{avg}$)}
    
\end{algorithm}

\subsubsection*{Double Balance Centroid Method (DBCM)} \label{fe:dbcm}

The Double Balance Centroid Method is a two-step centroid detection algorithm with an optimized acquisition-tracking readout method. Instead of requiring threshold processing to achieve super-resolution, the higher accuracy is obtained through simultaneously balancing multiple pixel areas. Hence, double balance refers to the use of two sub-windows to achieve the feature extraction process. The first step, acquisition mode, is used to coarsely establish an ROI around the sun image. Next, in the Sun tracking mode, the sun spot is tracked and then the centroid is calculated over successive measurements.

During the first step, the ROI is determined through the WTA segmentation technique. This provides a coarse window around the sun image for later processing. Once the ROI is established, the sun image is tracked within the window and the centroid is calculated as a double balance of two sub-windows. The first window is compared to the second window, which is shifted by one column. This information is used to calculate the centroid by balancing four known pixel areas.

The primary advantage of the method is in its ability to achieve both lower latency and power consumption by dividing the centroid algorithm into two optimized steps for acquisition and tracking. The WTA acquisition mode enables a fast readout method of the ROI without requiring additional circuitry or processing. Furthermore, the tracking mode enables sub-pixel resolution through the balancing of multiple pixel areas over the ROI \cite{Xie2014}. An overview of the Double Balance Centroid Method body of literature is presented in Table \ref{tab:felit}.

The work by Xie et al. \cite{Xie2014} is reviewed as a case-study for the Double Balance Centroid Method approach. In their study, the captured image is imported and the associated pixel matrix $I(x,y)$ is generated. The processed image space is then converted into grayscale. The number of pixel rows and columns are established. Lastly, the threshold value $\mu$ is set.

In the next phase, the acquisition mode of the algorithm is implemented. The image space is first pre-processed by applying the established threshold over each pixel. Each pixel is subtracted by a threshold fraction of the maximum intensity, where any values lower than the threshold are processed as zero.

The WTA process is then applied to each column and row profile. Here, a vector is generated of the most intensely illuminated pixels along the column and row. This is accomplished as a maximum operation over the image plane. The ROI centroid coordinates $(x_{ROI},y_{ROI})$ are computed as a window around the brightest pixel location. Finally, the ROI sub-array $I_{ROI}(x,y)$ is generated as a $21 \times 21$ pixel region, where the ROI centroid is bounded by a 10 pixel margin.

The second phase of the algorithm initiates the Sun tracking mode. In this mode the sum of the pixel areas in the ROI sub-array are computed for all five regions (A-E). Area A is defined as a 1x21 pixel region that resides in the second column of the image space. Likewise, area E is a 1x21 pixel region that resides at the end of the image column space in the 22nd column. Regions A and E are assumed to be non-illuminated and equal under uniform lighting conditions.

Area C is a 1x21 pixel region at the center of the image space in the 12th column. The areas B and D are 9x21 pixel regions that reside between areas A-C and C-E, respectively. The centroid is found by shifting two sub-windows in the ROI by one column space. Finally, the centroid coordinates ($x_{c},y_{c}$) are calculated by balancing the previously calculated summed regions. The pseudocode for the Double Balance Centroid Method, including acquisition and the Sun tracking modes, is presented in Algorithm \ref{alg:DBCM}.

\begin{algorithm}[htbp]
    \caption{Double Balance Centroid Method \cite{Xie2014}, \texttt{DBCM}}
    \label{alg:DBCM}
    \SetAlgoLined
    \SetKwInOut{Input}{input}
    \SetKwInOut{Output}{output}
    \SetKwProg{GetCentroid}{GetCentroid}{}
    \DontPrintSemicolon

    \Input{Captured Image.}
    
    \Output{Centroid Coordinates $C=(x_{c},y_{c})$.}
    
    // Initialization

    \quad \quad \quad Import image and generate pixel matrix, $I=I(x,y)$ 

    \quad \quad \quad Convert to grayscale
    
    \quad \quad \quad Set number of pixel matrix rows, $N_{r}$

    \quad \quad \quad Set number of pixel matrix columns, $N_{c}$
    
    \quad \quad \quad Set threshold value, $\mu$

    // Step 1: Acquisition mode

    Apply threshold to image as $I^{\prime} = I - \mu I_{max}$, where $I > 0$

    All data lower than the threshold is processed as 0

    // Apply Winner-Takes-It-All (WTA) algorithm

    \For{$j=1$ \KwTo $N_{c}$} {
        
        Generate a vector of the most heavily illuminated row pixels with
        $\displaystyle
        r_i = \max_j I_{i,j},j=0,1,\cdots ,N_{c}
        $\\

    }

    \For{$i=1$ \KwTo $N_{r}$} {
        
        Generate a vector of the most heavily illuminated column pixels with
        $\displaystyle
        c_j =\max_i I_{i,j},i=0,1,\cdots ,N_{r}
        $\\

    }

    Compute the ROI coordinates $(x_{ROI},y_{ROI})$

    Determine the ROI sub-array as $I_{ROI}(x,y)$, where the sub-array
    is a $21 \times 21$ pixel region (with centroid bounded by a 10 pixel margin)

    // Step 2: Sun tracking mode

    // Calculate the sum of the pixels for region (A-E)

    $\displaystyle
    S_{A}=\sum_{\text{area A}}I_{ROI}(x,y)
    $, where $A = (1 \times 21 \text{ pixel})$
    
    $\displaystyle
    S_{B}=\sum_{\text{area B}}I_{ROI}(x,y)
    $, where $B = (9 \times 21 \text{ pixel})$

    $\displaystyle
    S_{C}=\sum_{\text{area C}}I_{ROI}(x,y)
    $, where $C = (1 \times 21 \text{ pixel})$

    $\displaystyle
    S_{D}=\sum_{\text{area D}}I_{ROI}(x,y)
    $, where $D = (9 \times 21 \text{ pixel})$

    $\displaystyle
    S_{E}=\sum_{\text{area E}}I_{ROI}(x,y)
    $, where $E = (1 \times 21 \text{ pixel})$\\

    Compute the centroid coordinates
    $\displaystyle
    (x_{c},y_{c})=(x_{ROI},y_{ROI})+\frac{1}{2}\frac{S_{D}-S_{B}}{S_{C}-S_{A}}
    $, assuming that $S_{A}=S_{E}$\\

    \KwRet{($x_{c},y_{c}$)}
    
\end{algorithm}

\subsubsection*{Multiple Threshold Averaging Centroid Method (MT-ACM)} \label{fe:mtacm}

The Multiple Threshold Averaging Centroid Method uses multiple thresholds to improve subpixel resolution during the feature extraction process by averaging the estimates. This method is a direct improvement to the Basic Centroid Threshold Method, however it is more computationally expensive. The estimation accuracy is proportional to the number of binary images that are sampled and averaged \cite{Massari2004}.

The process is applied to the x and y axes using the superposition principle to reconstruct the light intensity profiles. The light profile is assumed to be Gaussian and a couple of symmetric points are found as the binary image bounds. Next, a mean value is calculated for each threshold, which is later used to find the average final estimate for each profile. The primary advantage of this method is the improvement to the subpixel accuracy. The increase in the computational load and algorithmic complexity is a drawback to this approach. An overview of the Multiple Threshold Averaging Centroid Method body of literature is presented in Table \ref{tab:felit}.

The work by Massari et al. \cite{Massari2004} is reviewed as a case-study for the Multiple Threshold Averaging Centroid Method. In their study, the captured image is imported and the associated pixel matrix $I(x,y)$ is generated. The processed image space is then converted into grayscale. Next, the number of pixel rows and columns are established and the pixel pitch $\Delta X$ is set. Lastly, the minimum number of thresholds required $N_{TH}$ to achieve the position detection accuracy $\Delta x_{m}$ is found.

The algorithm proceeds with calculating the light intensity profiles. The profile of the row image space is found by generating a vector of the sum of the row intensities. Likewise, the profile of the column image space is found by generating a vector of the sum of the column intensities. The pseudocode is presented in Algorithm \ref{alg:MTACM1} and continued in Algorithm \ref{alg:MTACM2}.

\begin{algorithm}[htbp]
    \caption{Multiple Threshold Averaging Centroid Method \cite{Massari2004}, \texttt{MT-ACM}}
    \label{alg:MTACM1}
    \SetAlgoLined
    \SetKwInOut{Input}{input}
    \SetKwInOut{Output}{output}
    \SetKwProg{GetCentroid}{GetCentroid}{}
    \DontPrintSemicolon

    \Input{Captured Image.}
    
    \Output{Centroid Coordinates $C=(x_{c},y_{c})$.}
    
    // Initialization

    \quad \quad \quad Import image and generate pixel matrix, $I=I(x,y)$ 

    \quad \quad \quad Convert to grayscale
    
    \quad \quad \quad Set number of pixel matrix rows, $N_{r}$

    \quad \quad \quad Set number of pixel matrix columns, $N_{c}$

    \quad \quad \quad Set pixel pitch, $\Delta X$

    \quad \quad \quad Set the minimum number of thresholds as
        $\displaystyle
        N_{TH}>\frac{\sigma \sqrt{e}}{\Delta x_{m}}
        $, where $\Delta x_{m}$ is the position detection accuracy\\

    \For{$j=1$ \KwTo $N_{c}$} {
        
        Generate a vector of the sum of the row intensities with
        $\displaystyle
        r_{i}=\sum_{j=1}^{N_{c}}I_{i,j},j=0,1,\cdots ,N_{c}
        $\\

    }

    \For{$i=1$ \KwTo $N_{r}$} {
        
        Generate a vector of the sum of the column intensities with
        $\displaystyle
        c_{j}=\sum_{i=1}^{N_{r}}I_{i,j},i=0,1,\cdots ,N_{r}
        $\\

    }
    
\end{algorithm}

In the next phase of the algorithm each threshold $T_{Hl}$ is iterated through for further processing. At least $N_{TH}$ binary images $b_{Il}$ are generated during this phase to ensure the required accuracy $\Delta x_{m}$ is achieved. Each pixel in the row image space is evaluated and compared to the current threshold value. If the intensity is lower than the threshold then the binary image is zero, otherwise the binary image is one. This process is then repeated for the column image space.

The leftmost binary transition is defined as $h_{l}=$, whereas the rightmost binary transition is defined as $k_{l}=$. The mean value of the current threshold intensity profile for a given row $x_{ml}$ or column $y_{ml}$ is calculated using the previously obtained binary transitions and pixel pitch. The final position estimation is then calculated for a given row $x_{c}$ or column $y_{c}$ profile as the average of all the threshold mean values. The final centroid solution is returned as a row and column pair ($x_{c},y_{c}$).

\begin{algorithm}[h]
    \caption{Multiple Threshold Averaging Centroid Method \cite{Massari2004}, \texttt{MT-ACM}}
    \label{alg:MTACM2}
    \SetAlgoLined
    \SetKwInOut{Input}{input}
    \SetKwInOut{Output}{output}
    \SetKwProg{GetCentroid}{GetCentroid}{}
    \DontPrintSemicolon
    \setcounter{AlgoLine}{13}
    
    \ForEach{threshold $T_{Hl}$} {

        // Create at least $N_{TH}$ binary images $b_{Il}$

        \ForEach{pixel $r_{i}$} {
    
            \eIf {$r_{i}>T_{Hl}$}{
    
            $b_{Il,i}=1$
    
            }{

            $b_{Il,i}=0$
    
            } 

        }

        \ForEach{pixel $c_{j}$} {
    
            \eIf {$c_{j}>T_{Hl}$}{
    
            $b_{Il,j}=1$
    
            }{

            $b_{Il,j}=0$
    
            }
        
        }

        $h_{l}=$ first $0\to 1$ from left
        
        $k_{l}=$ first $0\to 1$ from right

        The current mean value is extracted using
        $\displaystyle
        x_{ml}=y_{ml}=\frac{(h_{l}+k_{l})\Delta X}{2}
        $\\

        The final position estimation is calculated as
        $\displaystyle
        x_{c}=y_{c}=\frac{1}{N}\sum_{l=1}^{N}x_{ml}
        $\\

    }

    \KwRet{($x_{c},y_{c}$)}
    
\end{algorithm}

\subsubsection*{PixelMax (PM)} \label{fe:pm}

The PixelMax method is a two-step centroid detection algorithm that is proposed to reduce the processing time during feature extraction. The algorithm is simple to implement and reduces the computation time by eliminating image pre-processing steps in the feature extraction estimation. In particular, the PixelMax algorithm was developed as a faster alternative to the BCTM \cite{Coutinho2022}. 

In the first step of the algorithm the light intensity images are stored as row and column profile vectors for further processing. Next, the largest values of the vectors are searched and set as the centroid estimates. This process allows for fast feature extraction, however it trades execution time for sub-pixel resolution. The main advantage of the Pixelmax algorithm is the performance achieved without the need for pre-process filtering, such as thresholding or image segmentation.

The PixelMax method achieves similar performance to that of the BCTM. In addition, the algorithm has lower time complexity and reduces the processing latency compared to BCTM. PixelMax also demonstrated a lower latency compared to that of DBCM and MT-ACM. An overview of the PixelMax body of literature is presented in Table \ref{tab:felit}.

The work by Coutinho et al. \cite{Coutinho2022} is reviewed as a case-study for the PixelMax method. In their study, the captured image is imported and the associated pixel matrix $I(x,y)$ is generated. The processed image space is then converted into grayscale. Next, the number of pixel rows and columns are established.

The algorithm proceeds with the generation of the light intensity profiles of the row and column image spaces. The row profile is calculated as a vector of the sum of the row intensities, while the column profile is calculated as a vector of the sum of the column intensities. The row centroid position is found by stepping through the vector of the sum of the row intensities until the maximum value is found. In addition, the column centroid position is found by stepping through the vector of the sum of the column intensities until the maximum value is found. The centroid coordinate pair is returned as ($x_{c},y_{c}$). The pseudocode is presented in Algorithm \ref{alg:PM}.

\begin{algorithm}[h]
    \caption{PixelMax \cite{Coutinho2022}, \texttt{PM}}
    \label{alg:PM}
    \SetAlgoLined
    \SetKwInOut{Input}{input}
    \SetKwInOut{Output}{output}
    \DontPrintSemicolon
    \Input{Captured Image.}
    
    \Output{Centroid Coordinates $C=(x_{c},y_{c})$.}
    
    // Initialization

    \quad \quad \quad Import image and generate pixel matrix, $I=I(x,y)$ 

    \quad \quad \quad Convert to grayscale
    
    \quad \quad \quad Set number of pixel matrix rows, $N_{r}$

    \quad \quad \quad Set number of pixel matrix columns, $N_{c}$
    
    \For{$j=1$ \KwTo $N_{c}$} {
        
        Generate a vector of the sum of the row intensities with
        $\displaystyle
        r_{i}=\sum_{j=1}^{N_{c}}I_{i,j},j=0,1,\cdots ,N_{c}
        $\\

    }

    \For{$i=1$ \KwTo $N_{r}$} {
        
        Generate a vector of the sum of the column intensities with
        $\displaystyle
        c_{j}=\sum_{i=1}^{N_{r}}I_{i,j},i=0,1,\cdots ,N_{r}
        $\\

    }

    \For{$i=1$ \KwTo $N_{r}$} {
        
        Step through the vector of the sum of the row intensities

        \If {$r_{i}=\textup{max}\left ( R \right )$}{
    
        $x_{c}=i$
    
        }

    }

    \For{$j=1$ \KwTo $N_{c}$} {
        
        Step through the vector of the sum of the column intensities

        \If {$c_{j}=\textup{max}\left ( C \right )$}{
    
        $y_{c}=j$
    
        }

    }

    return ($x_{c},y_{c}$)
    
\end{algorithm}

\subsubsection*{Event Sensor Centroid Method (ESCM)} \label{fe:escm}

The Event Sensor Centroid method is a feature extraction method developed to achieve sub-pixel accuracy on spiking luminance sensors. The approach improves the accuracy of spiking luminance sensors without increasing the complexity of the detector architecture. Temporal information is used from pixel events to achieve sub-pixel accuracy \cite{Farian2022}. 

During ESCM operation, time stamped events from the Sun sensor are processed to reconstruct the light intensity profile on the detector and compute the associated centroid position. While asynchronous sun sensors enable improved low latency performance, they have a lower spatial resolution compared to APS sun sensors due to the increased complexity of I\&F pixels. The ESCM improves the accuracy of spiking luminance sensors through the use of temporal pixel events. The proposed method is implemented in a TFS operation on a L-slit mask.

The primary strengths of the ESCM are low latency, algorithmic simplicity, and sub-pixel resolution. An overview of the ESCM body of literature is presented in Table \ref{tab:felit}.

The work by Farian et al. \cite{Farian2022} is reviewed as a case-study for the Event Sensor Centroid method. In their study, the captured image is imported and processed into a series of $N_{events}$ time-stamped events $(x_{pix},t_{x})$. The time stamp of the $RESET$ signal used by the TFS framework is first recorded. The TFS operation is implemented by a TFS counter. The TFS operation globally resets of all pixels after the first $n$ pixels have spiked.

All of the incoming events $N_{events}$ of firing pixels are accumulated before the next $RESET$. The coordinates and timing information of the accumulated active pixels are then stored for each event $N_{events}$. The address position of the firing pixel is stored as $x_{pix}$, while the corresponding time-stamp information is stored as $t_{x}$.

The centroid sub-pixel coordinates $X_{sub\_pix}$ are calculated after the new $RESET$. The centroid is calculated as a weighted mean algorithm, where multiple winning pixels are used to interpolate the peak of the incident light profile. After the centroid $X_{sub\_pix}$ is computed, the sun position can be found next. The event counter $N_{events}$ is reset to zero, and the centroid coordinates $X_{sub\_pix}$ are returned. The pseudocode is presented in Algorithm \ref{alg:ECM}.

\begin{algorithm}[h]
    \caption{Event Sensor Centroid Method \cite{Farian2022}, \texttt{ESCM}}
    \label{alg:ECM}
    \SetAlgoLined
    \SetKwInOut{Input}{input}
    \SetKwInOut{Output}{output}
    \DontPrintSemicolon
    \Input{Events of firing pixels, $N_{events}$\\
        Time-stamped events, $(x_{pix},t_{x})$
    }
    
    \Output{Centroid sub-pixel coordinates, $X_{sub\_pix}$.}

    Record time stamp, $t_{reset}$, of the $RESET$ signal used by the TFS framework
    
    Before the next $RESET$ is elicited, accumulate all the incoming events $N_{events}$ of firing pixels

    // The coordinates and timing information of the active pixels are stored on memory

    \ForEach{event $N_{event}$} {
    Store the address position of the firing pixel, $x_{pix}$
    
    Store the corresponding time-stamp, $t_{x}$
    }

    // Once the new $RESET$ is elicited, compute the centroid sub-pixel coordinates
    
    \For{$n=1$ \KwTo $N_{events}$} {
    
    $\displaystyle
    X_{sub\_pix}=\frac{\sum_{n=1}^{N_{events}}\left ( \frac{1}{t_{x}-t_{reset}}\cdot x_{pix} \right )}{\sum_{n=1}^{N_{events}}\left ( \frac{1}{t_{x}-t_{reset}} \right )}
    $\\
    }

    Once centroid $X_{sub\_pix}$ is known, calculate the sun position $\phi$

    Reset the event counter as $N_{events}=0$
    
    \KwRet$X_{sub\_pix}$
    
\end{algorithm}

\subsubsection*{Black Sun Centroid Method (BSCM)} \label{fe:bscm}

The Black Sun Centroid Method is a feature extraction method that uses the black sun effect. The black sun effect is caused by CMOS image pixel oversaturation, in which electron overspill occurs. While the black sun effect is usually an undesirable phenomenon in imaging, it can be used during the feature extraction process to determine the incident light centroid for image-based sun sensors \cite{Saleem2019}. 

The BSCM process exploits blooming lines in the image space as features to improve the sun spot estimate. This methodology is not unique to the BCSM and has previously been demonstrated by Liu et al \cite{Liu2016} with a CCD detector using the Hough Transform (HT) method. Nevertheless, this algorithm was developed as an improvement to the HT method for irregular spot features due to glare. The primary strengths of the Black Sun Centroid Method are sub-pixel accuracy, adaptive spot tracking, and good performance independent of the spot location in the image space. An overview of the Black Sun Centroid Method body of literature is presented in Table \ref{tab:felit}.

The work by Saleem et al. \cite{Saleem2019} is reviewed as a case-study for the Black Sun Centroid Method. In their study, the captured image is imported and the associated pixel matrix $I(x,y)$ is generated. The image space has a Gaussian blur applied and is converted into grayscale. Next, the number of loops $f$ and threshold $T$ are set, where $\alpha$ is defined by the sensor performance.

The algorithm proceeds by iterating over the defined number of loops to find candidate points of the binary mask corners. To begin, the binary mask is generated for pixels with an intensity $I$ lower than the last iteration $I>T+f$ by decrementing the iterator $f$. The contour in the current binary mask is found and the index of the largest contour is determined. Strong corners in the current binary mask are obtained and the corner accuracy is refined with subpixels. Corner points inside the largest contour are saved that are away from the edge of the binary mask. The remaining points are marked for survival and accumulated over the defined iterations. 

All of the candidate points are accumulated by the end of the iteration count. Thereafter, the points that survived between iterations are obtained. If surviving points exist, then the point that has the largest radius and is greater than the required minimum radius is selected for feature extraction. Otherwise, if no surviving points exist, then accumulated corner points are used for feature extraction. The point with the largest radius is chosen due to the black sun being the largest segment inside the binary mask. The selected point is returned as the centroid ($C_{x},C_{y}$). The pseudocode is presented in Algorithm \ref{alg:BCSM}.

\begin{algorithm}[h]
    \caption{Black Sun Centroid Method \cite{Saleem2019}, \texttt{BSCM}}
    \label{alg:BCSM}
    \SetAlgoLined
    \SetKwInOut{Input}{input}
    \SetKwInOut{Output}{output}
    \DontPrintSemicolon
    \Input{Captured Image.}
    
    \Output{Black Sun Centroid Coordinates $C=(C_{x},C_{y})$.}
    
    // Initialization

    \quad \quad \quad Import image and generate pixel matrix, $I=I(x,y)$ 
    
    \quad \quad \quad Apply Gaussian Blur

    \quad \quad \quad Convert to grayscale
    
    \quad \quad \quad Set number of loops, $f$
    
    \quad \quad \quad Set threshold, $T=\text{maximum pixel intensity}-\alpha$
    
    \While  {$f\geq1$} {
        
        Generate binary mask for pixels with $I > T + f$
        
        Find contour in the binary mask

        Find the index of the largest contour

        Get strong corner points

        Find subpixel

        Save corner points inside the largest contour away from the edges

        Accumulate surviving points between iterations

        $f \gets f-1$

    }
    
    Accumulate corner points

    \eIf {no surviving points}{
    
    Get accumulated corner point

    Check for point with the largest radius $>$ minimum radius
    
    }{

    Get surviving points

    Check for point with the largest radius $>$ minimum radius
    
    }
    
    \KwRet($C_{x},C_{y}$)
    
\end{algorithm}

\subsubsection*{Hough Transform} \label{fe:ht}

The Basic Hough Transform (BHT) method is a feature extraction technique that detects complex patterns in binary images for centroid processing \cite{Liu2016}. The two methods discussed in this review are the Basic Hough Transform (BHT) method and the Circle Hough Transform (CHT) method. The HT was originally developed by Paul Hough in the form of a patent in 1962, however its modern form was invented by Richard Duda and Peter Hart in 1972 \cite{Duda1972}. It is a popular universal shape analysis technique for simple shapes such as straight lines, circles or ellipses. The method finds imperfect representations of features for a given shape through a voting process.

The CHT method was developed by Kimmie et al. \cite{Kimme1975} in 1975 to find the centroids of circular binary images. In the CHT, potential circle features are identified by accumulating votes in the Hough space, and the most likely circles correspond to the peaks in this accumulator. These methods are particularly useful to analyze images with noise and missing data.

The Basic Hough Transform method proposed in Liu et al. uses the CCD blooming effect, in which pixels are saturated and a blooming line is produced on the over-charged pixels. Generally the blooming effect is undesirable in imaging, however for this study the phenomenon is exploited as a feature. Since the blooming lines usually happen across the center of the sun image, they can be used to detect the centroid. In particular, the intersection of multiple blooming lines can extract the centroid with improved accuracy \cite{Liu2016}. The main advantage of the BHT method is that it is robust to noise and gaps in feature representations. An overview of the Hough Transform method body of literature is presented in Table \ref{tab:felit}.

The Circle Hough Transform method can be applied to detect and analyze a circular shape from the projected image. The method proposed in Adatrao et al. detects the centroids of circular masks on the image plane \cite{Adatrao2016}. In this case, the Circle Hough Transform searches for the radius and center coordinates of the parameterized circle. The radius of the projected image is assumed to be unknown, which makes the process a more complex 3D parameter space. The performance of the CHT method is highest when the shape of the projected image is known to be a circle in an imperfect image space. An overview of the Circle Hough Transform method body of literature is presented in Table \ref{tab:felit}.

The work by Liu et al. \cite{Liu2016} is reviewed as a case-study for the Basic Hough Transform method. In their study, the captured image is imported and the associated pixel matrix $I(x,y)$ is generated . The processed image space is then converted into grayscale. Next, the accumulator threshold parameter is set as $\tau$ to ensure only lines that get enough votes are selected as candidates. Lastly, the 2D accumulator array for the Hough space is initialized.

The main body of the algorithm starts with finding the edge features by processing the input image with any edge detector of choice. Some common methods for edge detection include canny, sobel, and adaptive thresholding. Captured images are then iterated through from each sun sensor camera. Each unique camera observation yields a new blooming line to use for later processing.

The captured image is searched through for each respective edge pixel found in the edge detection process. In the Hough transform, the shape is described in terms of its parameters. Therefore, the shape here is described by the parametric line equation $r$. Each line candidate is iterated through in the polar Hough parameter space. The accumulator cells $H[r,\theta]$ that lie along the parameterized curve are incremented.

The voting bins with with the highest values in the accumulator array represent likely line feature candidates. Hence, the line parameters ($r,\theta$) for the detected line is found where the accumulator array is maximal for all votes above the accumulator threshold. The associated line parameters are then substituted to define the detected line parameterization $r$. This step is repeated for each captured image until a respective set of detected line parameterizations is generated.

In theory, all the blooming lines should intersect at the estimated sun vector. The associated camera distortion correction model and coordinate transformations are applied to each blooming line. Each blooming line $r$ is projected into the HCS frame and the intersections for all blooming line pairs are found. The blooming line intersections are averaged to estimate the sun position using a weighted average, which is determined by the camera distortion and image disturbances. The blooming line intersection returned is the sun vector estimate ($\alpha,\beta$). The pseudocode is presented in Algorithm \ref{alg:BHT}.

\begin{algorithm}[h]
    \caption{Basic Hough Transform \cite{Liu2016}, \texttt{BHT}}
    \label{alg:BHT}
    \SetAlgoLined
    \SetKwInOut{Input}{input}
    \SetKwInOut{Output}{output}
    \SetKwProg{GetCentroid}{GetCentroid}{}
    \DontPrintSemicolon

    \Input{Captured Images.}
    
    \Output{Sun Vector ($\alpha,\beta$).}
    
    // Initialization

    \quad \quad \quad Import images and generate pixel matrices, $I=I(x,y)$ 

    \quad \quad \quad Convert to grayscale
    
    \quad \quad \quad Set threshold value, $\tau$

    \quad \quad \quad Initialize accumulator array indicating Hough space $H[r,\theta]=0$

    Find the image edge using any edge detector (canny, sobel, adaptive thresholding)

    \ForEach{image $\textbf{I}(x,y)$} {
    
    \ForEach{edge pixel ($x,y$)} {

        \For{$\theta=0$ \KwTo $\pi$} {

            $r\gets x\cos(\theta)+y\sin(\theta)$

            $H[r,\theta]\gets H[r,\theta]+1$

        }
        
    }

    Find $(r,\theta)$ where $H[r,\theta]$ is maximal as all votes above threshold $\tau$

    The detected line is $r=x\cos(\theta)+y\sin(\theta)$

    }

    Project each blooming line $r$ into HCS

    Find the intersections for all pairs of blooming lines

    Apply the associated camera distortion model and coordinate transformations

    Calculate the sun vector estimate ($\alpha,\beta$) as the weighted average of all pair intersections
    
    \KwRet{($\alpha,\beta$)}
    
\end{algorithm}

The work by Adatrao et al. \cite{Adatrao2016} is reviewed as a case-study for the Circle Hough Transform method. In the following study, the captured image is imported and the associated pixel matrix is generated $I(x,y)$. The processed image space is then converted into grayscale. Next, the accumulator threshold parameter is set as $\tau$ to ensure only lines that get enough votes are selected as candidates. Lastly, the 3D accumulator array for the Hough space is initialized.

The algorithm begins with finding the edge features by processing the input image with any edge detector of choice. As with the HT, common methods for edge detection include canny, sobel, and adaptive thresholding. For this algorithm the radius is assumed to be unknown. The range of potential feature radii are iterated through to a maximum of the diagonal image length. The captured image is searched through for each respective edge pixel found in the edge detection process.

In the Hough transform, the shape is described in terms of its parameters. Therefore, the shape here is described by the parametric circle equations $a$ and $b$. Each circle candidate is iterated through in the polar Hough parameter space. The accumulator cells $H[a,b,r]$ that lie along the parameterized curve are incremented.

The voting bins with with the highest values in the accumulator array represent likely circle feature candidates. Hence, the circle parameters ($a,b,r$) for the detected circle is found where the accumulator array is maximal for all votes above the accumulator threshold. The associated circle parameters are then substituted to define the detected circle parameterization $a$ and $b$. The circle center ($a,b$) is returned as the centroid feature estimate. The pseudocode is presented in Algorithm \ref{alg:CHT}.

\begin{algorithm}[h]
    \caption{Circle Hough Transform \cite{Adatrao2016}, \texttt{CHT}}
    \label{alg:CHT}
    \SetAlgoLined
    \SetKwInOut{Input}{input}
    \SetKwInOut{Output}{output}
    \SetKwProg{GetCentroid}{GetCentroid}{}
    \DontPrintSemicolon

    \Input{Captured Image.}
    
    \Output{Centroid Coordinates $C=(a,b)$.}
    
    // Initialization

    \quad \quad \quad Import image and generate pixel matrix, $I=I(x,y)$ 

    \quad \quad \quad Convert to grayscale
    
    \quad \quad \quad Set threshold value, $\tau$

    \quad \quad \quad Initialize accumulator array indicating Hough space $H[a,b,r]=0$

    Find the image edge using any edge detector (canny, sobel, adaptive thresholding)

    \For{$r=0$ \KwTo diagonal image length}{
    
    \ForEach{edge pixel ($x,y$)} {

        \For{$\theta=0$ \KwTo $2\pi$} {

            $a \gets x - r\cos(\theta)$
    
            $b \gets y - r\sin(\theta)$

            $H[a,b,r]\gets H[a,b,r]+1$

        }

    }

    }

    Find $(a,b,r)$ where $H[a,b,r]$ is maximal as all votes above threshold $\tau$

    // The detected circle
    
    $a = x - r\cos(\theta)$
    
    $b = y - r\sin(\theta)$
    
    \KwRet{($a,b$)}
    
\end{algorithm}

\subsubsection*{Fast Multi-Point MEANSHIFT (FMMS)} \label{fe:fmms}

The Fast Multi-Point MEANSHIFT Method was developed to achieve high-accuracy feature detection, while also enabling adaptive feature tracking and robustness against missing features. The algorithm is implemented in two parts. The first part of the algorithm establishes the feature description and feature similarity function. The second part of the algorithm determines the feature pointing and tracking. The FMMS method is developed as a performance and tracking capability improvement to the FEIC method \cite{You2011}. 

In the first step, the feature of the spot is defined by an isotropic, convex profile with a monotonically decreasing kernel function. The kernel function chosen in the selected study is the Epanechnikove kernel. The description of the feature goal and the candidate is established. Thereafter, the Bhattacharyya coefficient describes the degree of similarity between the feature goal and the candidate. 

In the second step, the feature pointing and tracking is obtained by searching for a new feature location in the current frame that minimizes the distance between the feature goal and candidate. The search process begins in the previous location of a spot and proceeds around the search neighborhood. The new feature location is then found as the kernel shift from the current location to the new predicted location. The FMMS method is developed with a multi-aperture architecture to ensure robustness to missing features by lowering the uncertain feature weights. 

The primary strengths of the FMMS method are sub-pixel accuracy, adaptive spot tracking, and robustness to missing features. An overview of the FMMS method body of literature is presented in Table \ref{tab:felit}.

The work by You et al. \cite{You2011} is reviewed as a case-study for the FMMS method. In their study, the captured image is imported and the associated pixel matrix $I(x,y)$ is generated. The processed image space is then converted into grayscale. Next, the number of sun spots  $N_{spot}$ is set and the convergence tolerance $\varepsilon$ is defined.

The algorithm begins by establishing the feature description and feature similarity function. The pixels values are computed to describe the quantitative feature space $b(x)$. The kernel function weighted for all quantified pixels is then selected to describe the feature of the spot. For this case-study, the Epanechnikove kernel $K_{E}(x)$ is implemented. The feature vector of the goal spot is then calculated for each pixel $x_{i}$ as $\hat{q}_{u}$, where $C$ is a normalization function. Lastly, the feature vector of the candidate spot is calculated for each pixel $x_{i}$ as $\hat{p}_{u}(y)$, where $C_{h}$ is a normalization factor. The pseudocode for the feature description and feature similarity function is presented in Algorithm \ref{alg:FMMS1}.

\begin{algorithm}[h]
    \caption{Fast Multi-Point MEANSHIFT (Spot Features Description and Similarity Function) \cite{You2011}, \texttt{FMMS}}
    \label{alg:FMMS1}
    \SetAlgoLined
    \SetKwInOut{Input}{input}
    \SetKwInOut{Output}{output}
    \SetKw{Break}{break}
    \SetKwProg{GetCentroid}{GetCentroid}{}
    \DontPrintSemicolon

    \Input{Captured Image.}
    
    \Output{Centroid Coordinates $\hat{y}$.}
    
    // Initialization

    \quad \quad \quad Import image and generate pixel matrix, $I=I(x,y)$ 

    \quad \quad \quad Convert to grayscale

    \quad \quad \quad Number of sun spots, $N_{spot}$
    
    \quad \quad \quad Set convergence tolerance, $\varepsilon$

    Compute pixels value in the quantitative feature space\linebreak
    $\displaystyle
    b(x)=\begin{cases}
    1 &x<t_{1}\\
    2 &t_{1}\leq x<t_{2}\\
    \vdots &\vdots\\
    m &t_{n}\leq x
    \end{cases}
    $\\

    Define the Epanechnikove kernel
    $\displaystyle
    K_{E}(x)=\begin{cases}
    C_{k}(1-\left \| x \right \|^{2}): &\left \| x \right \|<1\\
    0: &\left \| x \right \|\geq 1
    \end{cases}
    $\\

    \ForEach{pixel $x_{i}$}{
    Feature vector of the goal spot
    $\displaystyle
    \hat{q}_{u}=C\sum_{i=1}^{n}K_{E}(\left \| x_{i} \right \|^{2})\delta[b(x_{i})-u],\linebreak u=1,\cdots,m\quad
    $
    where $C$ makes
    $\displaystyle
    \sum_{u=1}^{m}\hat{q}_{u}=1 \to C=\frac{1}{\sum_{i=1}^{n}K_{E}(\left \| x_{i}^{*} \right \|^{2})}
    $\\

    Feature vector of candidate spot\quad
    $\displaystyle
    \hat{p}_{u}(y)=C_{h}\sum_{i=1}^{n_{h}}K_{E}\left ( \left \| \frac{y-x_{i}}{h} \right \|^{2} \right )\delta[b(x_{i})-u],\linebreak u=1,\cdots,m\quad
    $
    where $C_{h}$ makes
    $\displaystyle
    \sum_{u=1}^{m}\hat{p}_{u}=1 \to C_{h}=\frac{1}{\sum_{i=1}^{n}K_{E}(\left \| \frac{y-x_{i}}{h} \right \|^{2})}
    $\\
    }

\end{algorithm}

The FMMS method proceeds with the second step of calculating feature pointing and tracking for each spot feature in the image space. First, the feature vector of the candidate spot in the previous frame is computed as $\hat{p}_{u}(\hat{y}_{0})$. Next, the Bhattacharyya coefficient at the previous frame $\hat{y}_{0}$ is calculated as $\rho[\hat{p}(\hat{y}_{0}),\hat{q}]$ as a metric for the degree of similarity between the feature goal and the candidate. The algorithm iterates while the difference between the spot of the previous frame $\hat{y}_{0}$ and the current frame $\hat{y}_{1}$ is greater than the convergence tolerance.

The kernel density estimation is calculated as $w_{i}$, in which the maximum density estimation value is found in the neighborhood. Following this step, the updated spot location is calculated as the kernel shift from the current location $\hat{y}_{0}$ to the new location $\hat{y}_{1}$. Next, the Bhattacharyya coefficient at the current frame $\hat{y}_{1}$ is calculated as $\rho[\hat{p}(\hat{y}_{1}),\hat{q}]$. The current spot location is iteratively updated by the mean of the current and previous frame values until the Bhattacharyya coefficient of the current frame is greater than that of the previous frame.

If the convergence tolerance exit condition is not met, then the previous frame value for the start of the next loop iteration is updated as the current frame value. Otherwise, if the exit condition is met then the new spot location $\hat{y}_{1}$ is obtained for the n-th spot. This process is repeated until all $N_{spot}$ spots have been computed. Thereafter, the algorithm completes and the centroid coordinates are returned as $\hat{y}$. The pseudocode for the feature pointing and tracking is presented in Algorithm \ref{alg:FMMS2}.

\begin{algorithm}[h]
    \caption{Fast Multi-Point MEANSHIFT (Sun Spot Pointing and Tracking) \cite{You2011}, \texttt{FMMS}}
    \label{alg:FMMS2}
    \SetAlgoLined
    \SetKw{Break}{break}
    \SetKwProg{GetCentroid}{GetCentroid}{}
    \DontPrintSemicolon
    \setcounter{AlgoLine}{11}

    \ForEach{spot $N_{spot}$} {

    Calculate the feature vector of candidate spot $\{\hat{p}_{u}(\hat{y}_{0})\}_{u=1,\cdots,m}$

    Bhattacharyya coefficient at $\hat{y}_{0}$
    $\displaystyle
    \rho[\hat{p}(\hat{y}_{0}),\hat{q}]=\sum_{u=1}^{m}\sqrt{\hat{p}_{u}(\hat{y}_{0})\hat{q}_{u}}
    $\\

    \While  {$\left \| \hat{y}_{1}-\hat{y}_{0} \right \|>\varepsilon$} {

    \ForEach{pixel $x_{i}$}{

    $\displaystyle
    w_{i}=\sum_{u=1}^{m}\delta[b(x_{i})-u]\sqrt{\frac{\hat{q}_{u}}{\hat{p_{u}}(\hat{y_{1}})}}
    $\\

    Calculate updated spot location
    $\displaystyle
    \hat{y}_{1}=\frac{\sum_{i=1}^{n_{h}}x_{i}w_{i}g\left ( \left \| \frac{\hat{y}-x_{i}}{h} \right \|^{2} \right )}{\sum_{i=1}^{n_{h}}w_{i}g\left ( \left \| \frac{\hat{y}-x_{i}}{h} \right \|^{2} \right )},\linebreak
    \text{where } g(x)=-K_{E}^{'}(x)
    $\\

    }

    Bhattacharyya coefficient at $\hat{y}_{1}$
    $\displaystyle
    \rho[\hat{p}(\hat{y}_{1}),\hat{q}]=\sum_{u=1}^{m}\sqrt{\hat{p}_{u}(\hat{y}_{1})\hat{q}_{u}}
    $\\

    \While  {$\rho[\hat{p}(\hat{y}_{1}),\hat{q}]<\rho[\hat{p}(\hat{y}_{0}),\hat{q}]$} {

        $\displaystyle
        \hat{y}_{1}\gets\frac{1}{2}(\hat{y}_{0}+\hat{y}_{1})
        $\\

    }

    \If {$\left \| \hat{y}_{1}-\hat{y}_{0} \right \|>\varepsilon$}{
    $\hat{y}_{0}\gets\hat{y}_{1}$
    }

    }

    Get the $n$-th spot new location $\hat{y}_{1}$

    }

    \KwRet{$\hat{y}$}

\end{algorithm}

\subsubsection*{Image Filtering Method (IFM)} \label{fe:ifm}

The Image Filtering Method is a two-step feature extraction algorithm that reduces noise in the pre-processing phase through the use of smoothing filters. The second step of the algorithm proceeds with the Basic Centroiding Method. The technique was developed as an improvement to the BCM for low SNR images \cite{Chang2008}.

The IFM smooths the noise interference from the input image through neighbor intensity averaging. Specifically, the original intensity data of each subject pixel is updated with the average value of the neighboring pixels. This filtering approach can reduce the noise of the original input image. 

However, the feature extraction error can increase if the method is used for images with very high noise levels due to averaging. An overview of the IFM body of literature is presented in Table \ref{tab:felit}.

The work by Chang et al. \cite{Chang2008} is reviewed as a case-study for the IFM method. In their study, the captured image is imported and the associated pixel matrix $I(x,y)$ is generated. The processed image space is then converted into grayscale. The number of pixel rows and columns are then established.

The image space is first pre-processed by applying the intensity noise filter as neighbor intensity averaging over each pixel. The image space is first pre-processed by applying the intensity noise filter as neighbor intensity averaging over each pixel. Therefore, the original intensity data of each subject pixel is updated with the average value of the neighboring pixels. The filtering is achieved with a $3 \times 3$ mean filter kernel.

The centroid algorithm then proceeds identically to the Basic Centroid Method. Therefore, the total intensity $I_{Tot}$ is first calculated over the image space as the sum over each pixel. The centroid coordinates ($x_{c},y_{c}$) for each axis are then calculated as the weighted average over the image space. The pseudocode for the technique is presented in Algorithm \ref{alg:IFM} \cite{Chang2008}.

\begin{algorithm}[htbp]
    \caption{Image Filtering Method \cite{Chang2008}, \texttt{IFM}}
    \label{alg:IFM}
    \SetAlgoLined
    \SetKwInOut{Input}{input}
    \SetKwInOut{Output}{output}
    \SetKwProg{GetCentroid}{GetCentroid}{}
    \DontPrintSemicolon

    \Input{Captured Image.}
    
    \Output{Centroid Coordinates $C=(x_{c},y_{c})$.}
    
    // Initialization

    \quad \quad \quad Import image and generate pixel matrix, $I=I(x,y)$ 

    \quad \quad \quad Convert to grayscale
    
    \quad \quad \quad Set number of pixel matrix rows, $N_{r}$

    \quad \quad \quad Set number of pixel matrix columns, $N_{c}$

    //  Pre-processing phase
    
    \ForEach{pixel \textbf{I}$(x,y)$} {

    $\displaystyle
    I_{m,n}=\frac{1}{9}\sum_{i=m-1}^{m+1}\sum_{j=n-1}^{n+1}I_{i,j}
    $, where:
    \[I_{i,j} =
    \begin{cases}
    0 & \text{if } i,j < 1
    \end{cases}\]
    $\left( 1 \leq m \leq 256,\ 1 \leq n \leq 256 \right)$
    }

    // Centroiding algorithm
    
    \ForEach{pixel \textbf{I}$(m,n)$} {
    
        $\displaystyle
        I_{Tot}=\sum_{m=1}^{N_{r}}\sum_{n=1}^{N_{c}}I\left ( m,n \right )
        $\\
        
        $\displaystyle
        x_{c}=\frac{1}{I_{Tot}}\sum_{m=1}^{N_{r}}\sum_{n=1}^{N_{c}}x_{m}I\left ( m,n \right )
        $\\

        $\displaystyle
        y_{c}=\frac{1}{I_{Tot}}\sum_{m=1}^{N_{r}}\sum_{n=1}^{N_{c}}y_{n}I\left ( m,n \right )
        $\\

    }
    \KwRet{($x_{c},y_{c}$)}
    
\end{algorithm}

\subsubsection*{Template Method (TM)} \label{fe:tm}

The Template Method uses a theoretical spot image model to determine the centroid of the actual spot image. It is proposed that the best way of estimating the feature space of a noisy image is to compare the theoretical and actual image. Hence, this method assumes that the theoretical image shape is known a priori. The technique was developed to improve the accuracy of centroid estimation when the image data is affected by noise levels. In particular, it is an improvement over the BCTM and IFM for noisy image feature extraction \cite{Chang2008}.

The theoretical spot image is formulated based on the incident light angle. The image spot centroid is then found using the template method, in which a matching process between the actual and theoretical spot image is calculated. The difference between the actual and theoretical spot images is minimized to find the best template fit.

The Template Method yields a higher accuracy than BCTM and IFM. Furthermore, the method demonstrates a consistent level of accuracy across noise levels unlike BCTM and IFM, in which the centroid error increases with the noise level. Therefore, the Template Method is particularly effective in reliability critical applications with variable noise environments. An overview of the Template Method body of literature is presented in Table \ref{tab:felit}.

The work by Chang et al. \cite{Chang2008} is reviewed as a case-study for the Template Method method. In their study, the captured image is imported and the associated pixel matrix $I(x,y)$ is generated. The processed image space is then converted into grayscale. The number of pixel rows and columns are then established. Finally, the threshold value $\mu$ and FOV $\phi$ is set.

The main body of the algorithm begins by assigning the brightest points of the image to be the vector of candidate image centroids $\mathbf{P}_{c}$. The bi-level measured image $I^{\prime}(x,y)$ is recovered by applying a threshold to the original image, where $I_{max}$ is the maximum image intensity. The bi-level measured image is read as one if greater than the threshold, while all data lower than the threshold is processed as zero. The algorithm proceeds by iterating through each candidate image centroid to find the error that minimizes the theoretical and actual image error. The value of the current candidate centroid is defined as $p_{cn}$, which is in the set of candidate image centroids vector.

The sunlight vector for the current candidate centroid is calculated as $\overline{S}$. Thereafter, the points of the light source $p_{0}$ can be determined if the sunlight vector $\overline{S}$ and the FOV $\phi$ are known. Furthermore, all points within the mask hole $p_{1}$ can be determined from the  mask hole dimensions. Hence, all points of the light spot on the detector plane can be described by the parameterization $p_{t}$, where $F$ is the focal length. The theoretical image template for a given candidate centroid is fully described by the parameterization $p_{t}$.

The theoretical bi-level image $J(x,y)$ is generated from the previously defined parameterization, in which the value is one inside the projected light spot and zero otherwise. Next, the error for the current candidate centroid $E_{n}$ is calculated, which is defined as the difference between the measured and theoretical bi-level images. Therefore, the estimated image centroid $p^{*}_{c}$ can be defined as the candidate centroid $\mathbf{P}_{c}$ that minimizes the error value $E_{n}$. The centroid coordinates are returned as the pair ($x_{c},y_{c}$). The pseudocode for the technique is presented in Algorithm \ref{alg:TM} \cite{Chang2008}.

\begin{algorithm}[h]
    \caption{Template Method \cite{Chang2008}, \texttt{TM}}
    \label{alg:TM}
    \SetAlgoLined
    \SetKwInOut{Input}{input}
    \SetKwInOut{Output}{output}
    \SetKwProg{GetCentroid}{GetCentroid}{}
    \DontPrintSemicolon

    \Input{Captured Image.}
    
    \Output{Centroid Coordinates $C=(x_{c},y_{c})$.}
    
    // Initialization

    \quad \quad \quad Import image and generate pixel matrix, $I=I(x,y)$ 

    \quad \quad \quad Convert to grayscale
    
    \quad \quad \quad Set threshold value, $\mu$

    \quad \quad \quad Define FOV, $\phi$

    Assign brightest points of the image to be the vector of candidate image centroids\linebreak
    $\displaystyle
    \mathbf{P}_{c} =\argmaxA_{x,y} I(x,y),\quad \mathbf{P}_{c} = p_{c1},\cdots ,p_{cN}
    $\\

    Apply threshold to recover bi-level measured image as
    $\displaystyle
    I^{\prime}(x,y)= \begin{cases}
    0: &I<\mu I_{max}\\
    1: &I>\mu I_{max}
    \end{cases}
    $\\

    All data lower than the threshold is processed as 0
    
    \ForEach{centroid $\mathbf{P}_{c}$} {

        $p_{cn}=(x_{c},y_{c},z_{c})\in \mathbf{P}_{c}$

        Calculate the sunlight vector from ${p}_{cn}$
        $\overline{S}=-\vect{x_{c},y_{c},z_{c}}$

        Calculate $p_{0}$ and $p_{1}$ from $\overline{S}$ and $\phi$

        $\displaystyle
        x_{t}=\left ( \frac{z_{t}-z_{0}}{z_{1}-z_{0}} \right )\times \left ( x_{1}-x_{0} \right )+x_{0}
        $\\
        
        $\displaystyle
        y_{t}=\left ( \frac{z_{t}-z_{0}}{z_{1}-z_{0}} \right )\times \left ( y_{1}-y_{0} \right )+y_{0}
        $\\

        $\displaystyle
        z_{t}=F
        $\\

        $\displaystyle
        p_{t}=(x_{t},y_{t},F)
        $\\

        Compute theoretical bi-level image as
        $\displaystyle
        J(x,y)= \begin{cases}
        0: &x,y\neq x_{t},y_{t}\\
        1: &x,y = x_{t},y_{t}
        \end{cases}
        $\\

        \ForEach{pixel $I^{\prime}_{i,j}$} {

        $\displaystyle
        E_{n} = \sum_{i,j} \left [ I^{\prime}_{i,j}-J\left ( x_{i},y_{i} \right ) \right ]^2
        $\\
        
        }

    }

    Estimate image centroid $p^{*}_{c}$ as $\mathbf{P}_{c}$ value that minimizes error $E_{n}$ as
    $\displaystyle
    p^{*}_{c} =\argminA_{\mathbf{P}_{c}} E_{n},\quad \mathbf{P}_{c} = p_{c1},\cdots ,p_{cN}
    $\\
    
    \KwRet{($x_{c},y_{c}$)}
    
\end{algorithm}

\subsubsection*{Feature Extraction Image Correlation (FEIC)} \label{fe:feic}

The Feature Extraction Image Correlation method is an approach of implementing subsampling, multi-windowing and a prediction algorithm together. The sun spot image correlation (IC) and the centroiding algorithm operate together in what is known as the image correlation algorithm (ICA). The FEIC method is developed to improve upon traditional centroiding accuracy, while achieving reliability with missing or degraded sun spot features \cite{Xing2008}.

The FEIC technique operates through a process of iterative sun spot windowing and feature correlation matrix centroiding. The FE method enables the extraction of the search windows surrounding the sun spots and their displacements. The sun spot centroid displacements to their template centroid locations are assumed to be equal for efficiency. Furthermore, the ICA method works as the weight average of all the correlation matrix centroids. Therefore, the centroid of the sum of all correlation matrices is the sun spot centroid estimate. Sun spot tracking is not necessary since the spot location in the previous frame is used on the current frame as a predictive search window.

The primary strength of the FEIC method is the robustness enabled through multi-spot weighted averaging. This feature grants some immunity to missing sun spots and feature deterioration. However, the centroid accuracy is affected by the window size and center position. An overview of the FEIC method body of literature is presented in Table \ref{tab:felit}.

The work by Xing et al. \cite{Xing2008} is reviewed as a case-study for the FEIC method. In their study, the captured image is imported and the associated pixel matrix $I(x,y)$ is generated. The processed image space is then converted into grayscale. The number of sun spots projected $N_{s}$ is established. Finally, the correlation elements - 1 $N_{c}\times N_{c}$ and template pixels $N_{tpix}\times N_{tpix}$ are defined respectively.

The algorithm begins by iterating through each sun spot $N_{s}$ and applying the FEIC algorithm process. The template of the current sun spot is obtained as $T_{s}$, where $T_{s}(i,j)$ denotes the gray value of the $i$-th row and $j$-th column in template $T$. Next, the current associated template centroid location is obtained as $(\underline{x}_{s},\underline{y}_{s})$, where $s=1,\dotsc,N_{s}$. The initial value of the displacements of the sun spots in the FE image to the original template image are obtained as ($\Delta x, \Delta y$).

Thereafter, the sun spot window center pixel location is computed for the current spot as the coordinate pair $x_{s}$ and $y_{s}$, where function int$[x]$ denotes the maximal integer less than or equal to x. The sun spot window region $P_{s}$ is then extracted from the previously computed sun spot window center pixel locations for the current sun spot. With this information, the sun spot correlation matrix $C_{s}$ can be computed for the current sun spot using the sun spot window region $P_{s}$ and template $T_{s}$ respectively. The current sun spot correlation result matrix $C_{s}\left ( m,n \right )$ is calculated, which is composed of $m\times n$ elements. The calculation consists of a double summation over the template image space, where the template of the current sun spot is $T_{s}$, the feature extracted image window is $P_{s}$, and $T_{s}(i,j)$ and $P_{s}(i,j)$ are the gray values of the respective image window regions.

Moreover, a total of $N_{s}$ matrices $C_{s}$ are obtained as $C_{1},\dotsc,C_{N_{s}}$ through the algorithm process, which correspond to templates $T_{1},\dotsc,T_{N_{s}}$ and extracted windows $P_{1},\dotsc,P_{N_{s}}$. In the final step, the total centroid estimate ($x_{c},y_{c}$) is computed as the centroid of the sum of all correlation matrices. This step is iterated over each sun spot $s$ and correlation element $(m,n)$. Thereafter, the two-axis sun angle ($\alpha, \beta$) can be computed. The current frame centroid estimate ($x_{c}, y_{c}$) is then transferred to the new sun spot ($\Delta x, \Delta y$) for the next feature extraction. The centroid estimate ($x_{c},y_{c}$) is returned as the output. The pseudocode for the technique is presented in Algorithm \ref{alg:FEIC} \cite{Xing2008}.

\begin{algorithm}[h]
    \caption{Feature Extraction Image Correlation \cite{Xing2008}, \texttt{FEIC}}
    \label{alg:FEIC}
    \SetAlgoLined
    \SetKwInOut{Input}{input}
    \SetKwInOut{Output}{output}
    \SetKwFunction{Fint}{int}
    \DontPrintSemicolon

    \Input{Captured Image.}
    
    \Output{Centroid Coordinates $C=(x_{c},y_{c})$.}
    
    // Initialization

    \quad \quad \quad Import image and generate pixel matrix, $I=I(x,y)$ 

    \quad \quad \quad Convert to grayscale
    
    \quad \quad \quad Number of sun spots, $N_{s}$

    \quad \quad \quad Correlation elements - 1, $N_{c}\times N_{c}$

    \quad \quad \quad Template pixels, $N_{tpix}\times N_{tpix}$

    \For{$s=1$ \KwTo $N_{s}$} {

        Obtain each template as $T_{s}$, where $T_{s}(i,j)$ denotes the gray value of the $i$-th row and $j$-th column in template $T$

        Obtain each associated template centroid location as $(\underline{x}_{s},\underline{y}_{s})$, where $s=1,\cdots,N_{s}$

        Obtain initial value of ($\Delta x, \Delta y$)

        // Compute every sun spot window center
    
        $x_{s}= \Fint[\Delta x+\underline{x}_{s}+0.5]$

        $y_{s}= \Fint[\Delta y+\underline{y}_{s}+0.5]$, where function $\Fint[x]$ denotes the maximal integer less than or equal to x

        Extract every sun spot window region $P_{s}$ according to ($x_{s}, y_{s}$)

        // Compute the sun spot correlation $C_{s}$

        \ForEach{pixel i,j ($N_{tpix}\times N_{tpix}$)} {
    
            $\displaystyle
            C_{s}\left ( m,n \right )=\sum_{i=1}^{N_{tpix}}\sum_{j=1}^{N_{tpix}}P_{s}\left ( m+i,n+j \right )T_{s}\left ( i,j \right )\linebreak
            m,n=0,\cdots,N_{c} 
            $\\

        }

        \ForEach{element m,n ($N_{c}\times N_{c}$)} {

            $\displaystyle
            x_{c}=\Delta x+\frac{\sum_{s=1}^{N_{s}}\sum_{m=0}^{N_{c}}\sum_{n=0}^{N_{c}}C_{s}\left ( m,n \right ) \left ( m-4 \right )}{\sum_{s=1}^{N_{spot}}\sum_{m=0}^{N_{c}}\sum_{n=0}^{N_{c}}C_{s}\left ( m,n \right )}
            $\\
        
            $\displaystyle
            y_{c}=\Delta y+\frac{\sum_{s=1}^{N_{s}}\sum_{m=0}^{N_{c}}\sum_{n=0}^{N_{c}}C_{s}\left ( m,n \right )       \left ( n-4 \right )}{\sum_{s=1}^{N_{spot}}\sum_{m=0}^{N_{c}}\sum_{n=0}^{N_{c}}C_{s}\left ( m,n \right )}
            $\\
                
        }

    }

    Compute two-axis sun angle ($\alpha, \beta$)

    Transfer ($x_{c}, y_{c}$) to the new ($\Delta x, \Delta y$) for the next feature extraction

    \KwRet{($x_{c},y_{c}$)}
    
\end{algorithm}

\subsection{Parametric}

Feature extraction techniques can generally be categorized into two classes: conventional and parametric methods. Conventional methods extract features directly from the image space, relying on minimal prior knowledge of the expected sunspot pattern on the detector. In contrast, parametric methods enhance the feature extraction process by leveraging prior information about the anticipated shape of the sunspot. The following sections present two such parametric techniques: the linear-phase method and the eigen-analysis method.


\subsubsection*{Linear-phase} \label{fe:linearphase}

The linear-phase method is a parametric feature extraction technique that is based on the Discrete Fourier Transform (DFT). The centroid detection process is computed as a frequency domain estimation algorithm based on the classic Phase-Correlation Method. The parametric processing approach offers improved accuracy over conventional processing methods. Specifically, the linear-phase estimation achieves improved performance compared to the BCTM and PD techniques. The method offers a good balance of computational cost and performance gain \cite{Enright2007b}.  

The linear-phase feature extraction technique requires the DFT and LSQ mathematical operations to estimate the sun spot centroids. Some advantages and disadvantages of the feature extraction implementation include improved accuracy and overlapping feature errors, respectively. The linear-phase estimation technique is particularly effective at low noise levels. However, noise causes the linear-phase performance gain to degrade faster than the BCTM and eigen-analysis methods. 

For the parametric methods, increasing the pixel size reduces the effective resolution, while decreasing the pixel size has less effect on the effective resolution \cite{Enright2008b}. The technique can suffer from performance loss due to partially overlapping features leading to feature space ambiguity. The computational load is higher than conventional methods, but lower than eigen-analysis. An overview of the linear-phase method body of literature is presented in Table \ref{tab:felit}.

The work by Enright et al. \cite{Enright2007b} is reviewed as a case-study for the linear-phase method. In the study, the captured image is imported and the associated pixel matrix $I(x,y)$ is generated. The processed image space is then converted into grayscale. Furthermore, the sampling spacing $\Delta X$ is set. The algorithm begins with the definitions of the signal DFTs. The DFT of the zero shift signal is defined as $S_{0}[n]$ and the DFT of the shifted signal is defined as $S_{\alpha}[n]$. The discrete signal $I[n]$ and its DFT $S[k]$ is then computed.

The frequency domain effect of a space-domain shift is computed, where the illumination has been shifted by $\alpha$ samples. Here, a shift in the space domain is equal to multiplication by a linear-phase term in the frequency domain. The exponential term $Y[k]$ is isolated to find the phase angle of $Y[k]$ as $\Psi[k]$. The LSQ fit process is performed on the phase angle $\Psi[k]$ to get an estimate of the slope $\alpha$. With this information, the delay estimate $\tau$ can be found from the slope and sampling spacing. The centroid estimate $\tau$ is returned as the output. The pseudocode for the technique is presented in Algorithm \ref{alg:linearphase} \cite{Enright2007b}.

\begin{algorithm}[h]
    \caption{Linear-phase \cite{Enright2007b}}
    \label{alg:linearphase}
    \SetAlgoLined
    \SetKwInOut{Input}{input}
    \SetKwInOut{Output}{output}
    \DontPrintSemicolon

    \Input{Captured Image.}
    
    \Output{Peak Coordinate $\tau$.}
    
    // Initialization

    \quad \quad \quad Import image and generate pixel matrix, $I=I[n]$ 

    \quad \quad \quad Convert to grayscale

    \quad \quad \quad Sampling spacing $\Delta X$

    $S_{0}[n] \gets$ DFT of the zero shift signal

    $S_{\alpha}[n] \gets$ DFT of the shifted signal

    $\displaystyle
    I[n]\overset{DFT}\leftrightarrow S[k]
    $\\

    $\displaystyle
    I[n-\alpha]\overset{DFT}\leftrightarrow S[k]e^{-j\frac{2\pi\alpha k}{N}}
    $\\

    $\displaystyle
    Y[k]=\frac{S_{\alpha}[n]}{S_{0}[n]}=e^{-j\frac{2\pi\alpha k}{N}}
    $\\

    Calculating the phase angle of $Y[k]$ as
    $\displaystyle
    \Psi[k]=-\frac{2\pi\alpha k}{N}
    $\\

    Perform least squares fit on $\Psi[k]$ to get $\alpha$

    The delay estimate is
    $\displaystyle
    \tau=\alpha\Delta X
    $\\

    \KwRet{$\tau$}
    
\end{algorithm}

\subsubsection*{Eigen-analysis} \label{fe:eigena}

The eigen-analysis method is a parametric feature extraction technique that achieves sub-pixel resolution by exploiting the eigen-structure of the system covariance matrix. The approach is popularized by parametric time delay estimation (TDE) techniques. Therefore, the sun spot displacement problem is analogous to time-of-arrival (TOA) problems. The techniques separate the cross-correlation matrix into noise and signal subspaces by applying subspace decomposition via eigen-value decomposition (EVD). This approach was developed to improve the robustness of the feature extraction process to random noise and signal interference \cite{Enright2006b}.

The eigen-analysis feature extraction technique requires the DFT, cross-correlation, iterative minimization, and EVD mathematical operations to estimate the sun spot centroids. Advantages of the technique include robustness to system noise and the ability to estimate the linear shift in the signal with a limited number of observations. The eigen-analysis estimation technique is effective at the full range of noise levels. Furthermore, noise causes the performance gain to degrade much slower than that of the BCTM and linear-phase methods.

Similar to the linear-phase method, increasing the pixel size reduces the effective resolution, while decreasing the pixel size has little effect on the effective resolution \cite{Enright2008b}. The computational load for this algorithm is high, especially compared to conventional methods. An overview of the eigen-analysis method body of literature is presented in Table \ref{tab:felit}.

\begin{algorithm}[h]
    \caption{Eigen-analysis \cite{Enright2006b}}
    \label{alg:eigen}
    \SetAlgoLined
    \SetKwInOut{Input}{input}
    \SetKwInOut{Output}{output}
    \DontPrintSemicolon

    \Input{Captured Image.}
    
    \Output{Peak Coordinate $\tilde{\tau}$.}
    
    // Initialization

    \quad \quad \quad Import image and generate pixel matrix, $I=I[n]$ 

    \quad \quad \quad Convert to grayscale

    \quad \quad \quad Sampling spacing $\Delta X$

    $r_{1}(X)=a_{1}\cdot s(X)$ //  Reference signal

    $r_{2}(X)=a_{2}\cdot s(X-\tau)+w(X)$ // Delayed signal
    
    $r_{1}[n]=a_{1} \cdot s(n\Delta X)$

    $r_{2}[n]=a_{2} \cdot s(n\Delta X-\tau)+w(n\Delta X)\linebreak n=0,1,\cdots ,N-1$

    Zero-pad $r_{1}[n]$ and $r_{2}[n]$ to $K_{x} = 2N-1$

    $\displaystyle
    s[n]\overset{DFT}\leftrightarrow S[k]
    $\\

    $\displaystyle
    w[n]\overset{DFT}\leftrightarrow W[k]
    $\\

    // Cross-correlation between $r_{1}[n]$ and $r_{2}[n]$
    
    \For{$k=0$ \KwTo $K_{x}-1$} {
    $\displaystyle
    R_{r12}(\tau)=\sum_{k=0}^{K_{x}-1}r_{1}(n-\tau)\cdot r-{2}^{*}(n) =\sum_{k=0}^{K_{x}-1}(a_{1}a_{2}^{*}\left | S[k] \right |^{2}e^{j\left (\frac{2\pi}{K_{x}}  \right )k\tau}+a_{1}S[k]W[k])e^{-\left (\frac{2\pi}{K_{x}}  \right )k\tau}
    $\\
    }

    Zero-pad $R_{r12}[k]$ to length $2K_{x}-2$

    Correlation function
    $R_{x}[l]=DFT^{-1}[\left | R_{r12}[k] \right |^{2}]\linebreak k,l=0,1,\cdots ,2K_{x}-2$

    Truncated covariance matrix
    $\displaystyle
    \hat{R}_{\text{sub\_x}}=\begin{bmatrix}
    R_{x}[0] & R_{x}^{*}[1] & \cdots  & R_{x}^{*}[N-1]\\ 
    R_{x}[1] & R_{x}[0] & \cdots & R_{x}^{*}[N-2]\\ 
    \vdots  & \vdots & \ddots  & \vdots \\ 
    R_{x}[N-1] & R_{x}[N-2] & \cdots  & R_{x}[0]
    \end{bmatrix}
    $\\

    Create noise subspace matrix $\hat{E}_{\text{sub},n}$ composed of noise eigenvectors of $\hat{R}_{\text{sub\_x}}$

    Useful component of received signal
    $\displaystyle
    b(\tau)=\vect{\left | S[0] \right |^{2} \left | S[1] \right |^{2}e^{j\left (\frac{2\pi}{K_{x}}  \right )\tau},\linebreak\left | S[K_{x}-1] \right |^{2} e^{j\left (\frac{2\pi}{K_{x}}  \right )(K_{x}-1)\tau)}}^{\top}
    $\\

     Truncate $b(\tau)$ to length N to form the subvector $b_{\text{sub}}(\tau)$

     Estimate $\tau$ for minimal projection onto  noise subspace
    $\displaystyle
    \tilde{\tau}=\argmaxA_\tau \left \{ \left \| b_{\text{sub}}^{H}(\tau)\hat{E}_{\text{sub},n} \right \|^{2} \right \}
    $\\

    \KwRet{$\tilde{\tau}$}
    
\end{algorithm}

\begingroup
\begin{table*}[htb]

\renewcommand{\arraystretch}{1.5} 
\caption{Overview of feature extraction literature.}
\label{tab:felit}
\scriptsize
\centering
\newlength\qfe
\setlength\qfe{\dimexpr .0833\textwidth -2\tabcolsep}

\noindent\begin{tabular}{>{\rowmac}p{16mm}>{\rowmac}p{16mm}>{\rowmac}p{7.25mm}>{\rowmac}p{16mm}>{\rowmac}p{7.25mm}>{\rowmac}p{16mm}>{\rowmac}p{7.25mm}>{\rowmac}p{16mm}>{\rowmac}p{7.25mm}>{\rowmac}p{16mm}>{\rowmac}p{7.25mm}>{\rowmac}p{16mm}<{\clearrow}}
\Xhline{1.25pt}
\multicolumn{1}{l}{\setrow{\bfseries}Direct} & \multicolumn{1}{l}{\setrow{\bfseries}Voltage Balance} & \multicolumn{6}{l}{\setrow{\bfseries}Centroid Detection} & \multicolumn{2}{l}{\setrow{\bfseries}Parametric}\setrow{}\\
\cmidrule(lr){1-1}\cmidrule(lr){2-2}\cmidrule(lr){3-8}\cmidrule(lr){9-10}\cmidrule(lr){11-12}
\cite{Lu2017a,Wang2019,Allgeier2009,Fan2013,Yousefian2017,Richie2015,Nurgizat2021,Frezza2022,Lu2017b,Tsuno2019,Kondo2024,Barnes2014,Hermoso2022b,Zhang2022b,Zhang2023,Hermoso2021,Springmann2014,Mafi2023,Pita2014} & \cite{Zhuo2020,Ortega2010,Delgado2010,Chang2006,Delgado2012b,Chen2007,Boslooper2017,Rodrigues2006,Bahrampoor2018,Maqsood2010,Kim2005,Delgado2011,Abhilash2014,Delgado2012a,Sozen2021,Ortega2009,Miao2017,Faizullin2017,Delgado2013,Lin2016,Hales2002,Strietzel2002,Leijtens2020,Sun2023,Hermoso2022a,Pentke2022,Soken2023,OKeefe2017,Wu2001,Wu2002} & BCTM & \cite{Chang2008,Chang2007a,He2005,Wei2017,Post2013,Jing2016,Fan2015,Wei2013,Bolshakov2020,TrebiOllennu2001,Maji2015,He2013,Rufino2004a,Rhee2012,ONeill2020,Antonello2018,Chen2006,KeQiang2020,Zhang2021,Kohli2019,Diriker2021,Wei2011,Fan2016,deBoom2006,Wei2014,Wang2015,Zhang2022a,Merchan2021,Fan2017} & MCAM & \cite{Rufino2005,Buonocore2005,Rufino2012,Li2012,Rufino2017,Mobasser2003,Liebe2001,Liebe2002,Rufino2009b,Rufino2008,Rufino2009a,Liebe2004} & DBCM & \cite{Xie2008,deBoer2013,Xie2014,deBoom2003,Xie2011,Xie2012,Xie2013,Xie2010,deBoer2017} & Linear phase & \cite{Enright2006b,Enright2007b,Enright2008b,Enright2008a,Godard2006,Enright2008c}\\
& & BCM & \cite{Enright2008b,Chang2008,Chang2007a} & PM & \cite{Coutinho2022} & IFM & \cite{Chang2008,Chang2007a,Pelemeshko2020} & Eigen-analysis & \cite{Enright2006b,Enright2008b}\\
& & PPE & \cite{Enright2008b,Chum2003,Enright2007a,Ali2011,Alvi2014,Radan2019} & ESCM & \cite{Farian2018,Bardallo2017b,Farian2022,Farian2015,Bardallo2017a,Merchan2023} & PD & \cite{Enright2006b} & &\\
& & MT-ACM & \cite{Massari2004} & BSCM & \cite{Saleem2017,Kim2021,Saleem2019} & TM & \cite{Chang2008,Chang2007a} & &\\
& & FMMS & \cite{You2011} & HT & \cite{Liu2016,Minor2010} & FEIC & \cite{Xing2008,Arshad2018} & & \\
\Xhline{1.25pt}
\end{tabular}

\bigskip

\renewcommand{\arraystretch}{1} 
\footnotesize
\newcommand*\rot[1]{\hbox to1em{\hss\rotatebox[origin=br]{-60}{#1}}}
\newcommand*\feature[1]{\ifcase#1 -\or\LEFTcircle\or\CIRCLE\fi}
\newcommand*\f[3]{\feature#1&\feature#2&\feature#3}
\makeatletter
\newcommand*\ex[9]{#1\tnote{#2}&#3&%
    \f#4&\f#5&\f#6&\f#7&\f#8&\f#9&\expandafter\f\@firstofone
}
\makeatother
\newcolumntype{G}{c@{}c@{}c}
\begin{threeparttable}
\caption{Feature extraction assessment summary.}
\label{tab:fesummary}

\begin{tabular}{@{}ll G !{\kern2.3em} GG !{\kern2.3em} GG !{\kern2.3em} GG@{}}
\toprule
\setrow{\bfseries}Feature Extraction  & \setrow{\bfseries}Case Study & \multicolumn{3}{c}{\setrow{\bfseries}Algorithm} & \multicolumn{6}{c}{\setrow{\bfseries}Architecture} & \multicolumn{6}{c}{\setrow{\bfseries}Knowledge} & \multicolumn{6}{c}{\setrow{\bfseries}Performance}\\
\midrule
&& \rot{Computationally light}
 & \rot{No pre-processing}
 & \rot{Algorithmically simple}
 & \rot{Pinhole Mask}
 & \rot{Slit Mask}
 & \rot{Other Mask}
 & \rot{CMOS Detector}
 & \rot{CCD Detector}
 & \rot{Other Detector}
 & \rot{Feature profile agnostic}
 & \rot{Feature pattern agnostic}
 & \rot{Multi-feature}
 & \rot{Feature tracking}
 & \rot{Blooming features}
 & \rot{Resolution scaling}
 & \rot{Sub-pixel resolution}
 & \rot{High precision}
 & \rot{Feature location agnostic}
 & \rot{Robust to noise}
 & \rot{Noise level agnostic}
 & \rot{Robust to missing features}\\
\Xhline{1.25pt}
\ex{Voltage Balance}{}{\citet{Boslooper2017}}
{222} {222}{002} {220}{000} {000}{000}\\
\midrule
\ex{PD}{}{\citet{Enright2006b}}
{222} {020}{020} {022}{001} {010}{000}\\
\ex{PPE}{}{\citet{Enright2007a}}
{220} {020}{020} {002}{001} {210}{000}\\
\ex{BCM}{}{\citet{Chang2008}}
{122} {200}{200} {220}{002} {200}{000}\\
\ex{BCTM}{}{\citet{He2005}}
{102} {200}{200} {220}{002} {200}{100}\\
\ex{MCAM}{}{\citet{Rufino2009b}}
{000} {200}{200} {002}{002} {220}{202}\\
\ex{DBCM}{}{\citet{Xie2014}}
{202} {200}{200} {220}{202} {200}{100}\\
\ex{MT-ACM}{}{\citet{Massari2004}}
{000} {200}{200} {200}{002} {200}{220}\\
\ex{PM}{}{\citet{Coutinho2022}}
{222} {200}{200} {220}{001} {000}{000}\\
\ex{ESCM}{}{\citet{Farian2022}}
{222} {020}{002} {220}{000} {200}{000}\\
\ex{BSCM}{}{\citet{Saleem2019}}
{102} {002}{200} {000}{222} {202}{102}\\
\ex{BHT}{}{\citet{Liu2016}}
{000} {002}{020} {002}{022} {202}{222}\\
\ex{CHT}{}{\citet{Adatrao2016}}
{000} {002}{020} {000}{002} {202}{222}\\
\ex{FMMS}{}{\citet{You2011}}
{000} {200}{200} {002}{202} {220}{202}\\
\ex{IFM}{}{\citet{Chang2008}}
{202} {200}{200} {220}{002} {200}{200}\\
\ex{TM}{}{\citet{Chang2008}}
{000} {200}{200} {000}{002} {200}{220}\\
\ex{FEIC}{}{\citet{Xing2008}}
{020} {200}{200} {002}{202} {220}{202}\\
\midrule
\ex{Linear-phase}{}{\citet{Enright2007b}}
{121} {020}{020} {220}{000} {200}{210}\\
\ex{Eigen-analysis}{}{\citet{Enright2006b}}
{020} {020}{020} {220}{000} {200}{220}\\
\Xhline{1.25pt}
\end{tabular}
\begin{tablenotes}
\item \hfil$\feature2=\text{provides feature}$; $\feature1=\text{partially provides feature}$;
$\text{\feature0}=\text{does not provide feature}$;
\end{tablenotes}    \end{threeparttable}
\begin{flushleft}
Abbreviations: BCM, Basic Centroiding Method; BCTM, Basic Centroiding Thresholding Method; BSCM, Black Sun Centroiding Method; CHT, Circle Hough Transform; DBCM, Double Balance Centroiding Method; ESCM, Event Sensor Centroiding Method; FEIC, Feature extraction image correlation; FMMS, Fast Multi-Point MEANSHIFT; HT, Hough transform; IFM, Image filtering method; MCAM, Multiple Centroid Averaging Method; MT-ACM, Multiple-Threshold Averaging Centroiding Method; PD, Peak Detection; PM, PixelMax; PPE, Peak Position Estimate; TM, Template method.
\end{flushleft}

\end{table*}
\endgroup

The work by Enright et al. \cite{Enright2006b} is reviewed as a case-study for the eigen-analysis method. In their study, the captured image is imported and the associated pixel matrix is generated $I[n]$. The processed image space is then converted into grayscale. Furthermore, the sampling spacing $\Delta X$ is set. The algorithm begins with the definitions of the signal models. The ideal signal is defined as $r_{1}(X)$ and the delayed signal is defined as $r_{2}(X)$. In addition, $a_{1}$ and $a_{2}$ are the corresponding signal amplitudes, $w(X)$ is the AWGN process, and $\tau$ is the delay of the received signal.

The above signals are then sampled at the sampling spacing $\Delta X$ as $r_{1}[n]$ and $r_{2}[n]$, where the sampling period is 1 and the length of the sequences is $N$. The signals $r_{1}[n]$ and $r_{2}[n]$ are zero-padded to length $K_{x} = 2N-1$. The algorithm proceeds with the definitions of the signal DFTs. The DFT of $s[n]$ is defined as $S[k]$ and the DFT $w[n]$ is defined as $W[k]$, respectively. The cross-correlation function between $r_{1}[n]$ and $r_{2}[n]$ is then found using circular correlation as $R_{r12}(\tau)$. The cross-correlation function reduces noise, thereby improving time-delay resolution.

Next, the cross-correlation series $R_{r12}[k]$ is zero-padded to length $2K_{x}-2$. Thereafter, the correlation function $R_{x}[l]$ is determined from the signal DFTs via the Wiener-Khinchine theorem. The truncated covariance matrix is then calculated as $\hat{R}_{\text{sub\_x}}$ from the previously defined correlation function. The noise subspace matrix $\hat{E}_{\text{sub},n}$ is created, which is composed of the noise eigenvectors of the truncated covariance matrix. An expression for the useful component of the received signal is represented by the vector function $b(\tau)$, since it has limited projection onto the noise subspace. The vector function is calculated as the squared magnitude of $S$ and a complex exponential term.

The subvector $b_{\text{sub}}(\tau)$ is formed by truncating the function $b(\tau)$ to length $N$. Lastly, the feature extraction estimate of vector $\tilde{\tau}$ is calculated as the minimal projection onto the noise subspace. The centroid estimate $\tilde{\tau}$ is returned as the output. The pseudocode for the technique is presented in Algorithm \ref{alg:eigen} \cite{Enright2006b}. \phantomsection\label{rqsum:rqsum3} 

\begin{description}
\item[RQ 3 Summary] \textit{The most commonly used feature extraction techniques in the literature implement some form of centroid detection with thresholding. Higher performing methods typically either capture multiple features or employ iterative thresholding methods.} (See \hyperref[rq:rq3]{RQ 3})
\end{description}

We provide a comparative summary of the feature extraction methods presented in this section in Table \ref{tab:fesummary}.

\section{Challenges and future directions} \label{sec:futuredir}

This section discusses the research gaps and associated challenges identified in this study. It examines the key research categories, their underlying motivations, primary approaches, potential limitations, and suggested directions for future work. A summary of the findings derived from this study is provided in Table \ref{tab:finalsummary}.


\subsection*{Architecture}

Novel sensor architectures enable improved sensor performance through the evolution of mask, detector, and instrument designs. Each of these design categories will be further discussed below.

\textbf{Mask.} 
The most recent advancements in sun sensor masks include the development of encoded masks and Fresnel zone plate (FZP) based masks. Encoded masks are designed to achieve both high accuracy and a wide field of view (FOV) within a single sensor package \cite{Wei2014}. A potential challenge lies in the complexity of the aperture array pattern and the difficulty of manufacturing the mask within required tolerances for good image quality. Currently, this is solved through configurations such as the varying and coded aperture approach \cite{Wang2015}. Additionally, recent compound eye microsystems, such as the LCE coded subeye aperture array, further improve the performance achievable compared with existing methods \cite{Zhang2021,Zhang2022a}. One approach to addressing the low image quality resulting from manufacturing the LCE mask is to instead use a coded microlens array, however a lensed array may be less suitable for space applications \cite{Zhang2022a}. Another option is to improve the sensor performance through ML-based correlation algorithms to better map the sensor errors due to diffraction.


FZP-based masks are a type of lensless mask that use diffraction-based focusing. This mask type offers a more compact and lightweight design compared to its refractive optical counterparts, such as the pinhole aperture. In the work by \citet{Lee2024}, the authors propose to replace the existing pinhole-based sun sensor design with a FZP-based mask. Some limitations identified with pinhole sun sensors include: the extended focal length and inaccuracies due to excessive light capture on the detector. The FZP-based mask enables a more compact and accurate sun sensor due to a shorter focal length and enhanced diffraction-based focusing. While the application of this method to sun sensors is still nascent, we recommend further research into diffraction-based optical masks due to their numerous potential advantages.

\textbf{Detector.} 
Advances in detector resolution and pixel pitch have steadily improved the accuracy of sun sensors. However, these enhancements often come at the cost of increased computational overhead and power consumption, driving the need for more efficient sensing and processing techniques. Methods such as region-based segmentation help mitigate this by eliminating the need to read the entire pixel matrix; nonetheless, even within the ROI, non-informative dark pixels still require readout. To address this limitation, asynchronous detectors, such as event-based sun sensors, have been developed to reduce bandwidth requirements by reading only illuminated pixels \cite{Bardallo2017a}.

One challenge with event-based sensors is their low spatial resolution. Reducing pixel size to improve resolution often degrades low-light sensitivity, increases fixed pattern noise (FPN), and raises power consumption. Currently, this resolution constraint is addressed in software by interpolating the light spot centroid using multiple peak responses rather than relying on a single winner. We recommend further exploration of hybrid event/frame-based approaches for sun sensors. This method combines the rich feature information available in traditional images with the sparse, high-speed data provided by event-based sensing, potentially offering a balance between accuracy and efficiency \cite{Qiao2024}.


\textbf{Instrument.} 

Novel changes to the fundamental instrument package have the potential to significantly enhance sensor performance. Traditional sun sensors are limited in that they can only measure two axes of information, thereby preventing them from independently determining the full spacecraft attitude. One solution to this limitation involves introducing a third axis of attitude information by measuring the Sun’s rotation axis using the Zeeman effect \cite{Liebe2016}. The primary challenge with this architecture lies in reducing the size of the resulting system. We believe that incorporating measurements of a third information axis represents a promising path toward next-generation sun sensor capabilities and recommend continued research in this area.


\subsection*{Model representations}

Advancements in model representation are closely tied to the increasing complexity of sensor architectures. This study identifies four categories of novel model representations: multiplexing, neural networks, multi-sensor fusion, and online calibration methods.


\textbf{Multiplexing.} Multi-aperture model representations offer improved performance by averaging multiple spot observations. An evolution of the multi-aperture model is the multiplexing model, in which the aperture patterns are uniquely coded for an unambiguous estimate. This approach is motivated by the need for highly accurate sensing over a wide FOV. Here, coding rules are developed for the associated mask configuration to map the feature space. The main challenges to this technique are model uncertainty, manufacturing errors and diffraction errors due to the complex nature of the mask.

This problem has been traditionally approached by applying coding rules for periodic, and coded and varying illumination patterns. More recently, a promising new representation was developed by Zhang et al. \cite{Zhang2021,Zhang2022a} to address update rate and FOV, in which a more dense and larger subeye array is modeled for a lensless compound eye (LCE) microsystem. We recommend research into techniques, such as deep learning, that could automatically encode the feature mapping process to better capture model uncertainties and associated manufacturing errors \cite{Soken2023,Sun2023}. Moreover, such techniques could use diffraction effects as features rather than relying on a priori optimization of aperture sizes.

\textbf{Neural network.} While the adoption of neural networks for sun sensor model representations is not new, the research is rapidly evolving for deep learning space applications \cite{Kothari2020}. The topic has already proven successful on other space attitude sensors such as a multilayer perceptron (MLP) for DLAS Earth horizon sensor \cite{Koizumi2018,Kikuya2023} and real-time convolutional neural networks (CNN) for star trackers \cite{Zhao2024}. 

Traditional modeling techniques are often inflexible and specific to a sensor architecture. Furthermore, model inputs are generally mapped to a reduced feature space, such as centroids. These gaps motivate the exploration of neural networks due to their inherent flexibility, ability to learn rich feature mappings, and resilience to sensor noise. Some applications of neural networks in the literature include an ANN-based direct centroid to angle mapping of a multi-aperture digital sun sensor \cite{Rufino2004a}, a deep neural network (DNN) based error compensation model for a QPD \cite{Sun2023}, and a DNN-based albedo correction process for an analog sun sensor (ANSS) \cite{Soken2023}.

However, neural network approaches for sun sensor calibration are impeded by the required availability of large amounts of training data, potential for over-fitting of data, reduction of model interpretability, and limited attention thus far in the literature. The availability of data large datasets for training neural networks is a challenge due to the slow and resource intensive process of running ground experiments. The development of a sun sensor digital twin could be used to efficiently generate a large synthetic dataset to better train a sun sensor calibration deep learning network \cite{Mukhopadhyay2021}.  

One way to improve the neural network (NN) training and inference speed is to implement a physics-informed neural network (PINN). This method combines data and a physics-based loss function in the NN learning process, however it requires knowledge of the underlying physics partial differential equation (PDE) \cite{Mata2023}. Since most sensor models use simple features like centroids, one potential future research direction is the use of richer feature spaces via CNNs. This approach could be especially useful for complex systems under many difficult-to-represent error sources, such as an encoded system. In this case, the system could be calibrated without manual characterization of the error sources and the coding rules would be automatically learned. In addition, the effects of diffraction could potentially be leveraged as a learned feature rather than mitigated.

Another research path is the development of DNNs for event-based sun sensors. Two potential network architectures for this task are asynchronous event-based graph neural networks (AEGNN) and spatio-temporal fusion spiking neural networks (STF-SNN). AEGNNs work by processing events as "evolving" spatio-temporal graphs \cite{Schaefer2022}. By only updating nodes affected by each new event, the event-by-event processing overhead and latency is greatly reduced. This network architecture could enable highly efficient event-based sun sensor processing. Next, we investigate the application of STF-SNNs, which fuse frame and event-based information together to combine the speed of event sensors and the high-resolution of frame-based sensors. The architecture works by combining two sensor fusion methods, feature-level fusion and decision-level fusion, to achieve spatio-temporal fusion. This network architecture could improve the accuracy limitations of current event-based sun sensors, while maintaining their low-latency operation.

Finally, we suggest the investigation of sparse network architectures, which can enable more efficient learning of high-dimensional and sparse data. These networks have proven useful in other domains, such as neutrino telescope data \cite{Yu2023}. In particular, frame-based digital sun sensor models struggle due to the inherent sparsity of the captured illumination images. The implementation of sparse submanifold convolutional neural networks (SSCNN) for digital sun sensors could greatly increase the inference speed, learning convergence, and mitigate the need for sun spot image segmentation. We see many opportunities for advancement with deep neural network-based methods and recommend further attention to this subject.

\textbf{Multi-sensor fusion.} The fusion of multiple sensor observations can be an economical way to improve the system accuracy and FOV compared to that of a single sensor. Furthermore, multi-modal sensor data synthesis can greatly extend the capabilities of a sensor. This problem has recently been approached through estimation via body mounted solar cells \cite{Hermoso2022b}. Moreover, multiple sensors and multi-aperture sensors are capable of performing distance measurements \cite{Barnes2014}. In fact, several fused sensors can accomplish three-dimensional target positioning \cite{Zhang2022a}.

\textbf{Online methods.} Traditionally, Kalman filtering (KF) methods have been used to great success in sun sensor online calibration. In particular, the extended Kalman filter (EKF), unscented Kalman filter (UKF), and cubature  Kalman filter (CKF) have been been implemented to improve non-linear attitude estimation. These classical methods excel in non-linear estimation, however they suffer from model and parameter mismatch and time-variations. Due to model complexity, a priori knowledge of the system model and parameters may not always be feasible. As such, Kalman filtering with model and parameter uncertainty has been addressed through filters such as the robust Kalman filter (RKF), adaptive Kalman filter (AKF), multiple model adapative estimation (MMAE) and other hybrid variants \cite{Bulut2012}. These techniques automatically correct the model and associated parameters based on the data, however they struggle to extract features from highly non-linear, noisy, and high-dimensional data \cite{Bai2023}.

The aforementioned challenges have motivated the development of hybrid Kalman filter and neural network models for state estimation. These approaches include external combination of KF and NN as the Kalman filter family neural network in succession (KFFNNS), and the internal integration of NN into the KF as the hybrid neural network trained Kalman filter family (NNTKFF). Hybrid models have been demonstrated to outperform both the NN model and KF families alone \cite{Feng2023}. We believe that the NNTKFF approach is a promising direction for future sun sensor online calibration. In particular, the combination of adaptive KF and DNN could address non-linearities, model mismatch, and time-variance for complex systems \cite{Revach2022,Revach2024,Cassinis2023}. Furthermore, the requirement to fully describe the KF state-space model limits its applicability for complex and high-dimensional digital sun sensor state estimation \cite{Buchnik2023,Buchnik2024}. The novel approach of a hybrid DNN-based encoding with learned KF in the latent space could address high-dimensional state estimation with partial domain knowledge, which is required for sun sensor estimation under uncertainties \cite{Buchnik2023,Buchnik2024,Becker2019}.

\subsection*{Feature extraction}
In order to achieve high-quality sensor mapping, highly accurate and rich feature capture is required.
We have identified two paths for future research directions for sun sensor feature extraction: improvements to classical centroiding techniques and the development of deep feature extraction methods.

\textbf{Classical feature extraction.} The sun sensor feature extraction process is traditionally applied through centroid detection with variations of thresholding methods to improve the centroid accuracy. Some examples of these current methods include BCTM \cite{He2005}, DBCM \cite{Xie2014}, and MT-ACM \cite{Massari2004}. Improvements to centroiding accuracy is limited by random noise generated during sun sensor operation. However, traditional centroid techniques for specific frames have reached diminishing returns for improvements \cite{Bao2022}. Therefore, time-domain extended image sequences can be exploited to further improve centroiding accuracy. Specifically, we identify adaptive energy filtering with time-domain energy sequences, as proposed by \citet{Bao2022}, as a potential research direction to improve sun sensor centroiding performance.

\textbf{Deep feature extraction.} Another direction to improve sun sensor feature extraction is through the use of deep learning models. Deep learning methods, such as CNNs, are able to more intelligently extract from a richer image feature space than their centroid detection counterparts. Moreover, they can also fuse the feature extraction and state mapping process into a single model for a more efficient and integrated approach. In fact, deep feature extraction has already been demonstrated for star tracker centroid detection and computation to great success \cite{Zapevalin2022,Zhao2024}. One such study by \citet{Zapevalin2022} surpassed the performance of classical centroiding by almost an order of magnitude. We believe that deep feature extraction is a promising research direction for both improved centroiding performance and direct mapping of complex illumination patterns in sun sensors.

\subsection*{Segmentation}

Sun sensor segmentation enables more efficient feature extraction by confining the region of pixels processed. The two main research directions identified include FOV and deep segmentation methods.

\textbf{Classical segmentation.} The most common sun sensor segmentation approaches are windowing and FOV subdivision. This is usually applied through thresholding techniques around a bounding box of some margin. Of these, FOV subdivision is especially popular for multi-aperture digital and multi-detector analog sensors. Multiple sub-FOVs can be spliced and processed to increase the FOV range while maintaining high accuracy throughout \cite{Zhang2022a}. However, these classical techniques are limited by the effects of image noise, dim and varying intensities, space debris, mask pollution and detector aging \cite{Xing2008}. One promising research direction is the segmenting of mask sub-regions by finding maximum correlation peaks with a mask template \cite{Zhang2022a}. The processing of dense and coded illumination patterns enables more accurate and robust segmentation.  

\textbf{Deep segmentation.} The use of deep learning for sun sensor image segmentation could address many of the shortcomings of classical segmentation techniques. This approach has already been successfully demonstrated for star tracker image segmentation using a U-Net architecture \cite{Mastrofini2023}. Furthermore, the approach improves segmentation accuracy even under noisy and dim image conditions without the need for re-calibration. In addition, semantic segmentation can add intelligent detection of image anomalies with masking during the segmentation process in the cases of space debris and mask pollution \cite{Mastrofini2023}. Finally, SSCNNs could enable segmentation-free digital sun sensor operation by ignoring non-lit pixels during processing. We recommend further research into the use of deep segmentation methods for sun sensor calibration, especially for the robust segmentation of complex illumination patterns under uncertain conditions.

\begingroup
\renewcommand{\arraystretch}{1.5} 
\begin{table*}[htb]
\caption{Overview of sun sensor calibration approaches.}
\label{tab:finalsummary}
\scriptsize
\centering
\newlength\qov
\setlength\qov{\dimexpr .167\textwidth -2\tabcolsep}
\noindent\begin{tabular}{>{\rowmac\raggedright}p{0.7\qov}>{\rowmac\raggedright}p{\qov}>{\rowmac\raggedright}p{\qov}>{\rowmac\raggedright}p{\qov}>{\rowmac\raggedright}p{\qov}>{\rowmac\raggedright\arraybackslash}p{1.3\qov}<{\clearrow}}
\hline
\multicolumn{2}{l}{\setrow{\bfseries}Categories} & \setrow{\bfseries}Main Motivation & \setrow{\bfseries}Approach & \setrow{\bfseries}Challenges & \setrow{\bfseries}Future Directions\\
\cline{1-2}
General & Specific & & & &\\
\Xhline{1.25pt}
\multirow[c]{2}{*}{ Architecture } & Mask & Accuracy, precision, FOV & Encoded mask & Manufacturing & LCE, coded microlens array, FZP-based mask\\
\cmidrule(lr){2-6}
& Detector & Temporal resolution, bandwidth, ROI-free & CMOS sensor,\hspace{0.5cm} event sensor & Spatial resolution & Winner interpolation, hybrid event/frame-based\\
\cmidrule(lr){2-6}
& Instrument & Lack of single sensor 3-axis estimate & Third attitude information axis & FOV, size, TRL & Solar rotation axis via Zeeman effect, other potential information axis\\
\hline
\multirow[c]{2}{*}{ \makecell{Model \\ Representation}} & Multiplexing & Accuracy, precision, FOV & Coding rules & Model uncertainty,
manufacturing errors, diffraction error & Periodic, varying and coded, LCE, CNN-based\\
\cmidrule(lr){2-6}
& Neural Network & Mask agnostic, automatic error characterization,\hspace{0.5cm} rich feature mapping, noise resilience  & Supervised learning of NN from experimental/ synthetic dataset & Computationally expensive, interpretability,\hspace{0.5cm} data availability, limited attention & DNN, CNN, PINN, SSCNN, AEGNN, STF-SNN\\
\cmidrule(lr){2-6}
& Multi-Sensor Fusion & FOV, cost, robustness  & Distributed sensing, multi-modal, mask-less & Limited accuracy, albedo errors & Solar cell, 3D target positioning\\
\cmidrule(lr){2-6}
& Online Calibration & Non-linearity, in-orbit calibration, noise tolerance, adaptive & EKF, UKF, CKF & Model and parameter uncertainty, time-variance, high-dimensionality & RKF, AKF, MMAE, KFFNNS, NNTKFF, hybrid variants\\
\hline
\multirow[c]{2}{*}{ \makecell{Feature \\ Extraction}} & Classical Feature Extraction & Accuracy, single-frame thresholding limits  &  Centroid detection and computation with thresholding & Time efficiency, random noise & Adaptive energy filtering with time-domain energy sequences\\
\cmidrule(lr){2-6}
& Deep Feature Extraction & Feature richness, generalizability, noise robustness, integrated model & Fused feature-model, supervised learning of NN & Interpretability, data availability, training time, limited attention & CNN, SSCNN\\
\hline
\multirow[c]{2}{*}{Segmentation} & Classical Segmentation & FOV, accuracy & Windowing, FOV subdivision & Image noise, dim and varying intensities, debris, pollution, aging & Segmenting of mask sub-regions via correlation operations with a mask template\\
\cmidrule(lr){2-6}
& Deep Segmentation & Accuracy, robustness, re-calibration not required, semantic masking & Image segmentation from learned experimental/ synthetic dataset & Interpretability, data availability, training time, limited attention & Semantic segmentation via U-Net, SSCNN as segmentation-free method\\
\hline
\multirow[c]{2}{*}{ \makecell{Model Trust}} & Adversarial & Security, reliability, blind trust, adversarial influence/control & LSI such as laser spoofing of sun vector & DSS gap, limited attention & Anomaly watchdog, spectral filtering, multi-sensor fusion, adversarial learning\\
\Xhline{1.25pt}
\end{tabular}
\begin{flushleft}
Abbreviations: AEGNN, Asynchronous Event-based Graph Neural Networks; AKF, Adaptive Kalman Filter; ANN, Artificial Neural Network; CKF, Cubature Kalman Filter; CMOS, complementary metal-oxide semiconductor; CNN, Convolutional Neural Network; DNN, Deep Neural Network; DSS, Digital Sun Sensor; EKF, Extended Kalman Filter; FOV, field of view; FZP, Fresnel zone plate; KFFNNS, Kalman Filter Family Neural Network in Succession; LCE, Lensless Compound Eye; LSI, Laser Signal Injection; MMAE, Multiple Model Adaptive Estimation; NN, Neural Network; NNTKFF, Neural Network Trained Kalman Filter Family; PINN, Physics-Informed Neural Network; RKF, Robust Kalman Filter; ROI, region of interest; SSCNN, Sparse Submanifold Convolutional Neural Network; STF-SNN, Spatio-Temporal Fusion Spiking Neural Network; TRL, Technology Readiness Levels; UKF, Unscented Kalman Filter.
\end{flushleft}
\end{table*}
\endgroup

\subsection*{Model trust}

\textbf{Adversarial.} The trustworthiness of sensor models vulnerable to adversarial attacks is of critical importance, as blind trust in sensor estimates can compromise the reliability and security of satellite operations. Spoofing attacks can allow adversarial influence or even direct control over sun sensors on-board. In particular, a high-powered laser could inject adversarial-controlled signals into sun sensor measurements, thereby spoofing a false sun vector \cite{Cyr2023}. While spoofing attacks have been successfully demonstrated on analog light sensors, executing such attacks over large distances in space remains challenging due to the need for irradiance levels comparable to sunlight for effective laser signal injection (LSI). However, we have identified a research gap due to the limited number of studies examining the effects of LSI on various digital sun sensor variants.

Currently, the presence of LSI attacks are primarily detected through anomaly detection schemes and optical sensing. Some possible directions for the mitigation of adversarial attacks include: anomaly watchdogs, spectral filtering and multi-sensor fusion. In addition, AI models are especially susceptible to adversarial attacks, however this could be mitigated through adversarial learning algorithms. We recommend further research into model-based approaches to detect and mitigate spoofing attacks on sun sensors, especially for digital sun sensors. \phantomsection\label{rqsum:rqsum4} 

\begin{description}
\item[RQ 4 Summary] \textit{The primary research gaps identified from this study include a lack of: (1) calibration algorithms that are mask agnostic or amenable to new sun sensor architectures, such as encoded, FZP masks, or event sensors; (2) deep learning calibration models applied to digital sun sensors, such as CNNs; (3) adaptive, high-dimensional deep online learning algorithms for sun sensor calibration; (4) feature extraction algorithms that use the full feature-space or are segmentation-free, such as SSCNNs; and (5) focus on adversarial threat mitigation for digital sensors.} (See \hyperref[rq:rq4]{RQ 4})
\end{description}

For a more detailed summary of the gaps and challenges for sun sensor calibration algorithms, see Table \ref{tab:finalsummary}.

\section{Conclusion} \label{sec:conc}

This survey presents a systematic mapping of 128 studies focused on sun sensor calibration algorithms. The calibration process is categorized into five key areas: sensor tasks, model representation, feature extraction, architecture, and performance. Additionally, a decision flow was developed and applied to guide the sun sensor selection process through the literature review. With this contribution, we aim to support researchers in identifying the most effective sun sensor calibration algorithms for their specific application needs, and to assist current practitioners in effectively implementing these algorithms.


Through our analysis, we identified accuracy as the most frequently prioritized sensor requirement, followed by cost and field of view. Notably, fine-accuracy sun sensors are the most prevalent, with coarse-accuracy sensors ranking second. Sensor requirements were found to strongly influence architecture selection. For example, digital sensors are the most commonly used detectors in fine-accuracy multi-aperture mask configurations, while photodiodes are the most common choice for coarse estimation, particularly in single-aperture and maskless designs. Similarly, identified error sources are closely tied to architecture selection, with alignment and manufacturing errors being the most prominent, followed by optical errors.
    

We also examined model representation approaches through case studies from the literature and compared various feature extraction techniques, including representative pseudocode. Our findings indicate that calibration algorithm selection is often driven by error sources, with specific feature extraction methods aligning closely with compatible model representations. Centroid detection combined with thresholding emerged as the most widely used feature extraction technique. Among model representations, geometric models were the most commonly implemented, followed by physics-informed and non-physical models---likely due to models often being developed for architecture-specific mask configurations.

From this review, we identified five key challenges in the current sun sensor calibration algorithm literature:
\begin{enumerate}
    \item There is a lack of publicly available datasets to test and train sun sensor calibration algorithms.
    \item Models are often tightly coupled to specific architectures and require manual error characterization.
    \item Online methods are limited by state-space model uncertainty and rarely integrate adaptive models with high-dimensional observation features.
    \item Feature extraction improvements face diminishing returns and are limited in their ability to leverage rich feature spaces for calibration mapping.
    \item There is a notable lack of research on adversarial sun sensor detection and correction approaches.
\end{enumerate}

Addressing these challenges holds significant potential to improve the accuracy, robustness, and operational reliability of sun sensor calibration for future space missions.

\appendix
\section{Data availability} \label{sec:appendixa}
The data compiled for this study is publicly available on \href{https://doi.org/10.5281/zenodo.15784429}{Zenodo} \cite{Herman2025a}. In addition, the data analysis is available on \href{https://public.tableau.com/app/profile/herman.michael/viz/SunSensorCalibrationAlgorithms/Dashboard}{Tableau Public} as an interactive dashboard \cite{Herman2025b}.


\bibliographystyle{model1-num-names}

\bibliography{cas-refs}





\end{document}